\documentclass{article}

\usepackage[final,arxiv]{neurips_2026}

\usepackage[utf8]{inputenc}
\usepackage[T1]{fontenc}
\usepackage{hyperref}
\usepackage{url}
\usepackage{booktabs}
\usepackage{amsfonts}
\usepackage{amsmath,amssymb}
\usepackage{nicefrac}
\usepackage{microtype}
\usepackage{xcolor}
\usepackage[table]{colortbl}
\usepackage{graphicx}
\usepackage{algorithm}
\usepackage{algpseudocode}
\usepackage{tikz}
\usetikzlibrary{positioning, arrows.meta, shapes.geometric, calc, decorations.pathreplacing}
\usepackage{multirow}
\usepackage{makecell}
\usepackage{subcaption}
\usepackage[capitalise,noabbrev]{cleveref}
\usepackage{placeins}  % FloatBarrier
\usepackage{etoolbox}  % \apptocmd / \pretocmd
\usepackage{arydshln}  % \hdashline, \cdashline, ':' column spec
\usepackage{xspace}
\usepackage{dsfont}
\usepackage{bbm}

\usepackage{amsmath,amsfonts,bm}

\def\eqref#1{equation~\ref{#1}}
\def\1{\bm{1}}

\def\mQ{{\bm{Q}}}

\DeclareMathAlphabet{\mathsfit}{\encodingdefault}{\sfdefault}{m}{sl}
\SetMathAlphabet{\mathsfit}{bold}{\encodingdefault}{\sfdefault}{bx}{n}

\definecolor{exprcolor}{HTML}{DA3832}
\definecolor{wprcolor}{HTML}{3C75AF}
\definecolor{promptcolor}{HTML}{000000}
\definecolor{draftcolor}{HTML}{000000}
\newcommand{\Gen}{\mathsf{G}}
\newcommand{\Ext}{\mathsf{E}}
\newcommand{\Guard}{\mathrm{Leak}}

\newcommand{\expr}{{\color{exprcolor}e}}
\newcommand{\exprest}{{\color{exprcolor}\hat{e}}}
\newcommand{\wpr}{{\color{wprcolor}w}}
\newcommand{\prompt}{{\color{promptcolor}\pi}}
\newcommand{\promptg}[1]{{\color{promptcolor}\prompt_{\Gen}^{#1}}}
\newcommand{\prompte}[1]{{\color{promptcolor}\prompt_{\Ext}^{#1}}}
\newcommand{\genqual}[1]{q^\Gen_{#1}}   % generation quality (col avg)
\newcommand{\extqual}[1]{q^\Ext_{#1}}   % extraction quality (row avg)

\newcommand{\errparse}{\ensuremath{\mathrm{Err}_{\mathrm{parse}}}}
\newcommand{\errarity}{\ensuremath{\mathrm{Err}_{\mathrm{arity}}}}
\newcommand{\errstruct}{\ensuremath{\mathrm{Err}_{\mathrm{struct}}}}
\newcommand{\errcontent}{\ensuremath{\mathrm{Err}_{\mathrm{content}}}}

\newcommand{\eg}{\emph{e.g.,}\xspace}
\newcommand{\ie}{\emph{i.e.,}\xspace}

\newcommand{\fmtwo}{Claude Haiku 4.5\xspace}

\newcommand{\rebuttal}[1]{{\color{draftcolor}#1}}

\definecolor{xavicolor}{HTML}{9966FF}

\title{The Communication Bottleneck: A Round-Trip Study of \rebuttal{Tree-Structured Expression} Serialization in Language Models}

\author{
  Xavier Suau \\
  Apple \\
  \And
  Alex Ferrando \\ 
  Apple \\
  \And
  Luca Zappella \\
  Apple \\
  \And
  Samy Bengio \\
  Apple \\
}

\begin{document}

\maketitle

% I find the abstract above lacking a bit of details. Here an alternative tweak of the abstract above.
\begin{abstract}
When language models reason in chain-of-thought or \rebuttal{exchange free-text intermediates}, they serialize structured information into natural language. How much \rebuttal{tree-structured compositional content} survives this bottleneck? We propose a round-trip protocol that answers this question empirically for \rebuttal{tree-structured expressions}. A generator converts a procedurally generated arithmetic expression into a word problem, a separate extractor recovers the expression from the word problem alone, and symbolic equivalence provides an exact oracle. Evaluating all pairwise combinations of sixteen models yields a communication matrix whose marginals
separate generation quality from extraction quality. Three main findings emerge. First, the channel is lossy and asymmetric: swapping which model generates and which extracts shifts accuracy by up to 60.4 points, and the best pair reaches 92.9\% by combining different models on each end rather than the same model on both. Second, at least 73.6\% of round-trip failures originate at generation, and difficulty is driven by tree structure (operator count, depth, right-branching) rather than model family. Third, the channel is trainable: $\sim\!3600$ fine-tuning examples that share the evaluation's operators and tree shapes lift every open-weight model above untrained Gemini-3.1-Pro, an upper bound under matched semantics. A disjoint-domain regime with new operators and vocabulary also raises every open-weight model, confirming the gain is not an artifact of matched semantics, though a gap to the frontier remains. Together these results identify \rebuttal{tree-structured expression} serialization as a primary limiting factor when models communicate hierarchical structure through natural language.
\end{abstract}

\begin{figure}[htb]
  \centering
  \includegraphics[width=0.75\linewidth]{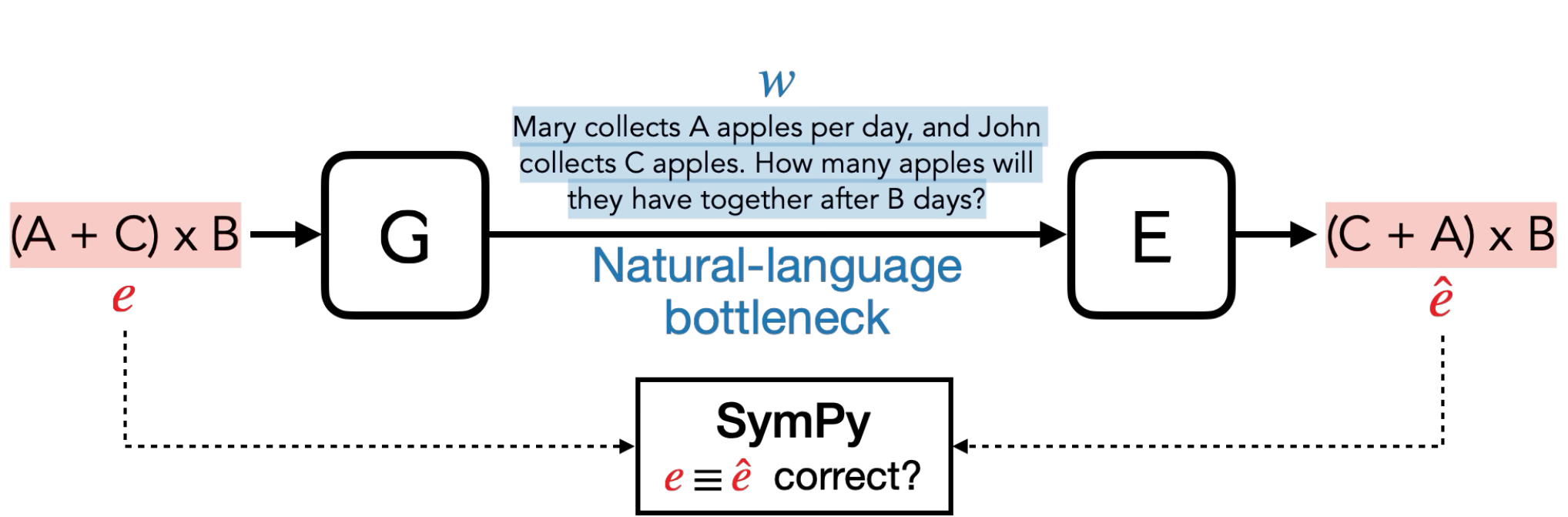}
     \caption{\textbf{The round-trip protocol.} A generator~$\Gen$ serializes an
     arithmetic expression~$\expr$ (a tree) into a word problem~$\wpr$ that
     passes through a {\color{wprcolor}natural-language bottleneck}. A separate
     extractor~$\Ext$ recovers~$\exprest$ from~$\wpr$ alone, and SymPy serves
     as an exact oracle for symbolic equivalence $\expr \equiv \exprest$. Any
     compositional detail the generator fails to encode is irreversibly lost,
         regardless of~$\Ext$.}
  \label{fig:protocol}
\end{figure}

\section{Introduction}

Language models routinely serialize structured information into
natural language (NL): chain-of-thought reasoning linearizes intermediate
computations into prose~\citep{wei2022chain}, model outputs are
consumed by downstream models in
pipelines~\citep{wu2024autogen,hong2023metagpt}, and tool-use
descriptions encode structured calls as
text~\citep{schick2023toolformer}. If this
serialization is lossy~\citep{herbst2025lost}, any structure that the encoding step fails to
express unambiguously is irreversibly lost, regardless of the
downstream consumer's capability. How much compositional structure
actually survives the natural-language bottleneck? \rebuttal{We
study this for hierarchical structure using tree-structured
expressions built from arithmetic and logical operators, which admit
an exact correctness oracle via symbolic equivalence.}

Answering this question requires measuring communication \emph{between}
models, but existing benchmarks are not designed for it. Whether
evaluating math~\citep{cobbe2021gsm8k,hendrycks2021math,glazer2025frontiermathbenchmarkevaluatingadvanced},
reasoning~\citep{rein2023gpqagraduatelevelgoogleproofqa,phan2025lastexam},
or code~\citep{jimenez2024swebench,jain2024livecodebench}, they test
single models on fixed problem sets. 
The closest prior art is \citet{allamanis2024unsupervised}, who round-trip code through natural language but deliberately keep encoder and decoder fixed to the same model to sidestep what they term the `communication chasm' between distinct models. We instead treat that chasm as the object of study: separating generator from extractor across an $N{\times}N$ grid of models yields a measurement instrument that exposes role asymmetry, fault attribution, and structural predictors of failure, all invisible under same-model analysis.

% One might ask why study natural language at all, since structured formats
% (code, JSON, S-expressions) can carry compositional content without
% ambiguity. We agree, and recommend structured formats when feasible
% (\Cref{sec:discussion}). But natural language remains the \emph{de
% facto} channel in all the settings listed above, despite evidence that
% alternative formats can improve both reasoning and
% communication~\citep{chen2024beyond}. 
One might ask why study natural language at all. While structured formats like JSON, code or S-expressions natively preserve compositional hierarchy and should be utilized whenever feasible (\Cref{sec:discussion}), they do not reflect how models typically operate. 
Natural language remains the \emph{de
facto} channel in all the settings listed above, despite evidence that
alternative formats can improve both reasoning and
communication~\citep{chen2024beyond}. 
Measuring NL compositional fidelity requires a domain with an exact correctness oracle,
independent control over structural complexity, and freedom from
contamination~\citep{zhang2024careful}. Arithmetic satisfies all
three: (i) symbolic equivalence gives a ground-truth oracle, (ii) expression
trees expose operator count, nesting depth, and branching topology
as controllable parameters, and (iii) instances can be procedurally
generated. 

% To quantify the channel's capacity for compositional structure, we
% propose a \textbf{round-trip protocol} (\Cref{fig:protocol}). A
% generator~$\Gen$ encodes an arithmetic expression~$\expr$, sampled
% from an enumerated space of tree skeletons, into a word
% problem~$\wpr$. A separate extractor~$\Ext$ recovers~$\exprest$ from
% $\wpr$ alone, and success is verified by symbolic equivalence. A
% rule-based guard ensures $\wpr$ contains no mathematical notation,
% forcing all structural information through a linguistic
% channel~(\Cref{sec:expressions}). Evaluating $N$ models in both
% roles yields an $N {\times} N$ \emph{communication matrix} whose
% marginals decompose performance into generation and extraction
% quality~(\Cref{sec:matrix}).
% We replicate the protocol on propositional logic as a robustness
% check (\Cref{app:logic}). Extending it to other structured data
% such as planning or algebra requires domain-specific oracles.
% We evaluate 15 models: 12 open-weight (0.6B--32B) across five
% families and 3 frontier, on 1790 expressions spanning operator
% counts 2--8 and depths 2--5. Our contributions are:

   To quantify the channel's capacity for compositional structure, we
   propose a \textbf{round-trip protocol} (\Cref{fig:protocol}). A
   generator~$\Gen$ encodes an arithmetic expression~$\expr$, sampled
   from an enumerated space of tree skeletons, into a word
   problem~$\wpr$. A separate extractor~$\Ext$ recovers~$\exprest$ from
   $\wpr$ alone, verified by symbolic equivalence. A rule-based guard
   ensures $\wpr$ contains no mathematical notation, forcing structure
   through a linguistic channel~(\Cref{sec:expressions}). Evaluating $N$
   models in both roles yields an $N {\times} N$ \emph{communication
   matrix} whose marginals decompose performance into generation and
   extraction quality~(\Cref{sec:matrix}).
   We replicate the protocol on propositional logic as a robustness
   check (\Cref{app:logic}).
   
   We evaluate 16 models: 12 open-weight (0.6B--32B) across five
   families and 4 frontier, on 2450 expressions with operator counts
   2--8 and depths 2--6. Our contributions are:
\begin{itemize}

\item \textbf{A round-trip protocol for communication fidelity.} A
  generator encodes a procedurally generated expression as a word
  problem, an extractor recovers it, and symbolic equivalence serves
  as an exact oracle. Evaluating $N$ models in both roles yields an
  $N {\times} N$ communication matrix that decomposes pairwise
  performance into generation and extraction
  quality~(\Cref{sec:design}). Code will be released upon acceptance.

\item \textbf{Generation is the bottleneck, not extraction.} At least
  73.6\% of round-trip failures originate at the generation step. Role
  assignment alone shifts accuracy by up to 60.4~pp, and the highest
  fidelity is achieved by a cross-model pair, not by
  self-communication~(\Cref{sec:comm_matrix_results,sec:why_fails}).

\item \textbf{Failure is governed by tree structure, not model
  family.} Operator count, nesting depth, and a linearization
  complexity measure $\ell(T)$ dominate failure prediction.
  Right-branching trees yield lower accuracy than left-branching ones of
  identical size, and $\ell(T)$ captures this asymmetry
  directly~(\Cref{sec:why_fails}). A propositional-logic replication
  preserves the main rankings and fault pattern~(\Cref{app:logic}).

\item \textbf{The bottleneck is compositional serialization, not
  domain knowledge.} Fine-tuning on a novel operator vocabulary with no arithmetic overlap lifts every open-weight generator, and in-domain fine-tuning on approximately 3600 examples pushes all of them above the strongest untrained frontier generator. The bottleneck is trainable under matched semantics and partially trainable across domains~(\Cref{sec:finetuning}).

% \item \textbf{The bottleneck is compositional serialization, not
%   domain knowledge.} Fine-tuning on a novel operator vocabulary with
%   no arithmetic overlap transfers the linearization skill to weaker
%   models, while in-domain fine-tuning on approximately 3600 examples
%   lifts every open-weight model above the strongest untrained frontier
%   generator~(\Cref{sec:finetuning}).

\end{itemize}

% ─────────────────────────────────────────────────────────────────────────────
\section{Protocol Design}
\label{sec:design}
% ─────────────────────────────────────────────────────────────────────────────

% ─── 2.1 ─────────────────────────────────────────────────────────────────────
\subsection{Expression Suite}
\label{sec:expressions}

\begin{figure}[htb]
\centering
\begin{tikzpicture}[
    level distance=5mm,
    sibling distance=7mm,
    every node/.style={font=\small},
    skel/.style={circle, draw, fill=gray!30, minimum size=3mm, inner sep=0pt},
    sleaf/.style={rectangle, draw, fill=gray!10, minimum size=3mm, inner sep=0pt},
]
% --- Skeleton 1: (□ ○ □) ○ □ ---
\node[skel] (s1) at (0, 0) {}
    child { node[skel] {}
        child { node[sleaf] {} }
        child { node[sleaf] {} }
    }
    child { node[sleaf] {} };
\node[above=2pt of s1, font=\scriptsize\itshape] {$k{=}2, d{=}2$};
\node[anchor=west, font=\scriptsize, align=left] at (0.9, -0.7) {%
  $(A{+}B){\times}C$\\
  $(D{/}3){-}E$\\
  $(F{\times}G){+}7$};%\\
  %$\ldots$;

% --- Separator 1 ---
\draw[gray!40, line width=0.5pt] (2.8, 0.6) -- (2.8, -1.7);

% --- Skeleton 2: □ ○ (□ ○ □) ---
\node[skel] (s2) at (4.2, 0) {}
    child { node[sleaf] {} }
    child { node[skel] {}
        child { node[sleaf] {} }
        child { node[sleaf] {} }
    };
\node[above=2pt of s2, font=\scriptsize\itshape] {$k{=}2, d{=}2$};
\node[anchor=west, font=\scriptsize, align=left] at (5.1, -0.7) {%
  $A{-}(B{\times}C)$\\
  $F{/}(D{+}2)$\\
  $H{+}(E{\times}G)$};

% --- Separator 2 ---
\draw[gray!40, line width=0.5pt] (7.0, 0.6) -- (7.0, -1.7);

% --- Skeleton 3: ((□ ○ □) ○ □) ○ □ --- (slightly smaller to match height)
\begin{scope}[level distance=5mm, sibling distance=6mm]
\node[skel] (s3) at (8.6, 0) {}
    child { node[skel] {}
        child { node[skel] {}
            child { node[sleaf] {} }
            child { node[sleaf] {} }
        }
        child { node[sleaf] {} }
    }
    child { node[sleaf] {} };
\end{scope}
\node[above=2pt of s3, font=\scriptsize\itshape] {$k{=}3, d{=}3$};
\node[anchor=west, font=\scriptsize, align=left] at (9.5, -0.7) {%
  $((A{+}B){\times}C){-}D$\\
  $((F{/}3){+}G){\times}H$\\
  $((E{-}4){\times}A){+}B$};

\end{tikzpicture}
\caption{\textbf{Expression generation via skeleton enumeration.}  Circles
denote operator nodes, squares denote leaf positions.  We fill each operator node with a random
element of $\{+,-,\times,\div\}$ and each leaf with a variable or
constant, producing distinct expressions that share the same structure.
The first two skeletons share $(k{=}2, d{=}2)$ but differ in branching
direction; the third has $(k{=}3, d{=}3)$.}
\label{fig:skeletons}
\end{figure}
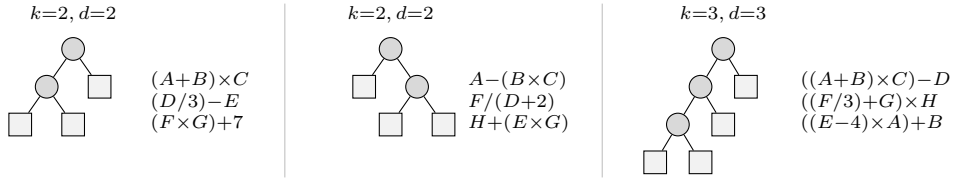

Standard math-word-problem datasets such as GSM8K offer no control over
structural complexity: expressions are hand-authored with uncontrolled
nesting depth, operator count, and tree shape. We instead generate
expressions procedurally from abstract syntax tree \emph{skeletons}.
Given a target operator count~$k$ and depth~$d$, we recursively
enumerate every binary-tree skeleton matching the pair $(k,d)$. Each skeleton is then instantiated by assigning operators
uniformly from $\{+,-,\times,\div\}$ to internal nodes and filling
leaves with named variables from $\{A,B,\ldots,H\}$ (probability~0.7)
or integer constants in $[1,10]$ (probability~0.3). Locally degenerate subexpressions are rejected at instantiation: same-leaf
patterns under a non-commutative operator ($X-X$, $X/X$) and multiplicative
identities ($X\times 1$, $1\times X$, $X/1$).
The operator set, variable alphabet, and leaf-type probabilities are
free parameters of the procedure. The values above are reasonable
defaults, not tuned choices. Since the skeleton fixes
the topology before instantiation, every resulting expression has the
target $k$ and $d$ by construction (\Cref{fig:skeletons}), enabling the
factorial analyses of depth, operator count, and tree shape reported in
\Cref{sec:results}.

% ─── 2.2 ─────────────────────────────────────────────────────────────────────
\subsection{Round-Trip Protocol}
\label{sec:protocol}

   The round-trip
   $\expr \! \xrightarrow{\Gen} \! \wpr \! \xrightarrow{\Ext} \! \exprest$
   (\Cref{fig:protocol}) proceeds as follows.
   A \textbf{generator}~$\Gen$, prompted with $\promptg{\gamma}$ containing
   $\gamma$ in-context $(\expr, \wpr)$ exemplars, produces a short word
   problem $\wpr = \Gen(\expr;\promptg{\gamma})$ using variable placeholders
   and no mathematical symbols. A rule-based \emph{leakage guard}
   ($\Guard:\wpr\mapsto\{0,1\}$) rejects outputs that leak mathematical
   syntax, ensuring structural information passes through a genuinely
   linguistic channel. Outputs missing required placeholders or
   exhibiting degenerate repetition are counted separately as generator
   faults (\Cref{app:guard}). Guard false negatives bias round-trip
   accuracy upward, so the reported figures are conservative
   estimates of channel fidelity.
   An \textbf{extractor}~$\Ext$, prompted with $\prompte{\varepsilon}$
   containing $\varepsilon$ exemplars, recovers
   $\exprest = \Ext(\wpr;\prompte{\varepsilon})$, verified by symbolic
   equivalence $\texttt{sympy.simplify}(\exprest - \expr) = 0$~\citep{sympy},
   which accepts valid rearrangements (\eg $A+B$ vs.~$B+A$) but catches
   all semantic errors. Any detail the generator fails to express
   unambiguously is irreversibly lost, regardless of the extractor.
   Prompts are detailed in \Cref{app:prompts}, worked examples in
   \Cref{app:examples}. Shot exemplars are drawn from a pool disjoint
   from the evaluation set, and the shot counts $\gamma, \varepsilon$
   are fixed in \Cref{sec:setup}.
   Because our contribution is the protocol itself, $\promptg{\gamma}$ and
   $\prompte{\varepsilon}$ are parameters rather than fixed choices, and
   prompt tuning is orthogonal to the measurements we report.

% ─── 2.3 ─────────────────────────────────────────────────────────────────────
\subsection{Communication Matrix}
\label{sec:matrix}

Given $N$ models, each serving as both $\Gen{}$ and $\Ext{}$, the protocol
produces an $N {\times} N$ communication matrix $\mQ$ whose entry
$q_{ij}$ is the round-trip accuracy when extractor~$i$ reads
generator~$j$'s word problems. For each $\expr \in D$, generator~$j$
produces $\wpr = \Gen_j(\expr)$. If $\wpr$ fails the leakage guard
($\Guard(\wpr) = 0$), the round trip scores zero, otherwise
extractor~$i$ recovers $\exprest = \Ext_i(\wpr) = \Ext_i\big(\Gen_j(\expr)\big)$:
\begin{equation}
  q_{ij}
  \;=\;
  \frac{1}{|D|}
  \sum_{\expr \in D}
  \Guard\big(\Gen_j(\expr)\big)
  \;\cdot\;
  \mathds{1}\!\bigl[\,
    \Ext_i\big(\Gen_j(\expr)\big) \equiv \expr
  \,\bigr].
  \label{eq:accuracy}
\end{equation}
The column average $\genqual{j} = \frac{1}{N}\sum_i q_{ij}$\vspace{-0.5mm} measures
the \textbf{generation quality} of model~$j$: how decodable its word
problems are across extractors. The row average
$\extqual{i} = \frac{1}{N}\sum_j q_{ij}$ measures the
\textbf{extraction quality} of model~$i$: how reliably it recovers
expressions across generators. The diagonal entry $q_{ii}$ is the
\textbf{self-communication} score.  
The denominator is the full suite~$|D|$, not each generator's
guard-passing subset, so guard failures count as communication
failures, eliminating the selection bias from evaluating generators
only on expressions they can encode. A consequence is that $\extqual{i}$ is panel-relative: guard failures enter as zeros in every extractor row, so absolute values reflect the generator mix. This balanced design, in which every
$\Gen/\Ext$ pair is evaluated on the same expression set, ensures that
marginal averages are fair comparisons. A latent-ability analysis
confirming that raw marginals track maximum-likelihood skill estimates
is reported in \Cref{app:irt}.

\subsection{Fault Attribution}
\label{sec:fault_method}

When a round trip fails ($\exprest\not\equiv\expr$), the error may
originate at the generation or extraction step. Because the expression actually
encoded by~$\wpr$ is unobserved, we estimate it via multi-extractor
consensus. For a word problem $\wpr = \Gen_j(\expr)$, we define the
\emph{extractor vote} for any candidate expression~$e'$ as
\begin{equation}
  P(e' \mid \wpr)
  \;=\;
  \frac{1}{N}\sum_{i=1}^{N}
    \mathds{1}\!\bigl[\Ext_i(\wpr) \equiv e'\bigr].
  \label{eq:consensus}
\end{equation}
If $P(\expr \mid \wpr) > 0$, at least one extractor recovers the
target, proving the word problem faithful, and any remaining failures are considered 
\textbf{extractor faults}. If $P(\expr \mid \wpr) = 0$, no extractor
succeeds and we classify the failure as a \textbf{generator fault}.
The severity of the generator fault is gauged by the \textbf{extraction
agreement} $p^{*}(\wpr) = \max_{e'} P(e' \mid \wpr)$,  \ie the fraction of
extractors that converge on the most common output.
High~$p^{*}$ means most extractors recover the same wrong expression,
indicating that the generator produced a coherent word problem that
systematically describes a different expression than~$\expr$.
Low~$p^{*}$ means extractors disagree, indicating that the word
problem is garbled or ambiguous. Attribution results are reported in \Cref{sec:fault_results}.

A generation counts as faithful if \emph{any} of the $N{=}16$
extractors succeeds, so enlarging the panel can only shift failures
from the generator-fault bucket to the extractor-fault bucket.
The reported generator-fault rate is therefore conservative with
respect to the bottleneck claim: every smaller sub-panel we tested
reports a higher rate, up to $91.1\%$. Misattribution is
negligible at this $N$ (\Cref{app:fault_sensitivity}).

\begin{table}[htb]
\centering
\caption{\textbf{The $\boldsymbol{16 \times 16}$ communication matrix on all generated expressions (guard failures counted as errors).} Each cell shows round-trip accuracy (\%).  Rows are extractors, columns are generators.  The rightmost column and bottom row show extraction quality (row average) and generation quality (column average). Generator and extractor are run with prompts $\promptg{0}$ and $\prompte{\varepsilon}$ respectively (configuration $\gamma{=}0, \varepsilon{=}3)$; see \Cref{app:shot_grid} for the full ablation. The $\pm$ values denote standard deviation across the 16 pairings in each row or column. The $\Guard$~\% row reports each model's leakage-guard pass rate as a generator. Thin rules separate open-weight auto-regressive, diffusion, and frontier models.}
\vspace{2mm}
\label{tab:matrix}
\small
\setlength{\tabcolsep}{2pt}
\setlength{\dashlinedash}{1pt}
\setlength{\dashlinegap}{1pt}
\resizebox{\textwidth}{!}{%
\begin{tabular}{c l c c c c c c c c c c !{\;\vrule width 0.3pt\;} c c !{\;\vrule width 0.3pt\;} c c c c !{\;\vrule width 0.75pt\;} c}
\toprule
\multicolumn{19}{c}{\textsc{Generators}} \\
 \multirow{20}{*}{\rotatebox{90}{\textsc{Extractors}}} & & \rotatebox{90}{\scriptsize Qwen3-0.6B} & \rotatebox{90}{\scriptsize Qwen3-1.7B} & \rotatebox{90}{\scriptsize Qwen3-4B} & \rotatebox{90}{\scriptsize Qwen3-8B} & \rotatebox{90}{\scriptsize Qwen3-14B} & \rotatebox{90}{\scriptsize Qwen3-32B} & \rotatebox{90}{\scriptsize Gemma-3-4B} & \rotatebox{90}{\scriptsize Gemma-3-12B} & \rotatebox{90}{\scriptsize Gemma-3-27B} & \rotatebox{90}{\scriptsize Phi-4} & \rotatebox{90}{\scriptsize Dream-7B} & \rotatebox{90}{\scriptsize LLaDA-8B} & \rotatebox{90}{\scriptsize C-Haiku-4.5} & \rotatebox{90}{\scriptsize GPT-5} & \rotatebox{90}{\scriptsize G-3-Flash} & \rotatebox{90}{\scriptsize G-3.1-Pro} & \rotatebox{90}{\scriptsize Extr.\ quality} \\
\cmidrule[0.5pt](l){2-19}
& {\scriptsize $\Guard$ pass \%} & \cellcolor[RGB]{229,247,229}{\scriptsize 20} & \cellcolor[RGB]{178,232,178}{\scriptsize 60} & \cellcolor[RGB]{135,219,135}{\scriptsize 94} & \cellcolor[RGB]{130,217,130}{\scriptsize 98} & \cellcolor[RGB]{128,217,128}{\scriptsize 99} & \cellcolor[RGB]{129,217,129}{\scriptsize 99} & \cellcolor[RGB]{129,217,129}{\scriptsize 98} & \cellcolor[RGB]{132,218,132}{\scriptsize 96} & \cellcolor[RGB]{132,218,132}{\scriptsize 96} & \cellcolor[RGB]{131,218,131}{\scriptsize 97} & \cellcolor[RGB]{173,230,173}{\scriptsize 64} & \cellcolor[RGB]{133,218,133}{\scriptsize 95} & \cellcolor[RGB]{128,216,128}{\scriptsize 100} & \cellcolor[RGB]{127,216,127}{\scriptsize 100} & \cellcolor[RGB]{127,216,127}{\scriptsize 100} & \cellcolor[RGB]{127,216,127}{\scriptsize 100} &  \\
\cmidrule[0.5pt](l){2-19}
& {\scriptsize Qwen3-0.6B} & \cellcolor[RGB]{249,251,254}{\scriptsize 0.0} & \cellcolor[RGB]{227,235,253}{\scriptsize 1.2} & \cellcolor[RGB]{210,223,252}{\scriptsize 3.2} & \cellcolor[RGB]{201,217,251}{\scriptsize 4.5} & \cellcolor[RGB]{208,222,252}{\scriptsize 3.4} & \cellcolor[RGB]{218,229,252}{\scriptsize 2.2} & \cellcolor[RGB]{246,248,254}{\scriptsize 0.1} & \cellcolor[RGB]{239,244,254}{\scriptsize 0.4} & \cellcolor[RGB]{241,245,254}{\scriptsize 0.3} & \cellcolor[RGB]{200,216,251}{\scriptsize 4.7} & \cellcolor[RGB]{236,242,253}{\scriptsize 0.5} & \cellcolor[RGB]{204,219,252}{\scriptsize 4.0} & \cellcolor[RGB]{187,207,251}{\scriptsize 7.2} & \cellcolor[RGB]{220,230,252}{\scriptsize 1.9} & \cellcolor[RGB]{220,230,252}{\scriptsize 1.9} & \cellcolor[RGB]{223,232,253}{\scriptsize 1.6} & \cellcolor[RGB]{216,228,252}{\scriptsize 2.3\scalebox{0.55}{$\pm$2}} \\
& {\scriptsize Qwen3-1.7B} & \cellcolor[RGB]{249,251,254}{\scriptsize 0.0} & \cellcolor[RGB]{221,231,253}{\scriptsize 1.8} & \cellcolor[RGB]{185,206,250}{\scriptsize 7.7} & \cellcolor[RGB]{171,195,250}{\scriptsize 11.2} & \cellcolor[RGB]{176,199,250}{\scriptsize 9.9} & \cellcolor[RGB]{184,205,250}{\scriptsize 7.9} & \cellcolor[RGB]{237,242,253}{\scriptsize 0.5} & \cellcolor[RGB]{219,230,252}{\scriptsize 2.0} & \cellcolor[RGB]{226,234,253}{\scriptsize 1.3} & \cellcolor[RGB]{159,187,249}{\scriptsize 14.4} & \cellcolor[RGB]{229,237,253}{\scriptsize 1.0} & \cellcolor[RGB]{178,201,250}{\scriptsize 9.2} & \cellcolor[RGB]{130,166,247}{\scriptsize 24.9} & \cellcolor[RGB]{170,195,250}{\scriptsize 11.3} & \cellcolor[RGB]{168,194,249}{\scriptsize 11.8} & \cellcolor[RGB]{162,189,249}{\scriptsize 13.6} & \cellcolor[RGB]{184,204,250}{\scriptsize 8.0\scalebox{0.55}{$\pm$7}} \\
& {\scriptsize Qwen3-4B} & \cellcolor[RGB]{249,251,254}{\scriptsize 0.0} & \cellcolor[RGB]{213,225,252}{\scriptsize 2.8} & \cellcolor[RGB]{162,189,249}{\scriptsize 13.6} & \cellcolor[RGB]{142,175,248}{\scriptsize 20.3} & \cellcolor[RGB]{139,173,248}{\scriptsize 21.3} & \cellcolor[RGB]{145,177,248}{\scriptsize 19.0} & \cellcolor[RGB]{234,240,253}{\scriptsize 0.7} & \cellcolor[RGB]{208,222,252}{\scriptsize 3.4} & \cellcolor[RGB]{210,223,252}{\scriptsize 3.1} & \cellcolor[RGB]{126,164,247}{\scriptsize 26.3} & \cellcolor[RGB]{218,229,252}{\scriptsize 2.1} & \cellcolor[RGB]{155,184,249}{\scriptsize 15.8} & \cellcolor[RGB]{79,131,244}{\scriptsize 49.1} & \cellcolor[RGB]{98,144,245}{\scriptsize 39.0} & \cellcolor[RGB]{107,151,246}{\scriptsize 34.6} & \cellcolor[RGB]{85,135,245}{\scriptsize 46.0} & \cellcolor[RGB]{147,178,248}{\scriptsize 18.6\scalebox{0.55}{$\pm$16}} \\
& {\scriptsize Qwen3-8B} & \cellcolor[RGB]{247,249,254}{\scriptsize 0.1} & \cellcolor[RGB]{214,226,252}{\scriptsize 2.7} & \cellcolor[RGB]{160,188,249}{\scriptsize 14.2} & \cellcolor[RGB]{143,176,248}{\scriptsize 19.8} & \cellcolor[RGB]{138,172,248}{\scriptsize 21.7} & \cellcolor[RGB]{141,174,248}{\scriptsize 20.7} & \cellcolor[RGB]{232,239,253}{\scriptsize 0.8} & \cellcolor[RGB]{210,223,252}{\scriptsize 3.1} & \cellcolor[RGB]{206,220,252}{\scriptsize 3.8} & \cellcolor[RGB]{123,162,247}{\scriptsize 27.6} & \cellcolor[RGB]{219,229,252}{\scriptsize 2.0} & \cellcolor[RGB]{153,183,249}{\scriptsize 16.5} & \cellcolor[RGB]{79,131,244}{\scriptsize 49.0} & \cellcolor[RGB]{88,137,245}{\scriptsize 44.2} & \cellcolor[RGB]{98,144,245}{\scriptsize 39.1} & \cellcolor[RGB]{73,127,244}{\scriptsize 52.4} & \cellcolor[RGB]{143,176,248}{\scriptsize 19.9\scalebox{0.55}{$\pm$17}} \\
& {\scriptsize Qwen3-14B} & \cellcolor[RGB]{247,249,254}{\scriptsize 0.1} & \cellcolor[RGB]{212,224,252}{\scriptsize 2.9} & \cellcolor[RGB]{154,184,249}{\scriptsize 16.1} & \cellcolor[RGB]{134,169,247}{\scriptsize 23.3} & \cellcolor[RGB]{130,166,247}{\scriptsize 24.9} & \cellcolor[RGB]{133,169,247}{\scriptsize 23.6} & \cellcolor[RGB]{232,239,253}{\scriptsize 0.8} & \cellcolor[RGB]{204,219,252}{\scriptsize 4.0} & \cellcolor[RGB]{202,218,251}{\scriptsize 4.3} & \cellcolor[RGB]{113,155,246}{\scriptsize 32.0} & \cellcolor[RGB]{215,227,252}{\scriptsize 2.4} & \cellcolor[RGB]{145,177,248}{\scriptsize 19.1} & \cellcolor[RGB]{65,121,243}{\scriptsize 57.1} & \cellcolor[RGB]{65,121,243}{\scriptsize 57.1} & \cellcolor[RGB]{77,129,244}{\scriptsize 50.6} & \cellcolor[RGB]{53,112,243}{\scriptsize 65.1} & \cellcolor[RGB]{132,168,247}{\scriptsize 24.0\scalebox{0.55}{$\pm$22}} \\
& {\scriptsize Qwen3-32B} & \cellcolor[RGB]{247,249,254}{\scriptsize 0.1} & \cellcolor[RGB]{210,223,252}{\scriptsize 3.1} & \cellcolor[RGB]{154,183,249}{\scriptsize 16.2} & \cellcolor[RGB]{134,169,247}{\scriptsize 23.3} & \cellcolor[RGB]{128,165,247}{\scriptsize 25.7} & \cellcolor[RGB]{130,166,247}{\scriptsize 24.8} & \cellcolor[RGB]{231,238,253}{\scriptsize 0.9} & \cellcolor[RGB]{205,220,252}{\scriptsize 3.9} & \cellcolor[RGB]{202,217,251}{\scriptsize 4.4} & \cellcolor[RGB]{114,156,246}{\scriptsize 31.3} & \cellcolor[RGB]{212,224,252}{\scriptsize 2.9} & \cellcolor[RGB]{145,177,248}{\scriptsize 19.3} & \cellcolor[RGB]{65,121,243}{\scriptsize 57.2} & \cellcolor[RGB]{54,113,243}{\scriptsize 64.1} & \cellcolor[RGB]{75,128,244}{\scriptsize 51.7} & \cellcolor[RGB]{45,106,242}{\scriptsize 70.2} & \cellcolor[RGB]{129,166,247}{\scriptsize 25.0\scalebox{0.55}{$\pm$23}} \\
& {\scriptsize Gemma-3-4B} & \cellcolor[RGB]{247,249,254}{\scriptsize 0.1} & \cellcolor[RGB]{218,229,252}{\scriptsize 2.1} & \cellcolor[RGB]{175,198,250}{\scriptsize 10.2} & \cellcolor[RGB]{163,190,249}{\scriptsize 13.5} & \cellcolor[RGB]{163,190,249}{\scriptsize 13.3} & \cellcolor[RGB]{172,196,250}{\scriptsize 10.8} & \cellcolor[RGB]{235,241,253}{\scriptsize 0.6} & \cellcolor[RGB]{212,224,252}{\scriptsize 2.9} & \cellcolor[RGB]{221,231,253}{\scriptsize 1.8} & \cellcolor[RGB]{149,180,248}{\scriptsize 17.8} & \cellcolor[RGB]{225,234,253}{\scriptsize 1.4} & \cellcolor[RGB]{169,194,249}{\scriptsize 11.7} & \cellcolor[RGB]{119,159,247}{\scriptsize 29.5} & \cellcolor[RGB]{150,181,248}{\scriptsize 17.4} & \cellcolor[RGB]{149,180,248}{\scriptsize 17.8} & \cellcolor[RGB]{143,176,248}{\scriptsize 19.8} & \cellcolor[RGB]{173,197,250}{\scriptsize 10.7\scalebox{0.55}{$\pm$8}} \\
& {\scriptsize Gemma-3-12B} & \cellcolor[RGB]{249,251,254}{\scriptsize 0.0} & \cellcolor[RGB]{216,227,252}{\scriptsize 2.4} & \cellcolor[RGB]{163,190,249}{\scriptsize 13.3} & \cellcolor[RGB]{147,179,248}{\scriptsize 18.4} & \cellcolor[RGB]{144,177,248}{\scriptsize 19.4} & \cellcolor[RGB]{148,180,248}{\scriptsize 18.0} & \cellcolor[RGB]{233,239,253}{\scriptsize 0.7} & \cellcolor[RGB]{210,223,252}{\scriptsize 3.2} & \cellcolor[RGB]{210,223,252}{\scriptsize 3.1} & \cellcolor[RGB]{131,167,247}{\scriptsize 24.5} & \cellcolor[RGB]{217,228,252}{\scriptsize 2.3} & \cellcolor[RGB]{156,185,249}{\scriptsize 15.5} & \cellcolor[RGB]{86,136,245}{\scriptsize 45.3} & \cellcolor[RGB]{100,146,245}{\scriptsize 38.0} & \cellcolor[RGB]{111,153,246}{\scriptsize 32.7} & \cellcolor[RGB]{88,137,245}{\scriptsize 44.3} & \cellcolor[RGB]{150,180,248}{\scriptsize 17.6\scalebox{0.55}{$\pm$15}} \\
& {\scriptsize Gemma-3-27B} & \cellcolor[RGB]{247,249,254}{\scriptsize 0.1} & \cellcolor[RGB]{213,225,252}{\scriptsize 2.8} & \cellcolor[RGB]{155,184,249}{\scriptsize 15.9} & \cellcolor[RGB]{138,173,248}{\scriptsize 21.5} & \cellcolor[RGB]{132,168,247}{\scriptsize 24.1} & \cellcolor[RGB]{134,170,247}{\scriptsize 23.1} & \cellcolor[RGB]{229,236,253}{\scriptsize 1.1} & \cellcolor[RGB]{205,220,252}{\scriptsize 3.8} & \cellcolor[RGB]{202,218,251}{\scriptsize 4.3} & \cellcolor[RGB]{118,158,246}{\scriptsize 29.8} & \cellcolor[RGB]{215,227,252}{\scriptsize 2.4} & \cellcolor[RGB]{149,180,248}{\scriptsize 17.6} & \cellcolor[RGB]{70,124,244}{\scriptsize 54.5} & \cellcolor[RGB]{67,122,243}{\scriptsize 56.3} & \cellcolor[RGB]{84,134,244}{\scriptsize 46.3} & \cellcolor[RGB]{53,112,243}{\scriptsize 64.8} & \cellcolor[RGB]{134,170,247}{\scriptsize 23.0\scalebox{0.55}{$\pm$21}} \\
& {\scriptsize Phi-4} & \cellcolor[RGB]{249,251,254}{\scriptsize 0.0} & \cellcolor[RGB]{212,224,252}{\scriptsize 2.9} & \cellcolor[RGB]{151,181,248}{\scriptsize 17.1} & \cellcolor[RGB]{134,169,247}{\scriptsize 23.3} & \cellcolor[RGB]{125,163,247}{\scriptsize 26.7} & \cellcolor[RGB]{131,167,247}{\scriptsize 24.5} & \cellcolor[RGB]{230,237,253}{\scriptsize 0.9} & \cellcolor[RGB]{203,218,251}{\scriptsize 4.2} & \cellcolor[RGB]{201,217,251}{\scriptsize 4.5} & \cellcolor[RGB]{111,153,246}{\scriptsize 32.8} & \cellcolor[RGB]{214,226,252}{\scriptsize 2.6} & \cellcolor[RGB]{143,176,248}{\scriptsize 19.9} & \cellcolor[RGB]{63,120,243}{\scriptsize 58.3} & \cellcolor[RGB]{53,112,243}{\scriptsize 65.0} & \cellcolor[RGB]{74,127,244}{\scriptsize 52.3} & \cellcolor[RGB]{41,104,242}{\scriptsize 72.8} & \cellcolor[RGB]{128,165,247}{\scriptsize 25.5\scalebox{0.55}{$\pm$24}} \\
\cmidrule[0.3pt](l){2-19}
& {\scriptsize Dream-7B} & \cellcolor[RGB]{249,251,254}{\scriptsize 0.0} & \cellcolor[RGB]{221,231,253}{\scriptsize 1.8} & \cellcolor[RGB]{186,206,250}{\scriptsize 7.5} & \cellcolor[RGB]{168,194,249}{\scriptsize 11.9} & \cellcolor[RGB]{173,197,250}{\scriptsize 10.6} & \cellcolor[RGB]{174,198,250}{\scriptsize 10.4} & \cellcolor[RGB]{238,243,254}{\scriptsize 0.4} & \cellcolor[RGB]{217,228,252}{\scriptsize 2.2} & \cellcolor[RGB]{221,231,253}{\scriptsize 1.8} & \cellcolor[RGB]{164,190,249}{\scriptsize 13.1} & \cellcolor[RGB]{228,236,253}{\scriptsize 1.1} & \cellcolor[RGB]{179,201,250}{\scriptsize 9.1} & \cellcolor[RGB]{135,170,247}{\scriptsize 22.7} & \cellcolor[RGB]{148,179,248}{\scriptsize 18.2} & \cellcolor[RGB]{144,177,248}{\scriptsize 19.4} & \cellcolor[RGB]{115,156,246}{\scriptsize 31.3} & \cellcolor[RGB]{175,198,250}{\scriptsize 10.1\scalebox{0.55}{$\pm$9}} \\
& {\scriptsize LLaDA-8B} & \cellcolor[RGB]{249,251,254}{\scriptsize 0.0} & \cellcolor[RGB]{217,228,252}{\scriptsize 2.3} & \cellcolor[RGB]{166,192,249}{\scriptsize 12.4} & \cellcolor[RGB]{151,181,248}{\scriptsize 17.1} & \cellcolor[RGB]{150,180,248}{\scriptsize 17.6} & \cellcolor[RGB]{155,184,249}{\scriptsize 15.9} & \cellcolor[RGB]{237,242,253}{\scriptsize 0.5} & \cellcolor[RGB]{210,223,252}{\scriptsize 3.2} & \cellcolor[RGB]{212,225,252}{\scriptsize 2.8} & \cellcolor[RGB]{138,173,248}{\scriptsize 21.5} & \cellcolor[RGB]{215,227,252}{\scriptsize 2.4} & \cellcolor[RGB]{157,186,249}{\scriptsize 15.1} & \cellcolor[RGB]{98,144,245}{\scriptsize 38.9} & \cellcolor[RGB]{101,146,245}{\scriptsize 37.7} & \cellcolor[RGB]{112,154,246}{\scriptsize 32.2} & \cellcolor[RGB]{97,144,245}{\scriptsize 39.4} & \cellcolor[RGB]{154,183,249}{\scriptsize 16.2\scalebox{0.55}{$\pm$14}} \\
\cmidrule[0.3pt](l){2-19}
& {\scriptsize C-Haiku-4.5} & \cellcolor[RGB]{249,251,254}{\scriptsize 0.0} & \cellcolor[RGB]{211,224,252}{\scriptsize 3.1} & \cellcolor[RGB]{153,183,249}{\scriptsize 16.5} & \cellcolor[RGB]{134,169,247}{\scriptsize 23.3} & \cellcolor[RGB]{125,163,247}{\scriptsize 26.7} & \cellcolor[RGB]{128,165,247}{\scriptsize 25.7} & \cellcolor[RGB]{231,238,253}{\scriptsize 0.9} & \cellcolor[RGB]{206,220,252}{\scriptsize 3.8} & \cellcolor[RGB]{201,217,251}{\scriptsize 4.6} & \cellcolor[RGB]{111,153,246}{\scriptsize 33.0} & \cellcolor[RGB]{211,224,252}{\scriptsize 3.0} & \cellcolor[RGB]{145,177,248}{\scriptsize 19.0} & \cellcolor[RGB]{61,118,243}{\scriptsize 59.6} & \cellcolor[RGB]{38,102,242}{\scriptsize 79.1} & \cellcolor[RGB]{67,122,243}{\scriptsize 56.3} & \cellcolor[RGB]{38,102,242}{\scriptsize 82.5} & \cellcolor[RGB]{124,162,247}{\scriptsize 27.3\scalebox{0.55}{$\pm$27}} \\
& {\scriptsize GPT-5} & \cellcolor[RGB]{247,249,254}{\scriptsize 0.1} & \cellcolor[RGB]{209,222,252}{\scriptsize 3.3} & \cellcolor[RGB]{150,180,248}{\scriptsize 17.6} & \cellcolor[RGB]{130,166,247}{\scriptsize 24.9} & \cellcolor[RGB]{122,161,247}{\scriptsize 28.1} & \cellcolor[RGB]{132,168,247}{\scriptsize 24.0} & \cellcolor[RGB]{231,238,253}{\scriptsize 0.9} & \cellcolor[RGB]{201,217,251}{\scriptsize 4.6} & \cellcolor[RGB]{202,217,251}{\scriptsize 4.4} & \cellcolor[RGB]{107,150,246}{\scriptsize 34.7} & \cellcolor[RGB]{212,224,252}{\scriptsize 2.9} & \cellcolor[RGB]{137,172,248}{\scriptsize 21.9} & \cellcolor[RGB]{53,112,243}{\scriptsize 64.7} & \cellcolor[RGB]{38,102,242}{\scriptsize 91.3} & \cellcolor[RGB]{54,113,243}{\scriptsize 64.5} & \cellcolor[RGB]{38,102,242}{\scriptsize 88.5} & \cellcolor[RGB]{118,158,246}{\scriptsize 29.8\scalebox{0.55}{$\pm$30}} \\
& {\scriptsize G-3-Flash} & \cellcolor[RGB]{247,249,254}{\scriptsize 0.1} & \cellcolor[RGB]{209,222,252}{\scriptsize 3.3} & \cellcolor[RGB]{152,182,248}{\scriptsize 16.8} & \cellcolor[RGB]{131,167,247}{\scriptsize 24.4} & \cellcolor[RGB]{124,162,247}{\scriptsize 27.2} & \cellcolor[RGB]{128,165,247}{\scriptsize 25.6} & \cellcolor[RGB]{230,237,253}{\scriptsize 1.0} & \cellcolor[RGB]{205,220,252}{\scriptsize 3.9} & \cellcolor[RGB]{202,218,251}{\scriptsize 4.4} & \cellcolor[RGB]{108,151,246}{\scriptsize 34.2} & \cellcolor[RGB]{212,224,252}{\scriptsize 2.9} & \cellcolor[RGB]{141,175,248}{\scriptsize 20.4} & \cellcolor[RGB]{57,115,243}{\scriptsize 62.5} & \cellcolor[RGB]{38,102,242}{\scriptsize 89.3} & \cellcolor[RGB]{60,117,243}{\scriptsize 60.6} & \cellcolor[RGB]{38,102,242}{\scriptsize 88.9} & \cellcolor[RGB]{120,159,247}{\scriptsize 29.1\scalebox{0.55}{$\pm$29}} \\
& {\scriptsize G-3.1-Pro} & \cellcolor[RGB]{249,251,254}{\scriptsize 0.0} & \cellcolor[RGB]{212,225,252}{\scriptsize 2.8} & \cellcolor[RGB]{152,182,248}{\scriptsize 16.7} & \cellcolor[RGB]{132,168,247}{\scriptsize 24.1} & \cellcolor[RGB]{124,163,247}{\scriptsize 27.1} & \cellcolor[RGB]{128,165,247}{\scriptsize 25.6} & \cellcolor[RGB]{232,239,253}{\scriptsize 0.8} & \cellcolor[RGB]{204,219,252}{\scriptsize 4.1} & \cellcolor[RGB]{202,217,251}{\scriptsize 4.4} & \cellcolor[RGB]{105,149,246}{\scriptsize 35.5} & \cellcolor[RGB]{211,224,252}{\scriptsize 3.0} & \cellcolor[RGB]{140,174,248}{\scriptsize 20.8} & \cellcolor[RGB]{56,114,243}{\scriptsize 62.9} & \cellcolor[RGB]{38,102,242}{\scriptsize 92.9} & \cellcolor[RGB]{57,115,243}{\scriptsize 62.4} & \cellcolor[RGB]{38,102,242}{\scriptsize 88.7} & \cellcolor[RGB]{119,159,247}{\scriptsize 29.5\scalebox{0.55}{$\pm$30}} \\
\cmidrule[0.5pt](l){2-19}
& {\scriptsize Gen.\ quality} & \cellcolor[RGB]{248,250,254}{\scriptsize 0.1\scalebox{0.55}{$\pm$0}} & \cellcolor[RGB]{214,226,252}{\scriptsize 2.6\scalebox{0.55}{$\pm$1}} & \cellcolor[RGB]{163,190,249}{\scriptsize 13.4\scalebox{0.55}{$\pm$4}} & \cellcolor[RGB]{145,177,248}{\scriptsize 19.0\scalebox{0.55}{$\pm$6}} & \cellcolor[RGB]{141,175,248}{\scriptsize 20.5\scalebox{0.55}{$\pm$7}} & \cellcolor[RGB]{146,178,248}{\scriptsize 18.8\scalebox{0.55}{$\pm$7}} & \cellcolor[RGB]{233,239,253}{\scriptsize 0.7\scalebox{0.55}{$\pm$0}} & \cellcolor[RGB]{209,222,252}{\scriptsize 3.3\scalebox{0.55}{$\pm$1}} & \cellcolor[RGB]{209,222,252}{\scriptsize 3.3\scalebox{0.55}{$\pm$1}} & \cellcolor[RGB]{127,165,247}{\scriptsize 25.8\scalebox{0.55}{$\pm$9}} & \cellcolor[RGB]{217,228,252}{\scriptsize 2.2\scalebox{0.55}{$\pm$1}} & \cellcolor[RGB]{155,184,249}{\scriptsize 15.9\scalebox{0.55}{$\pm$5}} & \cellcolor[RGB]{84,134,244}{\scriptsize 46.5\scalebox{0.55}{$\pm$17}} & \cellcolor[RGB]{77,129,244}{\scriptsize 50.2\scalebox{0.55}{$\pm$28}} & \cellcolor[RGB]{97,143,245}{\scriptsize 39.6\scalebox{0.55}{$\pm$19}} & \cellcolor[RGB]{70,124,244}{\scriptsize 54.4\scalebox{0.55}{$\pm$27}} &  \\
\bottomrule
\end{tabular}
}%  end resizebox
\end{table}

% ─────────────────────────────────────────────────────────────────────────────
\section{Experimental Results}
\label{sec:results}
% ─────────────────────────────────────────────────────────────────────────────

\subsection{Setup}
\label{sec:setup}

   We evaluate $N=16$ models across three groups. Ten open-weight
   auto-regressive: Qwen3~\citep{qwen3} (0.6B, 1.7B, 4B, 8B, 14B, 32B),
   Gemma-3~\citep{gemma3} (4B, 12B, 27B), and Phi-4~\citep{phi4}. Two
   diffusion LMs, Dream-v0-7B~\citep{dream} and LLaDA-8B~\citep{llada},
   to test whether parallel denoising yields a different serialization
   profile. Four frontier models via API: \fmtwo~\citep{claude_haiku_4_5},
   GPT-5~\citep{singh2025openai}, Gemini-3-Flash~\citep{gemini3flash}, and
   Gemini-3.1-Pro~\citep{gemini3pro}.
   All use greedy decoding (temperature~0, thinking disabled), \Cref{app:temperature} confirms
   $\mathrm{temp}=0$ yields the best accuracy. The expression suite
   contains $|D|=2450$ expressions, 5 instantiations per skeleton with
   up to 50 skeletons per $(k,d)$ cell, for $k \in [2,8],\; d \in [2,6]$, with uniform random subsample
   with fixed seed when a cell exceeds 50
   (not all pairs are feasible, see \Cref{app:expr_grid}). Guard pass rates
   are reported alongside all results in \Cref{tab:matrix}.

% \textbf{Shot configuration.} We ablate
% $(\gamma, \varepsilon) \in \{0, 3\}^2$ and report \Cref{tab:matrix}
% at $(\gamma, \varepsilon) = (0, 3)$, the configuration maximizing the
% panel-mean generation and extraction quality trade-off
% on the open-weight panel: generator-side shots ($\gamma{=}3$) do not improve
% generation quality for the strongest auto-regressive models (Qwen3 above 4B, Phi-4), while extractor-side shots uniformly improve
% generation quality over $\varepsilon{=}0$. Frontier models were evaluated only at $\gamma{=}0$,
% so we cannot characterize their sensitivity to generator shots. The
% full per-condition grids are in \Cref{app:shot_grid}.

   \textbf{Shot configuration.} We ablate
   $(\gamma, \varepsilon) \in \{0, 3\}^2$ and report \Cref{tab:matrix}
   at the trade-off optimum $(0, 3)$: generator shots ($\gamma{=}3$) do
   not help the strongest auto-regressive models (Qwen3 $\geq$4B, Phi-4),
   while extractor shots uniformly improve extraction over $\varepsilon{=}0$.
   Frontier models were evaluated only at $\gamma{=}0$ for tractability. Full per-condition
   grids are in \Cref{app:shot_grid}.

\subsection{Characterizing the Natural-Language Channel}
\label{sec:comm_matrix_results}

The full communication matrix (\Cref{tab:matrix}; per-depth
breakdowns in \Cref{app:depth_matrices}) reveals a channel whose
capacity rises with model capability and depends on how models are
paired across roles. Across the 12 open-weight models, mean
round-trip accuracy ($q_{ij}$, \Cref{eq:accuracy}) is $9.4\%$,
rising to $17.0\%$ at $\geq$8B.
Frontier models raise the ceiling: the $4 \times 4$ frontier
submatrix averages $74.7\%$, and the best cell reaches $92.9\%$
(Gemini-3.1-Pro reading GPT-5), above the best self-communication score
(max $91.3\%$, GPT-5) and $4.2$~pp above Gemini-3.1-Pro reading from itself. The best cell is itself a role-asymmetry result:
the strongest channel pairs the strongest generator with the
strongest extractor, not the same model on both sides. Accuracy
still degrades with complexity at every tier. Frontier generators
drop from $88\%$ at $k{=}3$ to $71\%$ at $k{=}8$, and open-weight
$\geq$8B generators from $64\%$ at $k{=}2$ to $13\%$ at $k{=}8$.
Diffusion parallel denoising does not escape the bottleneck. LLaDA-8B (gen/extr
$15.9 / 16.2$) and Dream-v0-7B ($2.2 / 10.1$) both sit within the
autoregressive envelope.

\begin{figure}[htb]
  \centering
  \includegraphics[width=1\linewidth]{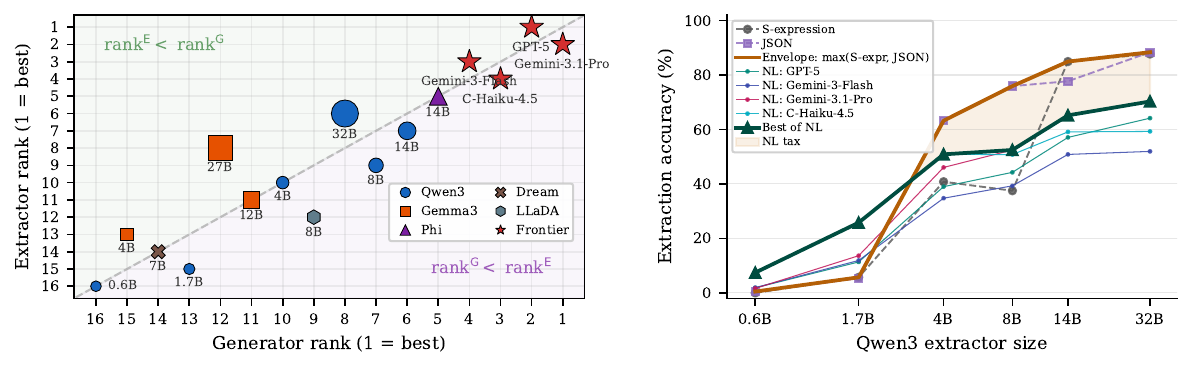}
  % \vspace{-2mm}
  \caption{\textbf{(Left)~Role asymmetry.} Generator rank vs.\ extractor rank (1 = best). Gemini-3.1-Pro leads generation but is second as extractor, GPT-5 is the top extractor but second generator. Swapping roles shifts round-trip accuracy by up to 60.4~pp.
    \textbf{(Right)~Channel asymmetry.} Qwen3 extraction on JSON/S-expressions (structured, unambiguous) vs.\ Frontier-generated word problems. The shaded area is the \emph{NL tax}: capacity large extractors have but no NL generator in our panel reaches.}
  \label{fig:broadcaster}
\end{figure}

\textbf{Generation and extraction are distinct skills.}
$\genqual{}$ and $\extqual{}$ are not directly comparable in absolute
terms because each marginal is upper-bounded by the panel's quality
on the opposite role, but within-cohort ranks are comparable. {\Cref{fig:broadcaster}-left}
plots each model's rank as $\Gen$ against its rank as $\Ext$. Had
both roles reflected one ability, models would sit on the diagonal, but they disperse instead. GPT-5 is the top extractor
($\extqual{}=29.8\%$, rank 1) yet ranks second as generator
($\genqual{}=50.2\%$); Gemini-3.1-Pro leads generation
($\genqual{}=54.4\%$, rank 1) but is only the second-best extractor
($\extqual{}=29.5\%$). The dispersion is largest for Gemma-3-27B
(rank 11 generator at $3.3\%$, rank 8 extractor at $23.0\%$): a
model whose narration loses structure more than its parsing does.

\textbf{Role assignment matters.} Swapping which model generates
and which extracts shifts accuracy by up to $60.4$~pp
(Gemma-3-27B with Gemini-3.1-Pro, larger than any row or column spread
in \Cref{tab:matrix}), where Gemma's stronger
extraction (rank 8) and weak generation (rank 11) make role
assignment decisive. The effect also appears within the frontier
panel: GPT-5 generating with Gemini-3.1-Pro
extracting yields $92.9\%$, the reverse $88.5\%$ ($4.4$~pp).
Pairing the strongest model on both sides is suboptimal when
the panel has a model with complementary strengths.

   \textbf{The NL channel is quantifiably lossy.}
   To separate channel loss from extractor weakness, we need a reference
   point where the encoding is lossless. We implement it by replacing
   the generator with deterministic S-expression and JSON-AST
   serializations and measure the $\text{NL tax} = \max(\extqual{\text{S-expr}}, \extqual{\text{JSON}}) - \extqual{\text{NL}}$,
   the extraction gap between a lossless encoding and natural language.
   {\Cref{fig:broadcaster}-right} shows this for the Qwen3
   family: at 14B--32B, lossless extraction reaches $85$--$88\%$ while
   NL extraction from Frontier models plateaus at $51$--$70\%$, an $18$--$20$~pp tax
   that grows with complexity (\Cref{app:sexpr}). The gap is relative to
   the panel's best generator and would narrow with a stronger one.
   Below 4B, models lack fluency in structured notation and extract
   similarly from prose, so the structured-format advantage requires
   extractor familiarity.

\subsection{Why Communication Fails}
\label{sec:why_fails}

\begin{figure}[htb]
  \centering
  \begin{minipage}[c]{0.27\linewidth}
    \centering
    \vspace{-7mm}

    \scriptsize
    \setlength{\tabcolsep}{3pt}
    \begin{tabular}{@{}l r@{}}
      \toprule
      Feature & Attribution \\
      \midrule
      Generator size  & 39.1\% $\uparrow$ \\
      $R_{\text{math}}$ & 19.5\% $\uparrow$ \\
      Extractor size  & 17.9\% $\uparrow$ \\
      Depth           & 9.2\% $\downarrow$ \\
      OpCount         & 7.8\% $\downarrow$ \\
      $\ell(T)$       & 4.6\% $\downarrow$ \\
      Family match    & 1.9\% $\uparrow$ \\
      \bottomrule
    \end{tabular}
  \end{minipage}%
  \hfill
  \begin{minipage}[c]{0.7\linewidth}
    \centering
    \includegraphics[width=\linewidth]{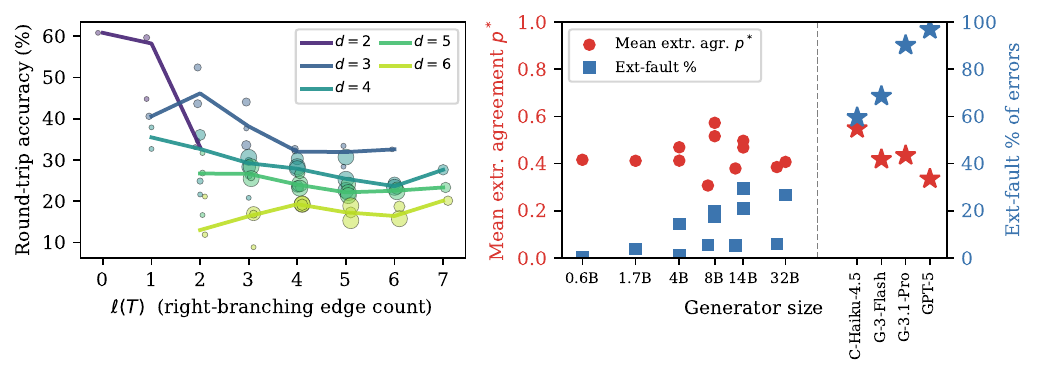}
  \end{minipage}
  \vspace{-2mm}
  \caption{\textbf{Why communication fails.} Over depths $2$--$6$.
  \textbf{(Left)}~Global SHAP attributions, computed on the 12
  open-weight models only (frontier parameter counts are unknown and
  would confound the size features): generator size $39.1\%$,
  $R_{\text{math}}$ $19.5\%$, and extractor size $17.9\%$ dominate.
  \textbf{(Center)}~Mean within-depth Spearman correlation between
  linearization complexity $\ell(T)$ and accuracy is
  $\overline{\rho_{|d}} = -0.48$: right-branching skeletons are harder
  at matched depth.
  \textbf{(Right)}~Extractor-fault rate rises with generator size,
  computed on all 16 models (12 open-weight plus Haiku, GPT-5 and the
  two Geminis), from ${<}4\%$ for the smallest open-weight generators
  to $60$--$97\%$ for the frontier models.}
  \label{fig:why_fails_new}
\end{figure}

\textbf{What predicts failure?}
We train a gradient-boosted classifier on all $(\expr, \Gen, \Ext)$
triples from the 12 open-weight models and compute SHAP
attributions~\citep{lundberg2017unified,lundberg2019explainable}
(\Cref{fig:why_fails_new}, left; details in \Cref{app:shap}). Frontier
models are excluded because their unknown parameter counts would
confound the model-size features. Model size dominates: generator
and extractor sizes together account for $57.0\%$ of the
attribution. $R_{\text{math}}$ ($19.5\%$, $\uparrow$), the fraction
of words in the word problem belonging to a curated arithmetic
lexicon (\Cref{app:rmath}), ranks second overall, ahead of
extractor size, and is comparable in magnitude to the combined
structural features ($21.6\%$),
indicating that pseudo-mathematical phrasing produces more
decodable problems. Among structural features, depth
($9.2\%$), operator count ($7.8\%$), and \emph{linearization complexity}
$\ell(T)$ ($4.6\%$) contribute in that order. We define $\ell(T)$ as
the number of internal nodes of $T$ whose right child is itself
internal, counted tree-wide ($\ell{=}0$ for a left-caterpillar of
any depth, $\ell{=}d{-}1$ for a right-caterpillar of depth~$d$),
reminiscent of the minimum register count in expression
evaluation~\citep{Sethi1970TheGO}. Controlling for depth, $\ell(T)$
predicts difficulty monotonically, with a mean within-depth
Spearman correlation of $\overline{\rho_{|d}}=-0.48$
(\Cref{fig:why_fails_new}, center). Right-branching skeletons are
therefore harder than left-branching ones of matched size. Per-cell
breakdowns appear in \Cref{app:com}. Family match contributes
$1.9\%$, providing no evidence of architecture-specific dialects.

\textbf{Where does failure occur?}
\label{sec:fault_results}
Applying the consensus-based attribution (\Cref{sec:fault_method}, with the full breakdown in \Cref{app:fault_attribution})
to all failed round trips, we find that at least $73.6\%$ are
generator faults (no extractor recovers the target) and at most
$26.4\%$ are extractor faults. Wrong operator count is the
dominant error type, accounting for $71.8\%$ of all errors
(\Cref{app:error_taxonomy}, with qualitative failure traces in \Cref{app:failure_examples}). The extractor-fault rate scales
steeply with generator capability, from under $4\%$ for sub-2B
open-weight generators to $60$--$97\%$ for frontier models
(\Cref{fig:why_fails_new}, right). The $73.6\%$ figure is robust to
panel composition: across nine extractor-panel subsets the rate
ranges $73.6$--$91.1\%$, with the full 16-model panel yielding
the lowest observation (\Cref{app:fault_sensitivity}). No panel
we tested reports a generator-fault share below $73.6\%$.

% ─────────────────────────────────────────────────────────────────────────────
\subsection{External Validity}
\label{sec:scaling_ext}

% A natural question is whether round-trip scores simply recapitulate
% existing benchmarks or capture a distinct capability. We correlate
% per-model round-trip scores with published results on 9 benchmarks
% across four categories, restricting to benchmarks with verified
% scores for at least seven models. Extraction quality correlates
% strongly with general-knowledge (MMLU $\rho = 0.98$), code
% (MBPP $\rho = 0.93$, LiveCodeBench $\rho = 0.93$), and reasoning benchmarks
% (GPQA-Diamond $\rho = 0.95$), but generation quality correlates
% only moderately with the same benchmarks ($\rho = 0.47$--$0.89$).
% This generation--extraction gap is consistent across all four
% categories and confirms that faithfully serializing compositional
% structure into prose is a partially distinct capability from the
% comprehension skills that standard benchmarks test. Details in
% \Cref{app:correlation}.

   Do round-trip scores recapitulate existing benchmarks or capture a
   distinct capability? We correlate per-model round-trip scores with
   published results on 9 benchmarks across four categories (at least
   seven verified scores each). Extraction correlates strongly with
   general knowledge (MMLU $\rho = 0.98$), code (MBPP $\rho = 0.93$,
   LiveCodeBench $\rho = 0.93$), and reasoning (GPQA-Diamond $\rho = 0.95$),
   but generation correlates only moderately ($\rho = 0.47$--$0.89$).
   This gap is consistent across all four categories, confirming that
   serializing compositional structure into prose is partially distinct
   from the comprehension skills standard benchmarks test. Details in
   \Cref{app:correlation}.

\textbf{Cross-domain replication.}
We re-run the protocol on propositional logic on a reduced
$14 \times 14$ roster (five connectives, depths 2--5, logical equivalence as oracle; frontier Gemini omitted to limit compute; \Cref{app:logic}). Absolute accuracy drops (mean cell $8.7\%$ vs.\
$14.9\%$ on the shared 14-model block), but model rankings mostly carry over: Spearman
$\rho = 0.85$ for generation and $0.69$ for extraction, with GPT-5
and Claude Haiku 4.5 again on top. The main exception is Phi-4,
which falls from extraction rank 3 to 14, suggesting its arithmetic
extraction relies on format-specific parsing. Rankings are stable overall but not uniform, reinforcing a core takeaway: the best model and role
assignment depend on the domain, not just on overall capability.

\section{Improving Communication Through Fine-Tuning}
\label{sec:finetuning}

Generation is the bottleneck: at least 73.6\% of round-trip failures
originate there (\Cref{sec:why_fails}). If the bottleneck is
trainable rather than architectural, what kind of training lifts
it and how far does it transfer? Two fine-tuning regimes test this,
both sharing the round-trip format.

\textbf{Arithmetic fine-tuning} probes how far structural fluency
carries across unseen variables and constants. Training pairs
($\expr \!\to\! \wpr$ and $\wpr \!\to\! \expr$) use the
$\{+,-,\times,\div\}$ operators and the evaluation tree topology,
with disjoint letters $\{Q\ldots Z\}$ and constants $[11\ldots 20]$.

\textbf{Assembly fine-tuning} probes whether the linearization
skill transfers across domains. Assembly has mixed-arity
operators (three unary, six binary, four ternary), breaking the
binary-only isomorphism, and coloured-object leaves
(13 colours $\times$ 13 objects) disjoint from any arithmetic
vocabulary. Each operator maps to multiple paraphrase templates
in a NL procedural style. Every (expression, NL) pair
yields two training examples ($\expr \!\to\! \wpr$ and
$\wpr \!\to\! \expr$). Full specification in
\Cref{app:finetuning_assembly}.

\textbf{Training and evaluation.} Both regimes use QLoRA~\citep{qlora}
(rank 32, $\alpha{=}64$, 4-bit NF4, 3 epochs, AdamW~\citep{adamw},
lr $2{\times}10^{-4}$, cosine schedule) on $\sim\!3600$ training
messages across the 10 open-weight autoregressive models. Adapters
are merged before evaluation.\footnote{A single hyperparameter
setting is used across models. Per-model tuning is orthogonal to
the structural-transfer question.}
We report each condition at its strongest shot configuration on the
full grid (\Cref{app:shot_grid}): base and arithmetic FT use
$\promptg{0}, \prompte{3}$, assembly FT uses $\promptg{3}, \prompte{3}$
because assembly-FT generators benefit from three exemplars but
base and arithmetic-FT generators do not.

\subsection{Fine-tuning Results}
\label{sec:ft_results}

% \Cref{tab:ft_assembly_summary} compares base, assembly-fine-tuned, and
% arithmetic-fine-tuned models.

\begin{table}[htb]
\centering
\caption{\textbf{Fine-tuning comparison across 10 open-weight models.} Three conditions are evaluated at their strongest shot configuration (\Cref{app:shot_grid}): Base $(\gamma{=}0, \varepsilon{=}3)$, assembly FT $(\gamma{=}3, \varepsilon{=}3)$, arithmetic FT $(\gamma{=}0, \varepsilon{=}3)$. $\Guard$ failures count as errors; all marginals are averaged over the 10-model extractor panel. \textbf{Bold} = best per row; \underline{underline} = second-best.}
% \vspace{2mm}
\label{tab:ft_assembly_summary}
\small
\setlength{\tabcolsep}{4pt}
\resizebox{\textwidth}{!}{%
\begin{tabular}{l ccc ccc ccc ccc}
\toprule
 & \multicolumn{3}{c}{$\Guard$ pass \%} & \multicolumn{3}{c}{$\genqual{}$} & \multicolumn{3}{c}{$\extqual{}$} & \multicolumn{3}{c}{Self} \\
\cmidrule(lr){2-4}\cmidrule(lr){5-7}\cmidrule(lr){8-10}\cmidrule(lr){11-13}
Model & Base & FT$_{\text{asm}}$ & FT$_{\text{arith}}$ & Base & FT$_{\text{asm}}$ & FT$_{\text{arith}}$ & Base & FT$_{\text{asm}}$ & FT$_{\text{arith}}$ & Base & FT$_{\text{asm}}$ & FT$_{\text{arith}}$ \\
\midrule
Qwen3-0.6B & 20 & \underline{52} & \textbf{100} & 0.1 & \underline{0.1} & \textbf{51.9} & 3.0 & \underline{3.7} & \textbf{14.1} & \underline{0.1} & 0.0 & \textbf{61.8} \\
Qwen3-1.7B & 59 & \underline{66} & \textbf{100} & 2.6 & \underline{6.5} & \textbf{57.2} & 6.6 & \underline{8.4} & \textbf{34.8} & 2.1 & \underline{3.8} & \textbf{70.4} \\
Qwen3-4B & \underline{94} & 82 & \textbf{100} & 14.0 & \underline{17.3} & \textbf{55.3} & 12.4 & \underline{20.9} & \textbf{64.4} & 15.3 & \underline{21.5} & \textbf{79.4} \\
Qwen3-8B & \underline{98} & 94 & \textbf{100} & 20.5 & \underline{32.6} & \textbf{59.5} & 12.9 & \underline{21.1} & \textbf{66.2} & 22.7 & \underline{37.3} & \textbf{83.0} \\
Qwen3-14B & \underline{99} & 98 & \textbf{100} & 21.4 & \underline{26.2} & \textbf{55.9} & 14.5 & \underline{24.2} & \textbf{72.7} & 27.3 & \underline{36.7} & \textbf{82.1} \\
Qwen3-32B & \underline{98} & 98 & \textbf{100} & 19.4 & \underline{27.6} & \textbf{58.0} & 14.7 & \underline{25.3} & \textbf{73.6} & 26.5 & \underline{41.6} & \textbf{85.6} \\
Gemma-3-4B & \underline{98} & 75 & \textbf{100} & 0.8 & \underline{5.5} & \textbf{55.4} & 8.4 & \underline{11.9} & \textbf{37.1} & 0.6 & \underline{4.6} & \textbf{48.4} \\
Gemma-3-12B & \underline{96} & 79 & \textbf{100} & 3.9 & \underline{16.7} & \textbf{57.8} & 11.7 & \underline{19.2} & \textbf{59.8} & 4.2 & \underline{21.1} & \textbf{81.2} \\
Gemma-3-27B & \underline{96} & 93 & \textbf{100} & 3.8 & \underline{20.4} & \textbf{58.4} & 14.2 & \underline{24.2} & \textbf{70.6} & 5.1 & \underline{27.2} & \textbf{84.6} \\
Phi-4 & \underline{97} & 97 & \textbf{100} & 26.9 & \underline{32.7} & \textbf{58.2} & 15.1 & \underline{26.6} & \textbf{74.4} & 35.7 & \underline{59.3} & \textbf{84.0} \\
\bottomrule
\end{tabular}
}%  end resizebox
\end{table}

   \textbf{Arithmetic fine-tuning closes the gap to the frontier.}
   Arithmetic FT raises generation quality from a baseline range of
   $0.1$--$26.9\%$ to $51.9$--$59.5\%$ across all 10 open-weight models
   (\Cref{tab:ft_assembly_summary}), and guard pass rates reach $100\%$,
   eliminating leakage entirely. Self-communication reaches $79$--$86\%$
   above 4B and $62\%$ even for Qwen3-0.6B. The bottleneck quantified in
   \Cref{sec:results} is therefore not architectural: training on the
   same operator set with disjoint variables and constants is enough to
   push every panel model above the strongest untrained frontier generator
   (Gemini-3.1-Pro, $\genqual{}=45.1\%$). Because training and evaluation
   expressions share operator semantics and tree topology, this is best
   read as an upper bound on what the channel admits under matched
   semantics, not a generalization claim.

\textbf{Structural fine-tuning lifts all generators, weakest most.}
Assembly FT raises generation quality for every open-weight model, by $+3.3$ to $+16.6$~pp, with the largest gains on the weakest generators (Gemma-3-12B $+12.8$, Gemma-3-27B $+16.6$) and smaller but consistently positive gains on the strongest (Qwen3-8B $+12.1$, Phi-4 $+5.8$). Extraction quality improves uniformly, $+0.7$ to $+11.5$~pp, with the strongest extractors gaining most. Self-communication moves with both, reaching $59.3\%$ for Phi-4 and $41.6\%$ for Qwen3-32B. Guard pass rate falls on average from $85.5\%$ (base) to $83.4\%$. The drop is concentrated in Gemma-3-4B ($-22.9$~pp) and Gemma-3-12B ($-17.5$), while Phi-4 and the larger Qwen3 models are essentially unchanged (\Cref{app:finetuning_assembly}).

\textbf{Cross-domain gains are real but a gap to the frontier persists.}
Because the 10 base extractors never see the FT data, assembly-FT gains reflect more decodable problems rather than a shared artifact. Even so, the strongest assembly-FT generator (Phi-4, $32.7\%$) stays $12.4$~pp below untrained Gemini-3.1-Pro ($45.1\%$). Arithmetic FT closes that gap: every open-weight model exceeds Gemini-3.1-Pro, but only by sharing operator semantics and tree topology with evaluation, so it is best read as an upper bound.
% ─────────────────────────────────────────────────────────────────────────────

\section{Related Work}
\label{sec:related}

    \textbf{Round-trip and consistency evaluation.}
Round-trip protocols are established quality proxies, going back to
back-translation~\citep{sennrich2016improving}. Closest to us,
\citet{allamanis2024unsupervised} formalize round-trip correctness
for code LLMs, using the \emph{same} model on both ends to avoid a
``communication chasm''. \citet{maveli2026compress} round-trip code
through compression--decompression bijections, finding
persistent gaps across prompting, fine-tuning, and self-reflection,
and \citet{kong2025rtrl} extend the framing as a
training reward for molecule--text LLMs.
Both use a same-model round-trip.
\citet{wigler2026stories} use a cross-model NL round-trip like
ours to recover personality profiles from life stories, but
transmit a trait vector at fixed complexity. We study
Allamanis' chasm directly for compositional content: an $N{\times} N$
generator--extractor matrix, symbolic equivalence as the oracle,
and controlled complexity, with failures attributed via row and
column marginals. Unlike self-consistency~\citep{wang2022self},
which aggregates reasoning paths from one model, we test whether a
second reader recovers the sender's structure through the NL
channel, complementing work on internal
faithfulness~\citep{lanham2023measuring, turpin2023language} with
an externally verifiable signal.

\textbf{Multi-agent LLM systems.}
MetaGPT~\citep{hong2023metagpt}, CAMEL~\citep{li2023camel},
AutoGen~\citep{wu2024autogen}, and
Mixture-of-Agents~\citep{wang2024mixture} route coordination through
NL but do not measure information loss at that
interface, even when debate improves
factuality~\citep{du2024improving}. \citet{zhou2025why} argue NL is
fundamentally misaligned with LLM representations.
\citet{he2026information} formalize this as an information bottleneck,
and \citet{tran2026singleagent} show that multi-agent decomposition inherently loses information through
the NL channel. Our communication matrix provides direct empirical
support, and the 60.4~pp role-swap effect has immediate implications
for pipeline design.
Cross-model~\citep{yin2023exchange} and heterogeneous-LLM~\citep{ye2025xmas}
pipelines improve over single-model baselines but do not measure
text-interface loss, while LatentMAS~\citep{zou2025latentmas} sidesteps
it by collaborating in a shared latent space.

\textbf{Structured channels for LLMs.}
Early frameworks bridged LLMs with structured interfaces via SQL-like
operations~\citep{jiang2023structgpt}, but recent work identifies a
persistent serialization gap: linearizing structured data into prose
breaks order invariance and makes reasoning brittle to arbitrary
formatting choices~\citep{herbst2025lost}. Remediation moves beyond
text-wrapping through hypergraph encodings~\citep{huang2025hyperg} and
table-native architectures~\citep{li2025table}, yet frontier models
still show format-dependent accuracy on structured
outputs~\citep{gpt4oprompts}. Our NL tax quantifies this loss in a
controlled symbolic setting.
Recent proposals replace NL with dense vectors~\citep{wu2025dense}
or a layered telecom-style protocol~\citep{li2025lacp}, and our
NL tax is the floor such alternatives would have to beat.
    
    % \textbf{Compositional generalization.}
    % SCAN~\citep{lake2018scan} and CFQ~\citep{keysers2020cfq} evaluate
    % compositional \emph{parsing} (text-to-structure). Transformer
    % performance decays with task depth~\citep{dziri2023faith}, exhibits a
    % ``compositionality gap''~\citep{press2023measuring}, and hits
    % architectural limits at high recursive depth~\citep{thomm2024limits}.
    % The same gradients appear in math reasoning under controlled
    % perturbations~\citep{stolfo2023causal} and procedurally generated
    % proof difficulty~\citep{opedal2024mathgap}, evaluated as solving
    % rather than transmission. We invert the direction to \emph{generation}
    % (structure-to-text), using linearization complexity $\ell(T)$ to
    % connect these failures to surface realization
    % constraints~\citep{Reiter_Dale_2000, zhang-etal-2006-synchronous}.
    % Recent work uses the equation-to-word-problem direction for data
    % augmentation~\citep{chen2024controlmath}, but does not study whether
    % the generated text preserves enough structure for a second model to
    % recover the original.

       \textbf{Compositional generalization.}
   SCAN~\citep{lake2018scan} and CFQ~\citep{keysers2020cfq} evaluate
   compositional \emph{parsing} (text-to-structure). Transformer
   performance decays with depth~\citep{dziri2023faith}, shows a
   ``compositionality gap''~\citep{press2023measuring}, and collapses at
   high complexity~\citep{thomm2024limits, mirzadeh2024gsm, shojaee2025illusion}. The
   same gradients appear in math reasoning~\citep{stolfo2023causal,opedal2024mathgap},
   evaluated as solving rather than transmission. We invert the
   direction to \emph{generation} (structure-to-text), connecting these
   failures to surface realization~\citep{Reiter_Dale_2000, zhang-etal-2006-synchronous}
   via linearization complexity $\ell(T)$. Recent work uses
   equation-to-word-problem generation for data
   augmentation~\citep{chen2024controlmath}, but does not ask whether
   the text preserves enough structure for a second model to recover
   the original.

\section{Discussion}
\label{sec:discussion}
% ─────────────────────────────────────────────────────────────────────────────

\textbf{The bottleneck is compositional serialization, not domain knowledge.}
Swapping arithmetic for logical operators changes absolute
difficulty but preserves rankings and the generator-dominant
fault pattern (\Cref{app:logic}). Fine-tuning on a novel
operator vocabulary transfers broadly, lifting every open-weight
generator but still leaving a gap to the frontier
(\Cref{sec:ft_results}). Under matched operator semantics,
$\sim\!3600$ examples lift every open-weight model above
untrained Gemini-3.1-Pro, so the limit is trainable rather than
architectural. What stays hard is flattening a
hierarchical expression into words another model can re-parse,
a skill chain-of-thought and multi-agent pipelines typically
rely on.

\textbf{Practical recommendations.}
Role assignment matters: swapping generator and extractor
shifts accuracy by up to $60.4$~pp, and the best pair
($92.9\%$) mixes two different models, so match models to roles
by their per-role rank rather than overall strength. Prefer structured intermediates when
the consumer can parse them~\citep{gao2023pal,chen2023pot}
\rebuttal{or when an interoperability standard such as
MCP~\citep{mcp2025} or A2A~\citep{a2a2025github} already provides the
schema}, the NL tax reaches 18--20~pp at 14B--32B
(\Cref{fig:broadcaster}-right).

\textbf{Limitations.}
The protocol covers arithmetic with 2--8 binary operators and a
logic replication (\Cref{app:logic}) \rebuttal{which both admit a
canonical form, so any two correct answers are provably
equivalent. Our protocol requires this kind of oracle, so it does
not directly transfer to domains where correctness admits many
answers with no canonical form to check against, only validators
specific to each instance, like planning, spatial reasoning, and
program synthesis.} Evaluation is monolingual (English)
with one prompt pair and no tuning, so scores are a reproducible lower
bound. Thinking
modes are disabled to isolate serialization from test-time reasoning
and keep the panel comparable.

% ─────────────────────────────────────────────────────────────────────────────
\section{Conclusion}
\label{sec:conclusion}
% ─────────────────────────────────────────────────────────────────────────────

We introduced a round-trip protocol that measures how faithfully
language models transmit compositional structure through natural
language. What fails is predictable: operator count and tree shape dominate,
while model family does not. A logic replication keeps most of the
ranking structure, so the failure pattern is not
unique to arithmetic even though absolute difficulty changes. What fails is also partially fixable:
arithmetic fine-tuning saturates the channel for every tested model,
and structural gains transfer to non-fine-tuned extractors. Natural language is a
lossy channel for compositional structure, one whose capacity scales
with model capability but remains structurally bounded. As language
models increasingly communicate with each other, understanding the
capacity and failure modes of this channel becomes a first-order
concern for system design.

\bibliographystyle{abbrvnat}
\bibliography{biblio}

\newpage
% Automatically insert \FloatBarrier before every appendix \section
\pretocmd{\section}{\FloatBarrier}{}{}
\appendix
\crefalias{section}{appendix}

% ─────────────────────────────────────────────────────────────────────────────
\section{Prompt Templates}
\label{app:prompts}
% ─────────────────────────────────────────────────────────────────────────────

\textbf{Generator system prompt.}
The following system prompt is used for the generation phase, with
\texttt{\{domain\}} and \texttt{\{variables\}} filled per expression:

\begin{quote}
\small
\begin{verbatim}
Convert a mathematical expression into a short word problem.

Rules:
1. Set the word problem in the domain of: {domain}.
2. Use EXACTLY these variable placeholders in curly braces: {variables}.
3. If the expression contains integer constants, use them as literal
   numbers in the word problem (e.g. the constant 8 becomes "8" in
   the text).
4. Each operation in the expression must correspond to a clear action
   in the story.
5. Do NOT include any mathematical symbols (+, -, *, /, parentheses)
   in your response.
6. Keep it to 2-3 sentences.
7. Output ONLY the word problem - no explanation, no title.
\end{verbatim}
\end{quote}

\textbf{Generator in-context exemplars.}
When the generator runs at $\gamma{=}3$, three exemplars are appended
to its prompt to form $\promptg{3}$. The exemplars are drawn from a
fixed stratified pool (five entries spanning $k \in \{2,3,4,6,8\}$ and
$d \in \{2,3,4,5\}$, disjoint from the evaluation set); the first
three entries are used at $\gamma{=}3$:

\begin{quote}
\small
\begin{verbatim}
Expression: ( A + B ) * C  [domain: baking]
Word problem: A bakery made {A} loaves in the morning and {B} in the
afternoon. Each loaf was packed into a box with {C} rolls. How many
rolls were packed in total?

Expression: ( C - ( A + B + 2 ) )  [domain: real estate]
Word problem: A developer had {C} plots of land. They sold {A} plots
in the spring, {B} plots in the summer, and set aside 2 for a park.
How many plots remained?

Expression: ( ( E + D ) - A ) - ( A / C )  [domain: mountaineering]
Word problem: A team of climbers reached a peak with {E} supplies and
found {D} more along the way. They used {A} for shelter and rationed
{A} units of fuel over {C} days, consuming one day's ration. How many
supplies remained?
\end{verbatim}
\end{quote}

\textbf{Extractor system prompt.}
The following system prompt is used for the extraction phase:

\begin{quote}
\small
\begin{verbatim}
You are given a word problem where some numbers are replaced by
variable placeholders and others appear as literal integers.
Identify the mathematical expression that represents the computation
described.

Rules:
1. Use ONLY the variable names listed and any integer constants from
   the text.
2. Use parentheses to make the order of operations explicit.
3. Use plain ASCII operators: +, -, *, /. Do NOT use LaTeX, \frac,
   \times, or any markup.
4. Output ONLY the expression - no explanation, no prose.
\end{verbatim}
\end{quote}

\textbf{Extractor in-context exemplars.}
When the extractor runs at $\varepsilon{=}3$, three exemplars are
appended to form $\prompte{3}$, drawn from a stratified pool
paired with the generator exemplars above but reformatted as
\texttt{Variables / Problem / Expression} triples:

\begin{quote}
\small
\begin{verbatim}
Variables: A, B, C
Problem: A bakery made {A} loaves in the morning and {B} in the
afternoon. Each loaf was packed into a box with {C} rolls. How many
rolls were packed in total?
Expression: ( A + B ) * C

Variables: A, B, C
Problem: A developer had {C} plots of land. They sold {A} plots
in the spring, {B} plots in the summer, and set aside 2 for a park.
How many plots remained?
Expression: C - ( A + B + 2 )

Variables: A, C, D, E
Problem: A team of climbers reached a peak with {E} supplies and
found {D} more along the way. They used {A} for shelter and rationed
{A} units of fuel over {C} days, consuming one day's ration. How many
supplies remained?
Expression: ( ( E + D ) - A ) - ( A / C )
\end{verbatim}
\end{quote}

The main matrix evaluates every pairing at
$(\gamma, \varepsilon) = (0, 3)$; see \Cref{app:shot_grid} for the full
ablation over $(\gamma, \varepsilon) \in \{0, 3\}^2$.

% ─────────────────────────────────────────────────────────────────────────────
\section{Expression Suite Details}
\label{app:expr_grid}
% ─────────────────────────────────────────────────────────────────────────────

\Cref{tab:expr_grid} shows the number of expressions generated for each
feasible combination of operator count~$k$ and depth~$d$.  Each cell
is populated by enumerating all skeletons for that $(k, d)$ pair (up to
50) and sampling 5 instantiations per skeleton, following the skeleton
enumeration procedure described in \Cref{sec:expressions}.
Not all $(k, d)$ pairs are feasible: $k$ operators require at least
$\lceil \log_2(k+1) \rceil$ levels of nesting, so shallow depths cannot
accommodate high operator counts.  All 2450 expressions are used in the
main evaluation; each column of the communication matrix is evaluated on
the subset for which that generator passed the leakage guard.

\begin{table}[htb]
\centering
\caption{Number of expressions per operator count~$k$ and depth~$d$.
Cells marked ``---'' correspond to infeasible or empty $(k, d)$ pairs.}
\label{tab:expr_grid}
\begin{tabular}{l rrrrrr}
\toprule
 & $d{=}2$ & $d{=}3$ & $d{=}4$ & $d{=}5$ & $d{=}6$ & Total \\
\midrule
$k{=}2$ & 10  & --- & --- & --- & --- &  10 \\
$k{=}3$ &  5  &  20 & --- & --- & --- &  25 \\
$k{=}4$ & --- &  30 &  40 & --- & --- &  70 \\
$k{=}5$ & --- &  30 & 100 &  80 & --- & 210 \\
$k{=}6$ & --- &  20 & 200 & 250 & 160 & 630 \\
$k{=}7$ & --- &   5 & 250 & 250 & 250 & 755 \\
$k{=}8$ & --- & --- & 250 & 250 & 250 & 750 \\
\midrule
Total   & 15  & 105 & 840 & 830 & 660 & 2450 \\
\bottomrule
\end{tabular}
\end{table}

% Skeleton enumeration algorithm removed — standard recursive
% decomposition of binary trees, now described inline in Sec 2.1.

% ─────────────────────────────────────────────────────────────────────────────
\section{Leakage Guard and Generator Faults}
\label{app:guard}
% ─────────────────────────────────────────────────────────────────────────────

The \emph{leakage guard} is a rule-based filter that rejects any
generated word problem in which mathematical syntax has leaked into
the prose. Two additional checks catch further generator faults
(missing placeholders and degenerate repetition); these are counted
as generator errors in the reported accuracies but are not reported
as leakage.

\textbf{1.\ Leakage check: operator adjacency rejection.}
Four regular-expression patterns reject any word problem in which a
mathematical operator appears adjacent to a variable letter or digit:
\begin{itemize}
  \item \texttt{[A-Z0-9]\textbackslash s*[+*/]} \quad (variable or digit followed by operator)
  \item \texttt{[+*/]\textbackslash s*[A-Z0-9]} \quad (operator followed by variable or digit)
  \item \texttt{\textbackslash(\textbackslash s*[A-Z]} \quad (open parenthesis followed by variable)
  \item \texttt{[A-Z]\textbackslash s*\textbackslash)} \quad (variable followed by close parenthesis)
\end{itemize}
Before pattern matching, all placeholder tokens \texttt{\{X\}} are
replaced with the letter~\texttt{X}, so the check operates on the
surface form that a human reader would see.  These patterns catch
expressions like ``A + B'' or ``(A * C)'' that leak the formal
structure into the word problem.

\textbf{2.\ Placeholder completeness (generator fault).}
Every variable in the original expression must appear as a
curly-brace placeholder (e.g.\ \texttt{\{A\}}, \texttt{\{B\}}) in
the generated text.  A missing placeholder means the model dropped a
variable, making faithful extraction impossible. This is a
generator fault rather than a leakage failure.

\textbf{3.\ Repetition loop detection (generator fault).}
The guard checks for degenerate outputs containing three or more
consecutive repetitions of any word-level $n$-gram (for $n \in
\{2, 3, 4, 5\}$).  Such repetition indicates the model has entered a
generation loop rather than producing a coherent word problem. This
is also a generator fault rather than a leakage failure.

A word problem must pass all three checks to enter the extraction
phase.  The $\Guard$~\% rows in the communication matrices report
pass rates on the leakage check only; the other two fault types are
counted as errors in the reported accuracies but are not shown as
leakage errors.

\paragraph{A tunable guard.}
The guard is a deliberately simple, rule-based filter that
downstream users can tighten or relax for their own setting.
Practitioners building round-trip evaluations on new domains can
reuse the three checks directly, swap in stricter syntactic
patterns (e.g., additional operator glyphs or domain-specific
markers), or replace the rule-based check with a learned
classifier. The bias-direction argument below holds under any
stricter replacement that still counts guard rejections as
round-trip failures.

\paragraph{Known regex gaps.}
Two specifications in the leakage check are deliberately permissive
and admit narrow leak forms that a stricter filter would catch.
The operator class \texttt{[+*/]} omits the minus sign because
\texttt{-} also serves as an English hyphen, and including it
would reject compound adjectives and hyphenated numerals in
otherwise leak-free prose. Leak strings of the form
``\texttt{A - B}'' or ``\texttt{5 - 3}'' therefore pass. The
parenthesis patterns match only letter-adjacent parentheses, so
numeric expressions such as ``\texttt{(5 - 3)}'' also pass. Both
are instances of guard false negatives: they admit word problems
into the extraction phase that a stricter filter would reject. The
bias-direction argument below covers this case. A stricter guard
(adding \texttt{-} contextually, and digit-adjacent parentheses)
would catch more leaks at the cost of some false positives on
natural prose. Because guard rejections remain charged as
round-trip failures, the lower-bound direction on the NL tax is
preserved under any such tightening.

\paragraph{Bias direction under guard false negatives.}
The leakage guard is conservative with respect to the paper's
central claims. If the guard admits a word problem that contains
implicit structural hints (a guard false negative), extractors have
an easier time recovering the expression, inflating round-trip
accuracy. Because guard rejections are already charged as
round-trip failures in our protocol (\Cref{sec:matrix}), a stricter
guard can only move additional word problems into the failure
bucket. Consequently, the NL tax (\Cref{sec:results}) and the
generator-fault rate (\Cref{sec:fault_results}) are lower bounds on
their true values, and the direction of bias strengthens rather
than undermines the conclusion that natural language is a lossy
channel for compositional structure. The reported role-swap
asymmetry is a difference of two accuracies and its sign is not
pinned down by this argument, so we do not claim it as a lower
bound.

% ─────────────────────────────────────────────────────────────────────────────
\section{$R_{\text{math}}$ Definition and Lexicon}
\label{app:rmath}
% ─────────────────────────────────────────────────────────────────────────────

$R_{\text{math}}$ measures the density of mathematical vocabulary in a
generated word problem.  Given a word problem $w = (w_1, w_2, \ldots,
w_n)$ and a curated lexicon~$\mathcal{L}$, we define
\begin{equation}
  R_{\text{math}}(w)
  \;=\;
  \frac{\sum_{i=1}^{n} \mathds{1}[w_i \in \mathcal{L}]}{n},
  \label{eq:rmath}
\end{equation}
where each $w_i$ is lowercased and stripped of punctuation before
lookup.  Higher $R_{\text{math}}$ indicates the model relied more
heavily on arithmetic-signalling vocabulary (e.g.\ ``sum,'' ``divided,''
``remaining'') rather than domain-specific narrative.

The lexicon~$\mathcal{L}$ contains 109 terms curated from 200 sampled
word problems from the full expression set, organised into seven semantic
categories.  The complete lexicon is listed below.

\smallskip
\noindent\textbf{Addition} (11):
\emph{added, adding, adds, combine, combined, plus, sum, total,
altogether, collected, gathered.}

\smallskip
\noindent\textbf{Subtraction} (28):
\emph{subtracted, subtracting, subtracts, minus, remaining, remain,
remained, remains, left, leftover, removed, removing, remove, deducted,
deducting, lost, loses, fewer, less, spent, spend, spending, spends,
reduced, reducing, discarded, unused, withdrawn.}

\smallskip
\noindent\textbf{Multiplication} (9):
\emph{multiplied, multiplies, times, doubled, tripled, twice, per,
each, every.}

\smallskip
\noindent\textbf{Division} (20):
\emph{divided, divides, dividing, division, split, shared, share,
distributed, distributing, distribution, allocated, evenly, equally,
equal, among, portion, portions, fraction, half, halved.}

\smallskip
\noindent\textbf{Result / query} (19):
\emph{calculate, calculating, determine, determined, find, finds,
estimate, estimated, how, many, much, what, number, amount, result,
average, count, counted, counts.}

\smallskip
\noindent\textbf{Change} (18):
\emph{increase, increased, decrease, decreased, additional, extra,
more, gain, gained, gains, earn, earned, earnings, save, saved, cost,
costing, costs.}

\smallskip
\noindent\textbf{Scale} (4):
\emph{scaled, scaling, factor, rate.}

% ─────────────────────────────────────────────────────────────────────────────
\section{Temperature Sensitivity}
\label{app:temperature}
% ─────────────────────────────────────────────────────────────────────────────

All main experiments use greedy decoding (temperature $\tau=0$).
To verify that this choice does not critically bias results, we sweep
$\tau \in \{0.0, 0.1, \ldots, 1.0\}$ on Phi-4 with 5 independent seeds per
temperature point.

\begin{figure}[htb]
  \centering
  \includegraphics[width=\linewidth]{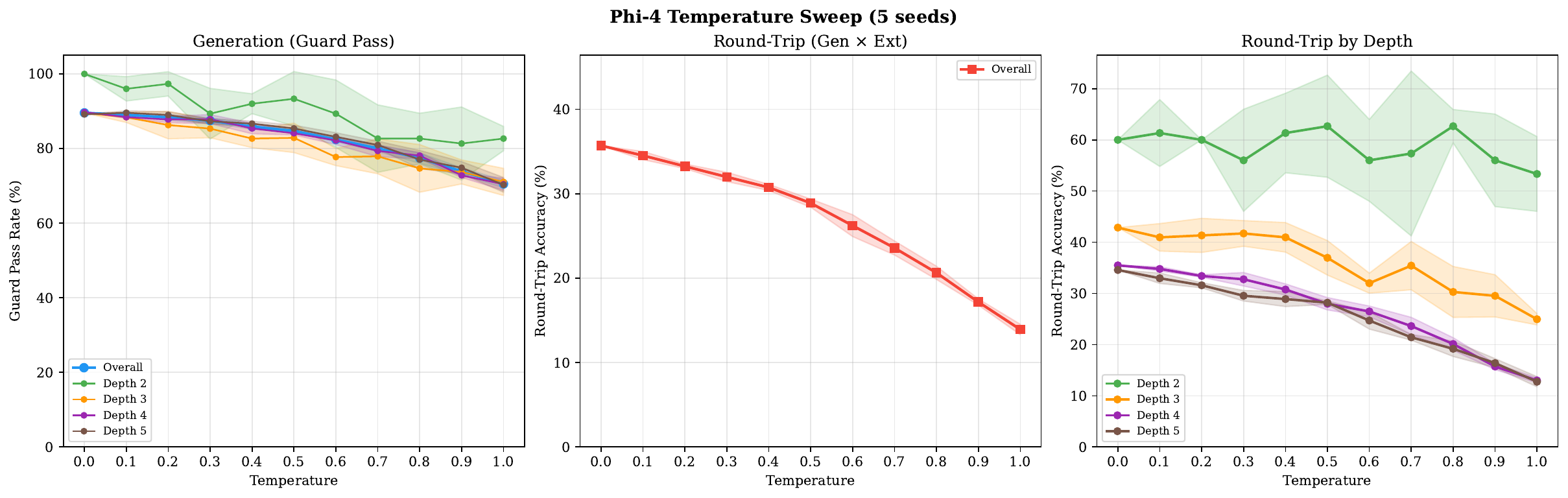}
  \caption{Phi-4 temperature sweep (5 seeds per point).
    Left: guard pass rate.
    Centre: overall round-trip accuracy.
    Right: round-trip accuracy by expression depth.
    Shaded bands show $\pm$1 standard deviation across seeds.}
  \label{fig:temp_sweep}
\end{figure}

\Cref{fig:temp_sweep} summarises the findings.
Guard pass rate (left panel) declines gradually from $89.6\%$
($\tau=0$) to $70.5\%$ ($\tau=1$), confirming that the generation
prompt remains usable across the sampling range.
Round-trip accuracy (centre panel) is relatively stable for
$\tau \le 0.4$ ($35.7\%$ $\to$ $30.7\%$, $-5$\,pp), then drops more
steeply to $13.9\%$ at $\tau=1.0$.
The per-depth breakdown (right panel) reveals that deeper expressions
are disproportionately affected: depth-5 accuracy falls from $34.6\%$
to $12.7\%$ ($-21.9$\,pp), whereas depth-2 accuracy stays in a
narrow range of $53$--$63\%$ across the entire sweep.
Across all temperature points the inter-seed standard deviation stays
below $1.5$\,pp for the overall round-trip metric, indicating low
run-to-run variance.

These results support the use of greedy decoding in the main experiments:
$\tau=0$ maximises round-trip accuracy while eliminating seed variance,
and the relative model rankings are unlikely to change at moderate
temperatures given the flat region below $\tau \approx 0.4$.

% ─────────────────────────────────────────────────────────────────────────────
\section{Per-Depth Communication Matrices}
\label{app:depth_matrices}
% ─────────────────────────────────────────────────────────────────────────────

\Cref{tab:matrix} in the main text aggregates across all depths.  \Cref{tab:matrix_d2,tab:matrix_d3,tab:matrix_d4,tab:matrix_d5,tab:matrix_d6}
show the communication matrix separately for each depth level,
revealing how pairwise accuracy degrades with increasing nesting.

\begin{table}[htb]
\centering
\caption{Communication matrix at depth $d = 2$ (15 expressions). Each cell shows round-trip accuracy (\%). Rows are extractors, columns are generators. Dotted lines separate open-weight AR, diffusion, and frontier models.}
\vspace{2mm}
\label{tab:matrix_d2}
\small
\setlength{\tabcolsep}{2pt}
\setlength{\dashlinedash}{1pt}
\setlength{\dashlinegap}{1pt}
\resizebox{\textwidth}{!}{%
\begin{tabular}{c l c c c c c c c c c c !{\;\vrule width 0.3pt\;} c c !{\;\vrule width 0.3pt\;} c c c c !{\;\vrule width 0.75pt\;} c}
\toprule
\multicolumn{19}{c}{\textsc{Generators}} \\
 \multirow{20}{*}{\rotatebox{90}{\textsc{Extractors}}} & & \rotatebox{90}{\scriptsize Qwen3-0.6B} & \rotatebox{90}{\scriptsize Qwen3-1.7B} & \rotatebox{90}{\scriptsize Qwen3-4B} & \rotatebox{90}{\scriptsize Qwen3-8B} & \rotatebox{90}{\scriptsize Qwen3-14B} & \rotatebox{90}{\scriptsize Qwen3-32B} & \rotatebox{90}{\scriptsize Gemma-3-4B} & \rotatebox{90}{\scriptsize Gemma-3-12B} & \rotatebox{90}{\scriptsize Gemma-3-27B} & \rotatebox{90}{\scriptsize Phi-4} & \rotatebox{90}{\scriptsize Dream-7B} & \rotatebox{90}{\scriptsize LLaDA-8B} & \rotatebox{90}{\scriptsize C-Haiku-4.5} & \rotatebox{90}{\scriptsize GPT-5} & \rotatebox{90}{\scriptsize G-3-Flash} & \rotatebox{90}{\scriptsize G-3.1-Pro} & \rotatebox{90}{\scriptsize Extr.\ quality} \\
\cmidrule[0.5pt](l){2-19}
& {\scriptsize $\Guard$ pass \%} & \cellcolor[RGB]{212,242,212}{\scriptsize 33} & \cellcolor[RGB]{161,226,161}{\scriptsize 73} & \cellcolor[RGB]{136,219,136}{\scriptsize 93} & \cellcolor[RGB]{127,216,127}{\scriptsize 100} & \cellcolor[RGB]{127,216,127}{\scriptsize 100} & \cellcolor[RGB]{127,216,127}{\scriptsize 100} & \cellcolor[RGB]{127,216,127}{\scriptsize 100} & \cellcolor[RGB]{127,216,127}{\scriptsize 100} & \cellcolor[RGB]{127,216,127}{\scriptsize 100} & \cellcolor[RGB]{127,216,127}{\scriptsize 100} & \cellcolor[RGB]{187,234,187}{\scriptsize 53} & \cellcolor[RGB]{127,216,127}{\scriptsize 100} & \cellcolor[RGB]{127,216,127}{\scriptsize 100} & \cellcolor[RGB]{127,216,127}{\scriptsize 100} & \cellcolor[RGB]{127,216,127}{\scriptsize 100} & \cellcolor[RGB]{127,216,127}{\scriptsize 100} &  \\
\cmidrule[0.5pt](l){2-19}
& {\scriptsize Qwen3-0.6B} & \cellcolor[RGB]{255,255,255}{\scriptsize 0.0} & \cellcolor[RGB]{190,209,251}{\scriptsize 6.7} & \cellcolor[RGB]{110,153,246}{\scriptsize 33.3} & \cellcolor[RGB]{72,125,244}{\scriptsize 53.3} & \cellcolor[RGB]{61,118,243}{\scriptsize 60.0} & \cellcolor[RGB]{125,163,247}{\scriptsize 26.7} & \cellcolor[RGB]{255,255,255}{\scriptsize 0.0} & \cellcolor[RGB]{190,209,251}{\scriptsize 6.7} & \cellcolor[RGB]{190,209,251}{\scriptsize 6.7} & \cellcolor[RGB]{110,153,246}{\scriptsize 33.3} & \cellcolor[RGB]{190,209,251}{\scriptsize 6.7} & \cellcolor[RGB]{125,163,247}{\scriptsize 26.7} & \cellcolor[RGB]{125,163,247}{\scriptsize 26.7} & \cellcolor[RGB]{96,143,245}{\scriptsize 40.0} & \cellcolor[RGB]{125,163,247}{\scriptsize 26.7} & \cellcolor[RGB]{110,153,246}{\scriptsize 33.3} & \cellcolor[RGB]{131,168,247}{\scriptsize 24.2\scalebox{0.55}{$\pm$18}} \\
& {\scriptsize Qwen3-1.7B} & \cellcolor[RGB]{255,255,255}{\scriptsize 0.0} & \cellcolor[RGB]{143,175,248}{\scriptsize 20.0} & \cellcolor[RGB]{143,175,248}{\scriptsize 20.0} & \cellcolor[RGB]{61,118,243}{\scriptsize 60.0} & \cellcolor[RGB]{110,153,246}{\scriptsize 33.3} & \cellcolor[RGB]{143,175,248}{\scriptsize 20.0} & \cellcolor[RGB]{190,209,251}{\scriptsize 6.7} & \cellcolor[RGB]{143,175,248}{\scriptsize 20.0} & \cellcolor[RGB]{143,175,248}{\scriptsize 20.0} & \cellcolor[RGB]{72,125,244}{\scriptsize 53.3} & \cellcolor[RGB]{255,255,255}{\scriptsize 0.0} & \cellcolor[RGB]{163,190,249}{\scriptsize 13.3} & \cellcolor[RGB]{84,134,244}{\scriptsize 46.7} & \cellcolor[RGB]{61,118,243}{\scriptsize 60.0} & \cellcolor[RGB]{84,134,244}{\scriptsize 46.7} & \cellcolor[RGB]{61,118,243}{\scriptsize 60.0} & \cellcolor[RGB]{117,158,246}{\scriptsize 30.0\scalebox{0.55}{$\pm$21}} \\
& {\scriptsize Qwen3-4B} & \cellcolor[RGB]{255,255,255}{\scriptsize 0.0} & \cellcolor[RGB]{125,163,247}{\scriptsize 26.7} & \cellcolor[RGB]{110,153,246}{\scriptsize 33.3} & \cellcolor[RGB]{50,110,242}{\scriptsize 66.7} & \cellcolor[RGB]{40,103,242}{\scriptsize 73.3} & \cellcolor[RGB]{72,125,244}{\scriptsize 53.3} & \cellcolor[RGB]{190,209,251}{\scriptsize 6.7} & \cellcolor[RGB]{143,175,248}{\scriptsize 20.0} & \cellcolor[RGB]{110,153,246}{\scriptsize 33.3} & \cellcolor[RGB]{84,134,244}{\scriptsize 46.7} & \cellcolor[RGB]{163,190,249}{\scriptsize 13.3} & \cellcolor[RGB]{96,143,245}{\scriptsize 40.0} & \cellcolor[RGB]{50,110,242}{\scriptsize 66.7} & \cellcolor[RGB]{50,110,242}{\scriptsize 66.7} & \cellcolor[RGB]{40,103,242}{\scriptsize 73.3} & \cellcolor[RGB]{38,102,242}{\scriptsize 93.3} & \cellcolor[RGB]{87,137,245}{\scriptsize 44.6\scalebox{0.55}{$\pm$26}} \\
& {\scriptsize Qwen3-8B} & \cellcolor[RGB]{190,209,251}{\scriptsize 6.7} & \cellcolor[RGB]{143,175,248}{\scriptsize 20.0} & \cellcolor[RGB]{110,153,246}{\scriptsize 33.3} & \cellcolor[RGB]{72,125,244}{\scriptsize 53.3} & \cellcolor[RGB]{38,102,242}{\scriptsize 80.0} & \cellcolor[RGB]{61,118,243}{\scriptsize 60.0} & \cellcolor[RGB]{190,209,251}{\scriptsize 6.7} & \cellcolor[RGB]{125,163,247}{\scriptsize 26.7} & \cellcolor[RGB]{125,163,247}{\scriptsize 26.7} & \cellcolor[RGB]{61,118,243}{\scriptsize 60.0} & \cellcolor[RGB]{143,175,248}{\scriptsize 20.0} & \cellcolor[RGB]{84,134,244}{\scriptsize 46.7} & \cellcolor[RGB]{38,102,242}{\scriptsize 80.0} & \cellcolor[RGB]{38,102,242}{\scriptsize 80.0} & \cellcolor[RGB]{38,102,242}{\scriptsize 86.7} & \cellcolor[RGB]{38,102,242}{\scriptsize 93.3} & \cellcolor[RGB]{80,131,244}{\scriptsize 48.8\scalebox{0.55}{$\pm$29}} \\
& {\scriptsize Qwen3-14B} & \cellcolor[RGB]{190,209,251}{\scriptsize 6.7} & \cellcolor[RGB]{125,163,247}{\scriptsize 26.7} & \cellcolor[RGB]{110,153,246}{\scriptsize 33.3} & \cellcolor[RGB]{61,118,243}{\scriptsize 60.0} & \cellcolor[RGB]{40,103,242}{\scriptsize 73.3} & \cellcolor[RGB]{61,118,243}{\scriptsize 60.0} & \cellcolor[RGB]{163,190,249}{\scriptsize 13.3} & \cellcolor[RGB]{143,175,248}{\scriptsize 20.0} & \cellcolor[RGB]{110,153,246}{\scriptsize 33.3} & \cellcolor[RGB]{61,118,243}{\scriptsize 60.0} & \cellcolor[RGB]{143,175,248}{\scriptsize 20.0} & \cellcolor[RGB]{96,143,245}{\scriptsize 40.0} & \cellcolor[RGB]{38,102,242}{\scriptsize 80.0} & \cellcolor[RGB]{38,102,242}{\scriptsize 80.0} & \cellcolor[RGB]{38,102,242}{\scriptsize 93.3} & \cellcolor[RGB]{38,102,242}{\scriptsize 100.0} & \cellcolor[RGB]{78,130,244}{\scriptsize 50.0\scalebox{0.55}{$\pm$29}} \\
& {\scriptsize Qwen3-32B} & \cellcolor[RGB]{190,209,251}{\scriptsize 6.7} & \cellcolor[RGB]{125,163,247}{\scriptsize 26.7} & \cellcolor[RGB]{110,153,246}{\scriptsize 33.3} & \cellcolor[RGB]{61,118,243}{\scriptsize 60.0} & \cellcolor[RGB]{40,103,242}{\scriptsize 73.3} & \cellcolor[RGB]{72,125,244}{\scriptsize 53.3} & \cellcolor[RGB]{163,190,249}{\scriptsize 13.3} & \cellcolor[RGB]{143,175,248}{\scriptsize 20.0} & \cellcolor[RGB]{110,153,246}{\scriptsize 33.3} & \cellcolor[RGB]{61,118,243}{\scriptsize 60.0} & \cellcolor[RGB]{125,163,247}{\scriptsize 26.7} & \cellcolor[RGB]{84,134,244}{\scriptsize 46.7} & \cellcolor[RGB]{38,102,242}{\scriptsize 86.7} & \cellcolor[RGB]{38,102,242}{\scriptsize 80.0} & \cellcolor[RGB]{38,102,242}{\scriptsize 93.3} & \cellcolor[RGB]{38,102,242}{\scriptsize 100.0} & \cellcolor[RGB]{76,129,244}{\scriptsize 50.8\scalebox{0.55}{$\pm$29}} \\
& {\scriptsize Gemma-3-4B} & \cellcolor[RGB]{190,209,251}{\scriptsize 6.7} & \cellcolor[RGB]{143,175,248}{\scriptsize 20.0} & \cellcolor[RGB]{143,175,248}{\scriptsize 20.0} & \cellcolor[RGB]{72,125,244}{\scriptsize 53.3} & \cellcolor[RGB]{61,118,243}{\scriptsize 60.0} & \cellcolor[RGB]{84,134,244}{\scriptsize 46.7} & \cellcolor[RGB]{190,209,251}{\scriptsize 6.7} & \cellcolor[RGB]{110,153,246}{\scriptsize 33.3} & \cellcolor[RGB]{125,163,247}{\scriptsize 26.7} & \cellcolor[RGB]{96,143,245}{\scriptsize 40.0} & \cellcolor[RGB]{143,175,248}{\scriptsize 20.0} & \cellcolor[RGB]{143,175,248}{\scriptsize 20.0} & \cellcolor[RGB]{72,125,244}{\scriptsize 53.3} & \cellcolor[RGB]{50,110,242}{\scriptsize 66.7} & \cellcolor[RGB]{40,103,242}{\scriptsize 73.3} & \cellcolor[RGB]{38,102,242}{\scriptsize 93.3} & \cellcolor[RGB]{96,143,245}{\scriptsize 40.0\scalebox{0.55}{$\pm$24}} \\
& {\scriptsize Gemma-3-12B} & \cellcolor[RGB]{255,255,255}{\scriptsize 0.0} & \cellcolor[RGB]{110,153,246}{\scriptsize 33.3} & \cellcolor[RGB]{110,153,246}{\scriptsize 33.3} & \cellcolor[RGB]{61,118,243}{\scriptsize 60.0} & \cellcolor[RGB]{40,103,242}{\scriptsize 73.3} & \cellcolor[RGB]{61,118,243}{\scriptsize 60.0} & \cellcolor[RGB]{163,190,249}{\scriptsize 13.3} & \cellcolor[RGB]{125,163,247}{\scriptsize 26.7} & \cellcolor[RGB]{125,163,247}{\scriptsize 26.7} & \cellcolor[RGB]{72,125,244}{\scriptsize 53.3} & \cellcolor[RGB]{163,190,249}{\scriptsize 13.3} & \cellcolor[RGB]{84,134,244}{\scriptsize 46.7} & \cellcolor[RGB]{38,102,242}{\scriptsize 80.0} & \cellcolor[RGB]{50,110,242}{\scriptsize 66.7} & \cellcolor[RGB]{38,102,242}{\scriptsize 100.0} & \cellcolor[RGB]{38,102,242}{\scriptsize 93.3} & \cellcolor[RGB]{80,131,244}{\scriptsize 48.7\scalebox{0.55}{$\pm$29}} \\
& {\scriptsize Gemma-3-27B} & \cellcolor[RGB]{190,209,251}{\scriptsize 6.7} & \cellcolor[RGB]{125,163,247}{\scriptsize 26.7} & \cellcolor[RGB]{110,153,246}{\scriptsize 33.3} & \cellcolor[RGB]{61,118,243}{\scriptsize 60.0} & \cellcolor[RGB]{40,103,242}{\scriptsize 73.3} & \cellcolor[RGB]{72,125,244}{\scriptsize 53.3} & \cellcolor[RGB]{190,209,251}{\scriptsize 6.7} & \cellcolor[RGB]{125,163,247}{\scriptsize 26.7} & \cellcolor[RGB]{110,153,246}{\scriptsize 33.3} & \cellcolor[RGB]{61,118,243}{\scriptsize 60.0} & \cellcolor[RGB]{163,190,249}{\scriptsize 13.3} & \cellcolor[RGB]{84,134,244}{\scriptsize 46.7} & \cellcolor[RGB]{38,102,242}{\scriptsize 80.0} & \cellcolor[RGB]{38,102,242}{\scriptsize 93.3} & \cellcolor[RGB]{38,102,242}{\scriptsize 86.7} & \cellcolor[RGB]{38,102,242}{\scriptsize 100.0} & \cellcolor[RGB]{78,130,244}{\scriptsize 50.0\scalebox{0.55}{$\pm$30}} \\
& {\scriptsize Phi-4} & \cellcolor[RGB]{255,255,255}{\scriptsize 0.0} & \cellcolor[RGB]{125,163,247}{\scriptsize 26.7} & \cellcolor[RGB]{125,163,247}{\scriptsize 26.7} & \cellcolor[RGB]{84,134,244}{\scriptsize 46.7} & \cellcolor[RGB]{40,103,242}{\scriptsize 73.3} & \cellcolor[RGB]{61,118,243}{\scriptsize 60.0} & \cellcolor[RGB]{190,209,251}{\scriptsize 6.7} & \cellcolor[RGB]{143,175,248}{\scriptsize 20.0} & \cellcolor[RGB]{110,153,246}{\scriptsize 33.3} & \cellcolor[RGB]{61,118,243}{\scriptsize 60.0} & \cellcolor[RGB]{190,209,251}{\scriptsize 6.7} & \cellcolor[RGB]{84,134,244}{\scriptsize 46.7} & \cellcolor[RGB]{38,102,242}{\scriptsize 80.0} & \cellcolor[RGB]{38,102,242}{\scriptsize 80.0} & \cellcolor[RGB]{38,102,242}{\scriptsize 100.0} & \cellcolor[RGB]{38,102,242}{\scriptsize 100.0} & \cellcolor[RGB]{81,132,244}{\scriptsize 47.9\scalebox{0.55}{$\pm$32}} \\
\cmidrule[0.3pt](l){2-19}
& {\scriptsize Dream-7B} & \cellcolor[RGB]{255,255,255}{\scriptsize 0.0} & \cellcolor[RGB]{143,175,248}{\scriptsize 20.0} & \cellcolor[RGB]{163,190,249}{\scriptsize 13.3} & \cellcolor[RGB]{50,110,242}{\scriptsize 66.7} & \cellcolor[RGB]{72,125,244}{\scriptsize 53.3} & \cellcolor[RGB]{110,153,246}{\scriptsize 33.3} & \cellcolor[RGB]{190,209,251}{\scriptsize 6.7} & \cellcolor[RGB]{143,175,248}{\scriptsize 20.0} & \cellcolor[RGB]{125,163,247}{\scriptsize 26.7} & \cellcolor[RGB]{61,118,243}{\scriptsize 60.0} & \cellcolor[RGB]{255,255,255}{\scriptsize 0.0} & \cellcolor[RGB]{125,163,247}{\scriptsize 26.7} & \cellcolor[RGB]{72,125,244}{\scriptsize 53.3} & \cellcolor[RGB]{38,102,242}{\scriptsize 80.0} & \cellcolor[RGB]{50,110,242}{\scriptsize 66.7} & \cellcolor[RGB]{50,110,242}{\scriptsize 66.7} & \cellcolor[RGB]{102,147,246}{\scriptsize 37.1\scalebox{0.55}{$\pm$26}} \\
& {\scriptsize LLaDA-8B} & \cellcolor[RGB]{255,255,255}{\scriptsize 0.0} & \cellcolor[RGB]{143,175,248}{\scriptsize 20.0} & \cellcolor[RGB]{125,163,247}{\scriptsize 26.7} & \cellcolor[RGB]{61,118,243}{\scriptsize 60.0} & \cellcolor[RGB]{38,102,242}{\scriptsize 86.7} & \cellcolor[RGB]{61,118,243}{\scriptsize 60.0} & \cellcolor[RGB]{255,255,255}{\scriptsize 0.0} & \cellcolor[RGB]{110,153,246}{\scriptsize 33.3} & \cellcolor[RGB]{96,143,245}{\scriptsize 40.0} & \cellcolor[RGB]{61,118,243}{\scriptsize 60.0} & \cellcolor[RGB]{125,163,247}{\scriptsize 26.7} & \cellcolor[RGB]{72,125,244}{\scriptsize 53.3} & \cellcolor[RGB]{40,103,242}{\scriptsize 73.3} & \cellcolor[RGB]{38,102,242}{\scriptsize 93.3} & \cellcolor[RGB]{38,102,242}{\scriptsize 86.7} & \cellcolor[RGB]{38,102,242}{\scriptsize 93.3} & \cellcolor[RGB]{76,129,244}{\scriptsize 50.8\scalebox{0.55}{$\pm$30}} \\
\cmidrule[0.3pt](l){2-19}
& {\scriptsize C-Haiku-4.5} & \cellcolor[RGB]{255,255,255}{\scriptsize 0.0} & \cellcolor[RGB]{143,175,248}{\scriptsize 20.0} & \cellcolor[RGB]{84,134,244}{\scriptsize 46.7} & \cellcolor[RGB]{72,125,244}{\scriptsize 53.3} & \cellcolor[RGB]{38,102,242}{\scriptsize 80.0} & \cellcolor[RGB]{72,125,244}{\scriptsize 53.3} & \cellcolor[RGB]{163,190,249}{\scriptsize 13.3} & \cellcolor[RGB]{125,163,247}{\scriptsize 26.7} & \cellcolor[RGB]{110,153,246}{\scriptsize 33.3} & \cellcolor[RGB]{61,118,243}{\scriptsize 60.0} & \cellcolor[RGB]{125,163,247}{\scriptsize 26.7} & \cellcolor[RGB]{84,134,244}{\scriptsize 46.7} & \cellcolor[RGB]{38,102,242}{\scriptsize 86.7} & \cellcolor[RGB]{38,102,242}{\scriptsize 100.0} & \cellcolor[RGB]{38,102,242}{\scriptsize 86.7} & \cellcolor[RGB]{38,102,242}{\scriptsize 100.0} & \cellcolor[RGB]{74,127,244}{\scriptsize 52.1\scalebox{0.55}{$\pm$30}} \\
& {\scriptsize GPT-5} & \cellcolor[RGB]{190,209,251}{\scriptsize 6.7} & \cellcolor[RGB]{125,163,247}{\scriptsize 26.7} & \cellcolor[RGB]{125,163,247}{\scriptsize 26.7} & \cellcolor[RGB]{72,125,244}{\scriptsize 53.3} & \cellcolor[RGB]{38,102,242}{\scriptsize 80.0} & \cellcolor[RGB]{61,118,243}{\scriptsize 60.0} & \cellcolor[RGB]{190,209,251}{\scriptsize 6.7} & \cellcolor[RGB]{125,163,247}{\scriptsize 26.7} & \cellcolor[RGB]{84,134,244}{\scriptsize 46.7} & \cellcolor[RGB]{61,118,243}{\scriptsize 60.0} & \cellcolor[RGB]{163,190,249}{\scriptsize 13.3} & \cellcolor[RGB]{72,125,244}{\scriptsize 53.3} & \cellcolor[RGB]{40,103,242}{\scriptsize 73.3} & \cellcolor[RGB]{38,102,242}{\scriptsize 100.0} & \cellcolor[RGB]{38,102,242}{\scriptsize 93.3} & \cellcolor[RGB]{38,102,242}{\scriptsize 100.0} & \cellcolor[RGB]{75,128,244}{\scriptsize 51.7\scalebox{0.55}{$\pm$31}} \\
& {\scriptsize G-3-Flash} & \cellcolor[RGB]{190,209,251}{\scriptsize 6.7} & \cellcolor[RGB]{143,175,248}{\scriptsize 20.0} & \cellcolor[RGB]{96,143,245}{\scriptsize 40.0} & \cellcolor[RGB]{72,125,244}{\scriptsize 53.3} & \cellcolor[RGB]{38,102,242}{\scriptsize 80.0} & \cellcolor[RGB]{61,118,243}{\scriptsize 60.0} & \cellcolor[RGB]{190,209,251}{\scriptsize 6.7} & \cellcolor[RGB]{125,163,247}{\scriptsize 26.7} & \cellcolor[RGB]{96,143,245}{\scriptsize 40.0} & \cellcolor[RGB]{61,118,243}{\scriptsize 60.0} & \cellcolor[RGB]{125,163,247}{\scriptsize 26.7} & \cellcolor[RGB]{84,134,244}{\scriptsize 46.7} & \cellcolor[RGB]{38,102,242}{\scriptsize 86.7} & \cellcolor[RGB]{38,102,242}{\scriptsize 100.0} & \cellcolor[RGB]{38,102,242}{\scriptsize 86.7} & \cellcolor[RGB]{38,102,242}{\scriptsize 100.0} & \cellcolor[RGB]{73,126,244}{\scriptsize 52.5\scalebox{0.55}{$\pm$30}} \\
& {\scriptsize G-3.1-Pro} & \cellcolor[RGB]{255,255,255}{\scriptsize 0.0} & \cellcolor[RGB]{163,190,249}{\scriptsize 13.3} & \cellcolor[RGB]{96,143,245}{\scriptsize 40.0} & \cellcolor[RGB]{61,118,243}{\scriptsize 60.0} & \cellcolor[RGB]{40,103,242}{\scriptsize 73.3} & \cellcolor[RGB]{50,110,242}{\scriptsize 66.7} & \cellcolor[RGB]{190,209,251}{\scriptsize 6.7} & \cellcolor[RGB]{125,163,247}{\scriptsize 26.7} & \cellcolor[RGB]{110,153,246}{\scriptsize 33.3} & \cellcolor[RGB]{61,118,243}{\scriptsize 60.0} & \cellcolor[RGB]{125,163,247}{\scriptsize 26.7} & \cellcolor[RGB]{84,134,244}{\scriptsize 46.7} & \cellcolor[RGB]{40,103,242}{\scriptsize 73.3} & \cellcolor[RGB]{38,102,242}{\scriptsize 100.0} & \cellcolor[RGB]{38,102,242}{\scriptsize 100.0} & \cellcolor[RGB]{38,102,242}{\scriptsize 100.0} & \cellcolor[RGB]{75,128,244}{\scriptsize 51.7\scalebox{0.55}{$\pm$32}} \\
\cmidrule[0.5pt](l){2-19}
& {\scriptsize Gen.\ quality} & \cellcolor[RGB]{212,224,252}{\scriptsize 2.9\scalebox{0.55}{$\pm$3}} & \cellcolor[RGB]{137,171,248}{\scriptsize 22.1\scalebox{0.55}{$\pm$6}} & \cellcolor[RGB]{116,156,246}{\scriptsize 30.8\scalebox{0.55}{$\pm$8}} & \cellcolor[RGB]{65,121,243}{\scriptsize 57.5\scalebox{0.55}{$\pm$5}} & \cellcolor[RGB]{44,106,242}{\scriptsize 70.4\scalebox{0.55}{$\pm$13}} & \cellcolor[RGB]{75,128,244}{\scriptsize 51.7\scalebox{0.55}{$\pm$13}} & \cellcolor[RGB]{186,206,250}{\scriptsize 7.5\scalebox{0.55}{$\pm$4}} & \cellcolor[RGB]{133,168,247}{\scriptsize 23.8\scalebox{0.55}{$\pm$6}} & \cellcolor[RGB]{116,156,246}{\scriptsize 30.8\scalebox{0.55}{$\pm$9}} & \cellcolor[RGB]{68,123,244}{\scriptsize 55.4\scalebox{0.55}{$\pm$8}} & \cellcolor[RGB]{154,183,249}{\scriptsize 16.2\scalebox{0.55}{$\pm$9}} & \cellcolor[RGB]{95,142,245}{\scriptsize 40.4\scalebox{0.55}{$\pm$12}} & \cellcolor[RGB]{44,106,242}{\scriptsize 70.4\scalebox{0.55}{$\pm$16}} & \cellcolor[RGB]{38,102,242}{\scriptsize 80.4\scalebox{0.55}{$\pm$17}} & \cellcolor[RGB]{38,102,242}{\scriptsize 81.2\scalebox{0.55}{$\pm$20}} & \cellcolor[RGB]{38,102,242}{\scriptsize 89.2\scalebox{0.55}{$\pm$19}} &  \\
\bottomrule
\end{tabular}
}%  end resizebox
\end{table}

\begin{table}[htb]
\centering
\caption{Communication matrix at depth $d = 3$ (105 expressions). Each cell shows round-trip accuracy (\%). Rows are extractors, columns are generators. Dotted lines separate open-weight AR, diffusion, and frontier models.}
\vspace{2mm}
\label{tab:matrix_d3}
\small
\setlength{\tabcolsep}{2pt}
\setlength{\dashlinedash}{1pt}
\setlength{\dashlinegap}{1pt}
\resizebox{\textwidth}{!}{%
\begin{tabular}{c l c c c c c c c c c c !{\;\vrule width 0.3pt\;} c c !{\;\vrule width 0.3pt\;} c c c c !{\;\vrule width 0.75pt\;} c}
\toprule
\multicolumn{19}{c}{\textsc{Generators}} \\
 \multirow{20}{*}{\rotatebox{90}{\textsc{Extractors}}} & & \rotatebox{90}{\scriptsize Qwen3-0.6B} & \rotatebox{90}{\scriptsize Qwen3-1.7B} & \rotatebox{90}{\scriptsize Qwen3-4B} & \rotatebox{90}{\scriptsize Qwen3-8B} & \rotatebox{90}{\scriptsize Qwen3-14B} & \rotatebox{90}{\scriptsize Qwen3-32B} & \rotatebox{90}{\scriptsize Gemma-3-4B} & \rotatebox{90}{\scriptsize Gemma-3-12B} & \rotatebox{90}{\scriptsize Gemma-3-27B} & \rotatebox{90}{\scriptsize Phi-4} & \rotatebox{90}{\scriptsize Dream-7B} & \rotatebox{90}{\scriptsize LLaDA-8B} & \rotatebox{90}{\scriptsize C-Haiku-4.5} & \rotatebox{90}{\scriptsize GPT-5} & \rotatebox{90}{\scriptsize G-3-Flash} & \rotatebox{90}{\scriptsize G-3.1-Pro} & \rotatebox{90}{\scriptsize Extr.\ quality} \\
\cmidrule[0.5pt](l){2-19}
& {\scriptsize $\Guard$ pass \%} & \cellcolor[RGB]{239,250,239}{\scriptsize 12} & \cellcolor[RGB]{166,228,166}{\scriptsize 70} & \cellcolor[RGB]{128,217,128}{\scriptsize 99} & \cellcolor[RGB]{128,217,128}{\scriptsize 99} & \cellcolor[RGB]{127,216,127}{\scriptsize 100} & \cellcolor[RGB]{128,217,128}{\scriptsize 99} & \cellcolor[RGB]{127,216,127}{\scriptsize 100} & \cellcolor[RGB]{128,217,128}{\scriptsize 99} & \cellcolor[RGB]{131,217,131}{\scriptsize 97} & \cellcolor[RGB]{133,218,133}{\scriptsize 95} & \cellcolor[RGB]{172,230,172}{\scriptsize 65} & \cellcolor[RGB]{129,217,129}{\scriptsize 98} & \cellcolor[RGB]{128,217,128}{\scriptsize 99} & \cellcolor[RGB]{127,216,127}{\scriptsize 100} & \cellcolor[RGB]{127,216,127}{\scriptsize 100} & \cellcolor[RGB]{127,216,127}{\scriptsize 100} &  \\
\cmidrule[0.5pt](l){2-19}
& {\scriptsize Qwen3-0.6B} & \cellcolor[RGB]{230,237,253}{\scriptsize 1.0} & \cellcolor[RGB]{190,209,251}{\scriptsize 6.7} & \cellcolor[RGB]{163,190,249}{\scriptsize 13.3} & \cellcolor[RGB]{145,177,248}{\scriptsize 19.0} & \cellcolor[RGB]{163,190,249}{\scriptsize 13.3} & \cellcolor[RGB]{163,190,249}{\scriptsize 13.3} & \cellcolor[RGB]{255,255,255}{\scriptsize 0.0} & \cellcolor[RGB]{212,225,252}{\scriptsize 2.9} & \cellcolor[RGB]{230,237,253}{\scriptsize 1.0} & \cellcolor[RGB]{157,186,249}{\scriptsize 15.2} & \cellcolor[RGB]{200,216,251}{\scriptsize 4.8} & \cellcolor[RGB]{154,183,249}{\scriptsize 16.2} & \cellcolor[RGB]{145,177,248}{\scriptsize 19.0} & \cellcolor[RGB]{185,206,250}{\scriptsize 7.6} & \cellcolor[RGB]{166,192,249}{\scriptsize 12.4} & \cellcolor[RGB]{185,206,250}{\scriptsize 7.6} & \cellcolor[RGB]{177,200,250}{\scriptsize 9.6\scalebox{0.55}{$\pm$6}} \\
& {\scriptsize Qwen3-1.7B} & \cellcolor[RGB]{230,237,253}{\scriptsize 1.0} & \cellcolor[RGB]{177,200,250}{\scriptsize 9.5} & \cellcolor[RGB]{140,174,248}{\scriptsize 21.0} & \cellcolor[RGB]{128,165,247}{\scriptsize 25.7} & \cellcolor[RGB]{114,155,246}{\scriptsize 31.4} & \cellcolor[RGB]{137,172,248}{\scriptsize 21.9} & \cellcolor[RGB]{206,220,252}{\scriptsize 3.8} & \cellcolor[RGB]{166,192,249}{\scriptsize 12.4} & \cellcolor[RGB]{185,206,250}{\scriptsize 7.6} & \cellcolor[RGB]{119,159,247}{\scriptsize 29.5} & \cellcolor[RGB]{212,225,252}{\scriptsize 2.9} & \cellcolor[RGB]{112,154,246}{\scriptsize 32.4} & \cellcolor[RGB]{84,134,244}{\scriptsize 46.7} & \cellcolor[RGB]{106,150,246}{\scriptsize 35.2} & \cellcolor[RGB]{114,155,246}{\scriptsize 31.4} & \cellcolor[RGB]{104,148,246}{\scriptsize 36.2} & \cellcolor[RGB]{138,172,248}{\scriptsize 21.8\scalebox{0.55}{$\pm$14}} \\
& {\scriptsize Qwen3-4B} & \cellcolor[RGB]{230,237,253}{\scriptsize 1.0} & \cellcolor[RGB]{160,188,249}{\scriptsize 14.3} & \cellcolor[RGB]{125,163,247}{\scriptsize 26.7} & \cellcolor[RGB]{116,157,246}{\scriptsize 30.5} & \cellcolor[RGB]{96,143,245}{\scriptsize 40.0} & \cellcolor[RGB]{96,143,245}{\scriptsize 40.0} & \cellcolor[RGB]{200,216,251}{\scriptsize 4.8} & \cellcolor[RGB]{157,186,249}{\scriptsize 15.2} & \cellcolor[RGB]{163,190,249}{\scriptsize 13.3} & \cellcolor[RGB]{92,140,245}{\scriptsize 41.9} & \cellcolor[RGB]{181,203,250}{\scriptsize 8.6} & \cellcolor[RGB]{96,143,245}{\scriptsize 40.0} & \cellcolor[RGB]{50,110,242}{\scriptsize 66.7} & \cellcolor[RGB]{42,104,242}{\scriptsize 72.4} & \cellcolor[RGB]{67,122,243}{\scriptsize 56.2} & \cellcolor[RGB]{46,107,242}{\scriptsize 69.5} & \cellcolor[RGB]{109,152,246}{\scriptsize 33.8\scalebox{0.55}{$\pm$23}} \\
& {\scriptsize Qwen3-8B} & \cellcolor[RGB]{230,237,253}{\scriptsize 1.0} & \cellcolor[RGB]{163,190,249}{\scriptsize 13.3} & \cellcolor[RGB]{121,160,247}{\scriptsize 28.6} & \cellcolor[RGB]{110,153,246}{\scriptsize 33.3} & \cellcolor[RGB]{98,144,245}{\scriptsize 39.0} & \cellcolor[RGB]{92,140,245}{\scriptsize 41.9} & \cellcolor[RGB]{195,212,251}{\scriptsize 5.7} & \cellcolor[RGB]{163,190,249}{\scriptsize 13.3} & \cellcolor[RGB]{170,195,250}{\scriptsize 11.4} & \cellcolor[RGB]{96,143,245}{\scriptsize 40.0} & \cellcolor[RGB]{173,197,250}{\scriptsize 10.5} & \cellcolor[RGB]{89,138,245}{\scriptsize 43.8} & \cellcolor[RGB]{39,102,242}{\scriptsize 74.3} & \cellcolor[RGB]{40,103,242}{\scriptsize 73.3} & \cellcolor[RGB]{58,115,243}{\scriptsize 61.9} & \cellcolor[RGB]{38,102,242}{\scriptsize 81.0} & \cellcolor[RGB]{105,149,246}{\scriptsize 35.8\scalebox{0.55}{$\pm$25}} \\
& {\scriptsize Qwen3-14B} & \cellcolor[RGB]{230,237,253}{\scriptsize 1.0} & \cellcolor[RGB]{177,200,250}{\scriptsize 9.5} & \cellcolor[RGB]{116,157,246}{\scriptsize 30.5} & \cellcolor[RGB]{106,150,246}{\scriptsize 35.2} & \cellcolor[RGB]{98,144,245}{\scriptsize 39.0} & \cellcolor[RGB]{89,138,245}{\scriptsize 43.8} & \cellcolor[RGB]{212,225,252}{\scriptsize 2.9} & \cellcolor[RGB]{163,190,249}{\scriptsize 13.3} & \cellcolor[RGB]{154,183,249}{\scriptsize 16.2} & \cellcolor[RGB]{91,139,245}{\scriptsize 42.9} & \cellcolor[RGB]{170,195,250}{\scriptsize 11.4} & \cellcolor[RGB]{85,135,245}{\scriptsize 45.7} & \cellcolor[RGB]{38,102,242}{\scriptsize 77.1} & \cellcolor[RGB]{38,102,242}{\scriptsize 87.6} & \cellcolor[RGB]{40,103,242}{\scriptsize 73.3} & \cellcolor[RGB]{38,102,242}{\scriptsize 87.6} & \cellcolor[RGB]{99,145,245}{\scriptsize 38.6\scalebox{0.55}{$\pm$29}} \\
& {\scriptsize Qwen3-32B} & \cellcolor[RGB]{230,237,253}{\scriptsize 1.0} & \cellcolor[RGB]{163,190,249}{\scriptsize 13.3} & \cellcolor[RGB]{114,155,246}{\scriptsize 31.4} & \cellcolor[RGB]{102,147,246}{\scriptsize 37.1} & \cellcolor[RGB]{98,144,245}{\scriptsize 39.0} & \cellcolor[RGB]{80,131,244}{\scriptsize 48.6} & \cellcolor[RGB]{190,209,251}{\scriptsize 6.7} & \cellcolor[RGB]{163,190,249}{\scriptsize 13.3} & \cellcolor[RGB]{163,190,249}{\scriptsize 13.3} & \cellcolor[RGB]{82,133,244}{\scriptsize 47.6} & \cellcolor[RGB]{170,195,250}{\scriptsize 11.4} & \cellcolor[RGB]{91,139,245}{\scriptsize 42.9} & \cellcolor[RGB]{38,102,242}{\scriptsize 77.1} & \cellcolor[RGB]{38,102,242}{\scriptsize 89.5} & \cellcolor[RGB]{40,103,242}{\scriptsize 73.3} & \cellcolor[RGB]{38,102,242}{\scriptsize 93.3} & \cellcolor[RGB]{96,143,245}{\scriptsize 39.9\scalebox{0.55}{$\pm$29}} \\
& {\scriptsize Gemma-3-4B} & \cellcolor[RGB]{230,237,253}{\scriptsize 1.0} & \cellcolor[RGB]{173,197,250}{\scriptsize 10.5} & \cellcolor[RGB]{125,163,247}{\scriptsize 26.7} & \cellcolor[RGB]{123,162,247}{\scriptsize 27.6} & \cellcolor[RGB]{116,157,246}{\scriptsize 30.5} & \cellcolor[RGB]{121,160,247}{\scriptsize 28.6} & \cellcolor[RGB]{195,212,251}{\scriptsize 5.7} & \cellcolor[RGB]{166,192,249}{\scriptsize 12.4} & \cellcolor[RGB]{181,203,250}{\scriptsize 8.6} & \cellcolor[RGB]{98,144,245}{\scriptsize 39.0} & \cellcolor[RGB]{181,203,250}{\scriptsize 8.6} & \cellcolor[RGB]{106,150,246}{\scriptsize 35.2} & \cellcolor[RGB]{70,124,244}{\scriptsize 54.3} & \cellcolor[RGB]{82,133,244}{\scriptsize 47.6} & \cellcolor[RGB]{104,148,246}{\scriptsize 36.2} & \cellcolor[RGB]{80,131,244}{\scriptsize 48.6} & \cellcolor[RGB]{126,164,247}{\scriptsize 26.3\scalebox{0.55}{$\pm$16}} \\
& {\scriptsize Gemma-3-12B} & \cellcolor[RGB]{230,237,253}{\scriptsize 1.0} & \cellcolor[RGB]{166,192,249}{\scriptsize 12.4} & \cellcolor[RGB]{123,162,247}{\scriptsize 27.6} & \cellcolor[RGB]{116,157,246}{\scriptsize 30.5} & \cellcolor[RGB]{100,145,245}{\scriptsize 38.1} & \cellcolor[RGB]{104,148,246}{\scriptsize 36.2} & \cellcolor[RGB]{190,209,251}{\scriptsize 6.7} & \cellcolor[RGB]{163,190,249}{\scriptsize 13.3} & \cellcolor[RGB]{173,197,250}{\scriptsize 10.5} & \cellcolor[RGB]{94,141,245}{\scriptsize 41.0} & \cellcolor[RGB]{181,203,250}{\scriptsize 8.6} & \cellcolor[RGB]{85,135,245}{\scriptsize 45.7} & \cellcolor[RGB]{43,105,242}{\scriptsize 71.4} & \cellcolor[RGB]{53,112,243}{\scriptsize 64.8} & \cellcolor[RGB]{67,122,243}{\scriptsize 56.2} & \cellcolor[RGB]{46,107,242}{\scriptsize 69.5} & \cellcolor[RGB]{110,153,246}{\scriptsize 33.3\scalebox{0.55}{$\pm$23}} \\
& {\scriptsize Gemma-3-27B} & \cellcolor[RGB]{230,237,253}{\scriptsize 1.0} & \cellcolor[RGB]{170,195,250}{\scriptsize 11.4} & \cellcolor[RGB]{114,155,246}{\scriptsize 31.4} & \cellcolor[RGB]{104,148,246}{\scriptsize 36.2} & \cellcolor[RGB]{96,143,245}{\scriptsize 40.0} & \cellcolor[RGB]{91,139,245}{\scriptsize 42.9} & \cellcolor[RGB]{190,209,251}{\scriptsize 6.7} & \cellcolor[RGB]{163,190,249}{\scriptsize 13.3} & \cellcolor[RGB]{154,183,249}{\scriptsize 16.2} & \cellcolor[RGB]{89,138,245}{\scriptsize 43.8} & \cellcolor[RGB]{177,200,250}{\scriptsize 9.5} & \cellcolor[RGB]{91,139,245}{\scriptsize 42.9} & \cellcolor[RGB]{38,102,242}{\scriptsize 77.1} & \cellcolor[RGB]{38,102,242}{\scriptsize 84.8} & \cellcolor[RGB]{49,109,242}{\scriptsize 67.6} & \cellcolor[RGB]{38,102,242}{\scriptsize 84.8} & \cellcolor[RGB]{100,145,245}{\scriptsize 38.1\scalebox{0.55}{$\pm$27}} \\
& {\scriptsize Phi-4} & \cellcolor[RGB]{230,237,253}{\scriptsize 1.0} & \cellcolor[RGB]{163,190,249}{\scriptsize 13.3} & \cellcolor[RGB]{114,155,246}{\scriptsize 31.4} & \cellcolor[RGB]{110,153,246}{\scriptsize 33.3} & \cellcolor[RGB]{98,144,245}{\scriptsize 39.0} & \cellcolor[RGB]{91,139,245}{\scriptsize 42.9} & \cellcolor[RGB]{195,212,251}{\scriptsize 5.7} & \cellcolor[RGB]{160,188,249}{\scriptsize 14.3} & \cellcolor[RGB]{160,188,249}{\scriptsize 14.3} & \cellcolor[RGB]{89,138,245}{\scriptsize 43.8} & \cellcolor[RGB]{173,197,250}{\scriptsize 10.5} & \cellcolor[RGB]{78,130,244}{\scriptsize 49.5} & \cellcolor[RGB]{38,102,242}{\scriptsize 75.2} & \cellcolor[RGB]{38,102,242}{\scriptsize 85.7} & \cellcolor[RGB]{50,110,242}{\scriptsize 66.7} & \cellcolor[RGB]{38,102,242}{\scriptsize 94.3} & \cellcolor[RGB]{99,144,245}{\scriptsize 38.8\scalebox{0.55}{$\pm$28}} \\
\cmidrule[0.3pt](l){2-19}
& {\scriptsize Dream-7B} & \cellcolor[RGB]{230,237,253}{\scriptsize 1.0} & \cellcolor[RGB]{177,200,250}{\scriptsize 9.5} & \cellcolor[RGB]{140,174,248}{\scriptsize 21.0} & \cellcolor[RGB]{137,172,248}{\scriptsize 21.9} & \cellcolor[RGB]{137,172,248}{\scriptsize 21.9} & \cellcolor[RGB]{137,172,248}{\scriptsize 21.9} & \cellcolor[RGB]{206,220,252}{\scriptsize 3.8} & \cellcolor[RGB]{173,197,250}{\scriptsize 10.5} & \cellcolor[RGB]{195,212,251}{\scriptsize 5.7} & \cellcolor[RGB]{125,163,247}{\scriptsize 26.7} & \cellcolor[RGB]{200,216,251}{\scriptsize 4.8} & \cellcolor[RGB]{114,155,246}{\scriptsize 31.4} & \cellcolor[RGB]{100,145,245}{\scriptsize 38.1} & \cellcolor[RGB]{106,150,246}{\scriptsize 35.2} & \cellcolor[RGB]{91,139,245}{\scriptsize 42.9} & \cellcolor[RGB]{78,130,244}{\scriptsize 49.5} & \cellcolor[RGB]{138,172,248}{\scriptsize 21.6\scalebox{0.55}{$\pm$14}} \\
& {\scriptsize LLaDA-8B} & \cellcolor[RGB]{230,237,253}{\scriptsize 1.0} & \cellcolor[RGB]{170,195,250}{\scriptsize 11.4} & \cellcolor[RGB]{123,162,247}{\scriptsize 27.6} & \cellcolor[RGB]{121,160,247}{\scriptsize 28.6} & \cellcolor[RGB]{132,168,247}{\scriptsize 23.8} & \cellcolor[RGB]{110,153,246}{\scriptsize 33.3} & \cellcolor[RGB]{220,230,252}{\scriptsize 1.9} & \cellcolor[RGB]{160,188,249}{\scriptsize 14.3} & \cellcolor[RGB]{185,206,250}{\scriptsize 7.6} & \cellcolor[RGB]{110,153,246}{\scriptsize 33.3} & \cellcolor[RGB]{170,195,250}{\scriptsize 11.4} & \cellcolor[RGB]{98,144,245}{\scriptsize 39.0} & \cellcolor[RGB]{59,117,243}{\scriptsize 61.0} & \cellcolor[RGB]{46,107,242}{\scriptsize 69.5} & \cellcolor[RGB]{61,118,243}{\scriptsize 60.0} & \cellcolor[RGB]{73,127,244}{\scriptsize 52.4} & \cellcolor[RGB]{118,158,246}{\scriptsize 29.8\scalebox{0.55}{$\pm$21}} \\
\cmidrule[0.3pt](l){2-19}
& {\scriptsize C-Haiku-4.5} & \cellcolor[RGB]{230,237,253}{\scriptsize 1.0} & \cellcolor[RGB]{160,188,249}{\scriptsize 14.3} & \cellcolor[RGB]{112,154,246}{\scriptsize 32.4} & \cellcolor[RGB]{102,147,246}{\scriptsize 37.1} & \cellcolor[RGB]{96,143,245}{\scriptsize 40.0} & \cellcolor[RGB]{82,133,244}{\scriptsize 47.6} & \cellcolor[RGB]{185,206,250}{\scriptsize 7.6} & \cellcolor[RGB]{166,192,249}{\scriptsize 12.4} & \cellcolor[RGB]{154,183,249}{\scriptsize 16.2} & \cellcolor[RGB]{89,138,245}{\scriptsize 43.8} & \cellcolor[RGB]{170,195,250}{\scriptsize 11.4} & \cellcolor[RGB]{85,135,245}{\scriptsize 45.7} & \cellcolor[RGB]{38,102,242}{\scriptsize 76.2} & \cellcolor[RGB]{38,102,242}{\scriptsize 95.2} & \cellcolor[RGB]{38,102,242}{\scriptsize 75.2} & \cellcolor[RGB]{38,102,242}{\scriptsize 91.4} & \cellcolor[RGB]{95,142,245}{\scriptsize 40.5\scalebox{0.55}{$\pm$29}} \\
& {\scriptsize GPT-5} & \cellcolor[RGB]{230,237,253}{\scriptsize 1.0} & \cellcolor[RGB]{160,188,249}{\scriptsize 14.3} & \cellcolor[RGB]{121,160,247}{\scriptsize 28.6} & \cellcolor[RGB]{102,147,246}{\scriptsize 37.1} & \cellcolor[RGB]{98,144,245}{\scriptsize 39.0} & \cellcolor[RGB]{94,141,245}{\scriptsize 41.0} & \cellcolor[RGB]{195,212,251}{\scriptsize 5.7} & \cellcolor[RGB]{154,183,249}{\scriptsize 16.2} & \cellcolor[RGB]{173,197,250}{\scriptsize 10.5} & \cellcolor[RGB]{91,139,245}{\scriptsize 42.9} & \cellcolor[RGB]{170,195,250}{\scriptsize 11.4} & \cellcolor[RGB]{84,134,244}{\scriptsize 46.7} & \cellcolor[RGB]{39,102,242}{\scriptsize 74.3} & \cellcolor[RGB]{38,102,242}{\scriptsize 95.2} & \cellcolor[RGB]{40,103,242}{\scriptsize 73.3} & \cellcolor[RGB]{38,102,242}{\scriptsize 94.3} & \cellcolor[RGB]{97,144,245}{\scriptsize 39.5\scalebox{0.55}{$\pm$30}} \\
& {\scriptsize G-3-Flash} & \cellcolor[RGB]{230,237,253}{\scriptsize 1.0} & \cellcolor[RGB]{160,188,249}{\scriptsize 14.3} & \cellcolor[RGB]{119,159,247}{\scriptsize 29.5} & \cellcolor[RGB]{98,144,245}{\scriptsize 39.0} & \cellcolor[RGB]{98,144,245}{\scriptsize 39.0} & \cellcolor[RGB]{91,139,245}{\scriptsize 42.9} & \cellcolor[RGB]{185,206,250}{\scriptsize 7.6} & \cellcolor[RGB]{163,190,249}{\scriptsize 13.3} & \cellcolor[RGB]{173,197,250}{\scriptsize 10.5} & \cellcolor[RGB]{85,135,245}{\scriptsize 45.7} & \cellcolor[RGB]{173,197,250}{\scriptsize 10.5} & \cellcolor[RGB]{80,131,244}{\scriptsize 48.6} & \cellcolor[RGB]{44,106,242}{\scriptsize 70.5} & \cellcolor[RGB]{38,102,242}{\scriptsize 95.2} & \cellcolor[RGB]{44,106,242}{\scriptsize 70.5} & \cellcolor[RGB]{38,102,242}{\scriptsize 97.1} & \cellcolor[RGB]{97,143,245}{\scriptsize 39.7\scalebox{0.55}{$\pm$30}} \\
& {\scriptsize G-3.1-Pro} & \cellcolor[RGB]{230,237,253}{\scriptsize 1.0} & \cellcolor[RGB]{160,188,249}{\scriptsize 14.3} & \cellcolor[RGB]{121,160,247}{\scriptsize 28.6} & \cellcolor[RGB]{106,150,246}{\scriptsize 35.2} & \cellcolor[RGB]{106,150,246}{\scriptsize 35.2} & \cellcolor[RGB]{110,153,246}{\scriptsize 33.3} & \cellcolor[RGB]{190,209,251}{\scriptsize 6.7} & \cellcolor[RGB]{151,181,248}{\scriptsize 17.1} & \cellcolor[RGB]{166,192,249}{\scriptsize 12.4} & \cellcolor[RGB]{82,133,244}{\scriptsize 47.6} & \cellcolor[RGB]{166,192,249}{\scriptsize 12.4} & \cellcolor[RGB]{77,129,244}{\scriptsize 50.5} & \cellcolor[RGB]{39,102,242}{\scriptsize 74.3} & \cellcolor[RGB]{38,102,242}{\scriptsize 100.0} & \cellcolor[RGB]{38,102,242}{\scriptsize 77.1} & \cellcolor[RGB]{38,102,242}{\scriptsize 98.1} & \cellcolor[RGB]{96,142,245}{\scriptsize 40.2\scalebox{0.55}{$\pm$31}} \\
\cmidrule[0.5pt](l){2-19}
& {\scriptsize Gen.\ quality} & \cellcolor[RGB]{230,237,253}{\scriptsize 1.0\scalebox{0.55}{$\pm$0}} & \cellcolor[RGB]{168,193,249}{\scriptsize 12.0\scalebox{0.55}{$\pm$2}} & \cellcolor[RGB]{124,162,247}{\scriptsize 27.3\scalebox{0.55}{$\pm$5}} & \cellcolor[RGB]{114,155,246}{\scriptsize 31.7\scalebox{0.55}{$\pm$6}} & \cellcolor[RGB]{108,151,246}{\scriptsize 34.3\scalebox{0.55}{$\pm$8}} & \cellcolor[RGB]{104,148,246}{\scriptsize 36.2\scalebox{0.55}{$\pm$10}} & \cellcolor[RGB]{198,215,251}{\scriptsize 5.1\scalebox{0.55}{$\pm$2}} & \cellcolor[RGB]{164,191,249}{\scriptsize 13.0\scalebox{0.55}{$\pm$3}} & \cellcolor[RGB]{172,196,250}{\scriptsize 11.0\scalebox{0.55}{$\pm$4}} & \cellcolor[RGB]{98,144,245}{\scriptsize 39.0\scalebox{0.55}{$\pm$8}} & \cellcolor[RGB]{178,201,250}{\scriptsize 9.3\scalebox{0.55}{$\pm$3}} & \cellcolor[RGB]{94,141,245}{\scriptsize 41.0\scalebox{0.55}{$\pm$8}} & \cellcolor[RGB]{53,113,243}{\scriptsize 64.6\scalebox{0.55}{$\pm$16}} & \cellcolor[RGB]{43,105,242}{\scriptsize 71.2\scalebox{0.55}{$\pm$26}} & \cellcolor[RGB]{63,119,243}{\scriptsize 58.4\scalebox{0.55}{$\pm$18}} & \cellcolor[RGB]{42,104,242}{\scriptsize 72.2\scalebox{0.55}{$\pm$26}} &  \\
\bottomrule
\end{tabular}
}%  end resizebox
\end{table}

\begin{table}[htb]
\centering
\caption{Communication matrix at depth $d = 4$ (840 expressions). Each cell shows round-trip accuracy (\%). Rows are extractors, columns are generators. Dotted lines separate open-weight AR, diffusion, and frontier models.}
\vspace{2mm}
\label{tab:matrix_d4}
\small
\setlength{\tabcolsep}{2pt}
\setlength{\dashlinedash}{1pt}
\setlength{\dashlinegap}{1pt}
\resizebox{\textwidth}{!}{%
\begin{tabular}{c l c c c c c c c c c c !{\;\vrule width 0.3pt\;} c c !{\;\vrule width 0.3pt\;} c c c c !{\;\vrule width 0.75pt\;} c}
\toprule
\multicolumn{19}{c}{\textsc{Generators}} \\
 \multirow{20}{*}{\rotatebox{90}{\textsc{Extractors}}} & & \rotatebox{90}{\scriptsize Qwen3-0.6B} & \rotatebox{90}{\scriptsize Qwen3-1.7B} & \rotatebox{90}{\scriptsize Qwen3-4B} & \rotatebox{90}{\scriptsize Qwen3-8B} & \rotatebox{90}{\scriptsize Qwen3-14B} & \rotatebox{90}{\scriptsize Qwen3-32B} & \rotatebox{90}{\scriptsize Gemma-3-4B} & \rotatebox{90}{\scriptsize Gemma-3-12B} & \rotatebox{90}{\scriptsize Gemma-3-27B} & \rotatebox{90}{\scriptsize Phi-4} & \rotatebox{90}{\scriptsize Dream-7B} & \rotatebox{90}{\scriptsize LLaDA-8B} & \rotatebox{90}{\scriptsize C-Haiku-4.5} & \rotatebox{90}{\scriptsize GPT-5} & \rotatebox{90}{\scriptsize G-3-Flash} & \rotatebox{90}{\scriptsize G-3.1-Pro} & \rotatebox{90}{\scriptsize Extr.\ quality} \\
\cmidrule[0.5pt](l){2-19}
& {\scriptsize $\Guard$ pass \%} & \cellcolor[RGB]{230,247,230}{\scriptsize 19} & \cellcolor[RGB]{182,233,182}{\scriptsize 57} & \cellcolor[RGB]{134,218,134}{\scriptsize 94} & \cellcolor[RGB]{130,217,130}{\scriptsize 98} & \cellcolor[RGB]{128,217,128}{\scriptsize 99} & \cellcolor[RGB]{130,217,130}{\scriptsize 98} & \cellcolor[RGB]{130,217,130}{\scriptsize 98} & \cellcolor[RGB]{132,218,132}{\scriptsize 96} & \cellcolor[RGB]{134,218,134}{\scriptsize 95} & \cellcolor[RGB]{132,218,132}{\scriptsize 96} & \cellcolor[RGB]{172,230,172}{\scriptsize 64} & \cellcolor[RGB]{134,218,134}{\scriptsize 95} & \cellcolor[RGB]{127,216,127}{\scriptsize 100} & \cellcolor[RGB]{127,216,127}{\scriptsize 100} & \cellcolor[RGB]{127,216,127}{\scriptsize 100} & \cellcolor[RGB]{127,216,127}{\scriptsize 100} &  \\
\cmidrule[0.5pt](l){2-19}
& {\scriptsize Qwen3-0.6B} & \cellcolor[RGB]{255,255,255}{\scriptsize 0.0} & \cellcolor[RGB]{223,233,253}{\scriptsize 1.5} & \cellcolor[RGB]{201,216,251}{\scriptsize 4.6} & \cellcolor[RGB]{196,213,251}{\scriptsize 5.5} & \cellcolor[RGB]{203,218,251}{\scriptsize 4.2} & \cellcolor[RGB]{218,229,252}{\scriptsize 2.1} & \cellcolor[RGB]{246,248,254}{\scriptsize 0.1} & \cellcolor[RGB]{237,242,253}{\scriptsize 0.5} & \cellcolor[RGB]{240,244,254}{\scriptsize 0.4} & \cellcolor[RGB]{195,212,251}{\scriptsize 5.7} & \cellcolor[RGB]{232,238,253}{\scriptsize 0.8} & \cellcolor[RGB]{195,213,251}{\scriptsize 5.6} & \cellcolor[RGB]{180,202,250}{\scriptsize 8.8} & \cellcolor[RGB]{216,227,252}{\scriptsize 2.4} & \cellcolor[RGB]{215,227,252}{\scriptsize 2.5} & \cellcolor[RGB]{219,229,252}{\scriptsize 2.0} & \cellcolor[RGB]{212,224,252}{\scriptsize 2.9\scalebox{0.55}{$\pm$2}} \\
& {\scriptsize Qwen3-1.7B} & \cellcolor[RGB]{255,255,255}{\scriptsize 0.0} & \cellcolor[RGB]{223,233,253}{\scriptsize 1.5} & \cellcolor[RGB]{176,199,250}{\scriptsize 9.8} & \cellcolor[RGB]{161,189,249}{\scriptsize 13.8} & \cellcolor[RGB]{166,192,249}{\scriptsize 12.6} & \cellcolor[RGB]{181,203,250}{\scriptsize 8.6} & \cellcolor[RGB]{240,244,254}{\scriptsize 0.4} & \cellcolor[RGB]{219,229,252}{\scriptsize 2.0} & \cellcolor[RGB]{230,237,253}{\scriptsize 1.0} & \cellcolor[RGB]{152,182,248}{\scriptsize 16.7} & \cellcolor[RGB]{225,233,253}{\scriptsize 1.4} & \cellcolor[RGB]{171,195,250}{\scriptsize 11.2} & \cellcolor[RGB]{123,162,247}{\scriptsize 27.6} & \cellcolor[RGB]{163,190,249}{\scriptsize 13.5} & \cellcolor[RGB]{158,186,249}{\scriptsize 14.9} & \cellcolor[RGB]{154,184,249}{\scriptsize 16.1} & \cellcolor[RGB]{178,200,250}{\scriptsize 9.4\scalebox{0.55}{$\pm$8}} \\
& {\scriptsize Qwen3-4B} & \cellcolor[RGB]{255,255,255}{\scriptsize 0.0} & \cellcolor[RGB]{218,229,252}{\scriptsize 2.1} & \cellcolor[RGB]{156,185,249}{\scriptsize 15.6} & \cellcolor[RGB]{127,164,247}{\scriptsize 26.0} & \cellcolor[RGB]{133,169,247}{\scriptsize 23.6} & \cellcolor[RGB]{141,175,248}{\scriptsize 20.5} & \cellcolor[RGB]{235,241,253}{\scriptsize 0.6} & \cellcolor[RGB]{208,222,252}{\scriptsize 3.5} & \cellcolor[RGB]{213,225,252}{\scriptsize 2.7} & \cellcolor[RGB]{120,159,247}{\scriptsize 29.0} & \cellcolor[RGB]{215,227,252}{\scriptsize 2.5} & \cellcolor[RGB]{148,179,248}{\scriptsize 18.2} & \cellcolor[RGB]{66,122,243}{\scriptsize 56.4} & \cellcolor[RGB]{90,139,245}{\scriptsize 43.0} & \cellcolor[RGB]{98,144,245}{\scriptsize 39.0} & \cellcolor[RGB]{67,122,243}{\scriptsize 55.8} & \cellcolor[RGB]{139,173,248}{\scriptsize 21.2\scalebox{0.55}{$\pm$19}} \\
& {\scriptsize Qwen3-8B} & \cellcolor[RGB]{255,255,255}{\scriptsize 0.0} & \cellcolor[RGB]{217,228,252}{\scriptsize 2.3} & \cellcolor[RGB]{155,184,249}{\scriptsize 15.7} & \cellcolor[RGB]{127,164,247}{\scriptsize 26.0} & \cellcolor[RGB]{131,168,247}{\scriptsize 24.2} & \cellcolor[RGB]{136,171,248}{\scriptsize 22.3} & \cellcolor[RGB]{233,240,253}{\scriptsize 0.7} & \cellcolor[RGB]{208,222,252}{\scriptsize 3.5} & \cellcolor[RGB]{206,220,252}{\scriptsize 3.8} & \cellcolor[RGB]{112,154,246}{\scriptsize 32.3} & \cellcolor[RGB]{213,225,252}{\scriptsize 2.7} & \cellcolor[RGB]{147,179,248}{\scriptsize 18.3} & \cellcolor[RGB]{68,123,244}{\scriptsize 55.5} & \cellcolor[RGB]{78,130,244}{\scriptsize 49.8} & \cellcolor[RGB]{86,136,245}{\scriptsize 45.4} & \cellcolor[RGB]{53,112,243}{\scriptsize 64.9} & \cellcolor[RGB]{135,170,247}{\scriptsize 22.9\scalebox{0.55}{$\pm$21}} \\
& {\scriptsize Qwen3-14B} & \cellcolor[RGB]{255,255,255}{\scriptsize 0.0} & \cellcolor[RGB]{212,225,252}{\scriptsize 2.9} & \cellcolor[RGB]{147,179,248}{\scriptsize 18.5} & \cellcolor[RGB]{118,158,246}{\scriptsize 29.6} & \cellcolor[RGB]{121,160,247}{\scriptsize 28.6} & \cellcolor[RGB]{128,166,247}{\scriptsize 25.4} & \cellcolor[RGB]{235,241,253}{\scriptsize 0.6} & \cellcolor[RGB]{206,220,252}{\scriptsize 3.8} & \cellcolor[RGB]{203,218,251}{\scriptsize 4.2} & \cellcolor[RGB]{105,149,246}{\scriptsize 35.7} & \cellcolor[RGB]{211,224,252}{\scriptsize 3.0} & \cellcolor[RGB]{136,171,248}{\scriptsize 22.5} & \cellcolor[RGB]{55,114,243}{\scriptsize 63.6} & \cellcolor[RGB]{53,112,243}{\scriptsize 65.0} & \cellcolor[RGB]{69,124,244}{\scriptsize 54.6} & \cellcolor[RGB]{38,102,242}{\scriptsize 78.0} & \cellcolor[RGB]{124,162,247}{\scriptsize 27.2\scalebox{0.55}{$\pm$25}} \\
& {\scriptsize Qwen3-32B} & \cellcolor[RGB]{255,255,255}{\scriptsize 0.0} & \cellcolor[RGB]{211,224,252}{\scriptsize 3.0} & \cellcolor[RGB]{147,179,248}{\scriptsize 18.3} & \cellcolor[RGB]{119,159,247}{\scriptsize 29.2} & \cellcolor[RGB]{117,158,246}{\scriptsize 30.0} & \cellcolor[RGB]{127,164,247}{\scriptsize 26.1} & \cellcolor[RGB]{233,240,253}{\scriptsize 0.7} & \cellcolor[RGB]{201,216,251}{\scriptsize 4.6} & \cellcolor[RGB]{204,219,252}{\scriptsize 4.0} & \cellcolor[RGB]{108,151,246}{\scriptsize 34.3} & \cellcolor[RGB]{207,221,252}{\scriptsize 3.6} & \cellcolor[RGB]{135,170,247}{\scriptsize 22.9} & \cellcolor[RGB]{59,116,243}{\scriptsize 61.3} & \cellcolor[RGB]{42,105,242}{\scriptsize 71.9} & \cellcolor[RGB]{67,122,243}{\scriptsize 55.8} & \cellcolor[RGB]{38,102,242}{\scriptsize 80.7} & \cellcolor[RGB]{122,161,247}{\scriptsize 27.9\scalebox{0.55}{$\pm$26}} \\
& {\scriptsize Gemma-3-4B} & \cellcolor[RGB]{255,255,255}{\scriptsize 0.0} & \cellcolor[RGB]{220,230,252}{\scriptsize 1.9} & \cellcolor[RGB]{165,191,249}{\scriptsize 12.7} & \cellcolor[RGB]{151,181,248}{\scriptsize 17.3} & \cellcolor[RGB]{154,184,249}{\scriptsize 16.1} & \cellcolor[RGB]{166,192,249}{\scriptsize 12.4} & \cellcolor[RGB]{235,241,253}{\scriptsize 0.6} & \cellcolor[RGB]{209,222,252}{\scriptsize 3.3} & \cellcolor[RGB]{222,232,253}{\scriptsize 1.7} & \cellcolor[RGB]{138,172,248}{\scriptsize 21.8} & \cellcolor[RGB]{223,233,253}{\scriptsize 1.5} & \cellcolor[RGB]{160,187,249}{\scriptsize 14.4} & \cellcolor[RGB]{107,150,246}{\scriptsize 34.9} & \cellcolor[RGB]{145,177,248}{\scriptsize 19.2} & \cellcolor[RGB]{137,172,248}{\scriptsize 22.0} & \cellcolor[RGB]{134,169,247}{\scriptsize 23.3} & \cellcolor[RGB]{165,192,249}{\scriptsize 12.7\scalebox{0.55}{$\pm$10}} \\
& {\scriptsize Gemma-3-12B} & \cellcolor[RGB]{255,255,255}{\scriptsize 0.0} & \cellcolor[RGB]{217,228,252}{\scriptsize 2.3} & \cellcolor[RGB]{156,185,249}{\scriptsize 15.5} & \cellcolor[RGB]{129,166,247}{\scriptsize 25.0} & \cellcolor[RGB]{134,169,247}{\scriptsize 23.3} & \cellcolor[RGB]{142,175,248}{\scriptsize 20.4} & \cellcolor[RGB]{237,242,253}{\scriptsize 0.5} & \cellcolor[RGB]{204,219,252}{\scriptsize 4.0} & \cellcolor[RGB]{214,226,252}{\scriptsize 2.6} & \cellcolor[RGB]{122,161,247}{\scriptsize 28.2} & \cellcolor[RGB]{211,224,252}{\scriptsize 3.0} & \cellcolor[RGB]{148,179,248}{\scriptsize 18.2} & \cellcolor[RGB]{75,128,244}{\scriptsize 51.4} & \cellcolor[RGB]{85,135,245}{\scriptsize 46.1} & \cellcolor[RGB]{100,145,245}{\scriptsize 38.2} & \cellcolor[RGB]{69,123,244}{\scriptsize 55.1} & \cellcolor[RGB]{140,174,248}{\scriptsize 20.9\scalebox{0.55}{$\pm$18}} \\
& {\scriptsize Gemma-3-27B} & \cellcolor[RGB]{255,255,255}{\scriptsize 0.0} & \cellcolor[RGB]{216,227,252}{\scriptsize 2.4} & \cellcolor[RGB]{148,179,248}{\scriptsize 18.2} & \cellcolor[RGB]{123,162,247}{\scriptsize 27.5} & \cellcolor[RGB]{122,161,247}{\scriptsize 27.9} & \cellcolor[RGB]{129,166,247}{\scriptsize 25.2} & \cellcolor[RGB]{230,237,253}{\scriptsize 1.0} & \cellcolor[RGB]{204,219,252}{\scriptsize 4.0} & \cellcolor[RGB]{206,220,252}{\scriptsize 3.8} & \cellcolor[RGB]{109,152,246}{\scriptsize 33.9} & \cellcolor[RGB]{209,222,252}{\scriptsize 3.3} & \cellcolor[RGB]{142,175,248}{\scriptsize 20.2} & \cellcolor[RGB]{61,118,243}{\scriptsize 59.8} & \cellcolor[RGB]{55,114,243}{\scriptsize 63.5} & \cellcolor[RGB]{81,132,244}{\scriptsize 48.1} & \cellcolor[RGB]{38,102,242}{\scriptsize 77.5} & \cellcolor[RGB]{127,164,247}{\scriptsize 26.0\scalebox{0.55}{$\pm$24}} \\
& {\scriptsize Phi-4} & \cellcolor[RGB]{255,255,255}{\scriptsize 0.0} & \cellcolor[RGB]{217,228,252}{\scriptsize 2.3} & \cellcolor[RGB]{145,177,248}{\scriptsize 19.2} & \cellcolor[RGB]{120,160,247}{\scriptsize 28.7} & \cellcolor[RGB]{119,159,247}{\scriptsize 29.5} & \cellcolor[RGB]{126,164,247}{\scriptsize 26.3} & \cellcolor[RGB]{232,238,253}{\scriptsize 0.8} & \cellcolor[RGB]{203,218,251}{\scriptsize 4.3} & \cellcolor[RGB]{200,216,251}{\scriptsize 4.8} & \cellcolor[RGB]{106,149,246}{\scriptsize 35.4} & \cellcolor[RGB]{210,223,252}{\scriptsize 3.2} & \cellcolor[RGB]{137,172,248}{\scriptsize 22.0} & \cellcolor[RGB]{57,115,243}{\scriptsize 62.5} & \cellcolor[RGB]{43,105,242}{\scriptsize 71.5} & \cellcolor[RGB]{69,124,244}{\scriptsize 54.6} & \cellcolor[RGB]{38,102,242}{\scriptsize 83.8} & \cellcolor[RGB]{122,161,247}{\scriptsize 28.1\scalebox{0.55}{$\pm$26}} \\
\cmidrule[0.3pt](l){2-19}
& {\scriptsize Dream-7B} & \cellcolor[RGB]{255,255,255}{\scriptsize 0.0} & \cellcolor[RGB]{226,234,253}{\scriptsize 1.3} & \cellcolor[RGB]{180,202,250}{\scriptsize 8.9} & \cellcolor[RGB]{159,187,249}{\scriptsize 14.5} & \cellcolor[RGB]{163,190,249}{\scriptsize 13.5} & \cellcolor[RGB]{170,195,250}{\scriptsize 11.4} & \cellcolor[RGB]{242,246,254}{\scriptsize 0.2} & \cellcolor[RGB]{219,229,252}{\scriptsize 2.0} & \cellcolor[RGB]{219,229,252}{\scriptsize 2.0} & \cellcolor[RGB]{156,185,249}{\scriptsize 15.5} & \cellcolor[RGB]{227,235,253}{\scriptsize 1.2} & \cellcolor[RGB]{176,199,250}{\scriptsize 9.9} & \cellcolor[RGB]{128,165,247}{\scriptsize 25.6} & \cellcolor[RGB]{136,171,248}{\scriptsize 22.3} & \cellcolor[RGB]{135,170,247}{\scriptsize 23.0} & \cellcolor[RGB]{94,141,245}{\scriptsize 41.0} & \cellcolor[RGB]{168,193,249}{\scriptsize 12.0\scalebox{0.55}{$\pm$11}} \\
& {\scriptsize LLaDA-8B} & \cellcolor[RGB]{255,255,255}{\scriptsize 0.0} & \cellcolor[RGB]{220,230,252}{\scriptsize 1.9} & \cellcolor[RGB]{164,191,249}{\scriptsize 13.1} & \cellcolor[RGB]{138,172,248}{\scriptsize 21.8} & \cellcolor[RGB]{148,179,248}{\scriptsize 18.1} & \cellcolor[RGB]{156,185,249}{\scriptsize 15.5} & \cellcolor[RGB]{237,242,253}{\scriptsize 0.5} & \cellcolor[RGB]{209,222,252}{\scriptsize 3.3} & \cellcolor[RGB]{215,227,252}{\scriptsize 2.5} & \cellcolor[RGB]{131,167,247}{\scriptsize 24.3} & \cellcolor[RGB]{211,224,252}{\scriptsize 3.0} & \cellcolor[RGB]{152,182,248}{\scriptsize 16.9} & \cellcolor[RGB]{93,141,245}{\scriptsize 41.5} & \cellcolor[RGB]{88,137,245}{\scriptsize 44.2} & \cellcolor[RGB]{103,148,246}{\scriptsize 36.5} & \cellcolor[RGB]{83,134,244}{\scriptsize 46.8} & \cellcolor[RGB]{148,179,248}{\scriptsize 18.1\scalebox{0.55}{$\pm$16}} \\
\cmidrule[0.3pt](l){2-19}
& {\scriptsize C-Haiku-4.5} & \cellcolor[RGB]{255,255,255}{\scriptsize 0.0} & \cellcolor[RGB]{214,226,252}{\scriptsize 2.6} & \cellcolor[RGB]{148,179,248}{\scriptsize 18.1} & \cellcolor[RGB]{120,160,247}{\scriptsize 28.7} & \cellcolor[RGB]{119,159,247}{\scriptsize 29.5} & \cellcolor[RGB]{124,162,247}{\scriptsize 27.4} & \cellcolor[RGB]{233,240,253}{\scriptsize 0.7} & \cellcolor[RGB]{204,219,252}{\scriptsize 4.0} & \cellcolor[RGB]{203,218,251}{\scriptsize 4.2} & \cellcolor[RGB]{104,149,246}{\scriptsize 36.0} & \cellcolor[RGB]{207,221,252}{\scriptsize 3.6} & \cellcolor[RGB]{137,171,248}{\scriptsize 22.1} & \cellcolor[RGB]{56,114,243}{\scriptsize 63.1} & \cellcolor[RGB]{38,102,242}{\scriptsize 81.9} & \cellcolor[RGB]{60,117,243}{\scriptsize 60.6} & \cellcolor[RGB]{38,102,242}{\scriptsize 91.9} & \cellcolor[RGB]{118,158,246}{\scriptsize 29.7\scalebox{0.55}{$\pm$29}} \\
& {\scriptsize GPT-5} & \cellcolor[RGB]{255,255,255}{\scriptsize 0.0} & \cellcolor[RGB]{214,226,252}{\scriptsize 2.6} & \cellcolor[RGB]{147,178,248}{\scriptsize 18.6} & \cellcolor[RGB]{117,157,246}{\scriptsize 30.4} & \cellcolor[RGB]{116,157,246}{\scriptsize 30.7} & \cellcolor[RGB]{131,167,247}{\scriptsize 24.4} & \cellcolor[RGB]{237,242,253}{\scriptsize 0.5} & \cellcolor[RGB]{201,217,251}{\scriptsize 4.5} & \cellcolor[RGB]{203,218,251}{\scriptsize 4.2} & \cellcolor[RGB]{103,148,246}{\scriptsize 36.7} & \cellcolor[RGB]{210,223,252}{\scriptsize 3.2} & \cellcolor[RGB]{131,167,247}{\scriptsize 24.5} & \cellcolor[RGB]{52,112,243}{\scriptsize 65.4} & \cellcolor[RGB]{38,102,242}{\scriptsize 91.3} & \cellcolor[RGB]{50,110,242}{\scriptsize 66.8} & \cellcolor[RGB]{38,102,242}{\scriptsize 95.7} & \cellcolor[RGB]{115,156,246}{\scriptsize 31.2\scalebox{0.55}{$\pm$31}} \\
& {\scriptsize G-3-Flash} & \cellcolor[RGB]{255,255,255}{\scriptsize 0.0} & \cellcolor[RGB]{212,225,252}{\scriptsize 2.9} & \cellcolor[RGB]{147,178,248}{\scriptsize 18.6} & \cellcolor[RGB]{117,158,246}{\scriptsize 30.1} & \cellcolor[RGB]{114,156,246}{\scriptsize 31.3} & \cellcolor[RGB]{126,163,247}{\scriptsize 26.5} & \cellcolor[RGB]{233,240,253}{\scriptsize 0.7} & \cellcolor[RGB]{203,218,251}{\scriptsize 4.3} & \cellcolor[RGB]{205,219,252}{\scriptsize 3.9} & \cellcolor[RGB]{103,148,246}{\scriptsize 36.5} & \cellcolor[RGB]{208,222,252}{\scriptsize 3.5} & \cellcolor[RGB]{135,170,247}{\scriptsize 22.6} & \cellcolor[RGB]{52,111,243}{\scriptsize 65.6} & \cellcolor[RGB]{38,102,242}{\scriptsize 90.7} & \cellcolor[RGB]{56,114,243}{\scriptsize 63.1} & \cellcolor[RGB]{38,102,242}{\scriptsize 96.3} & \cellcolor[RGB]{115,156,246}{\scriptsize 31.0\scalebox{0.55}{$\pm$31}} \\
& {\scriptsize G-3.1-Pro} & \cellcolor[RGB]{255,255,255}{\scriptsize 0.0} & \cellcolor[RGB]{214,226,252}{\scriptsize 2.6} & \cellcolor[RGB]{149,180,248}{\scriptsize 17.9} & \cellcolor[RGB]{117,157,246}{\scriptsize 30.4} & \cellcolor[RGB]{116,156,246}{\scriptsize 30.8} & \cellcolor[RGB]{126,164,247}{\scriptsize 26.4} & \cellcolor[RGB]{235,241,253}{\scriptsize 0.6} & \cellcolor[RGB]{203,218,251}{\scriptsize 4.2} & \cellcolor[RGB]{203,218,251}{\scriptsize 4.2} & \cellcolor[RGB]{101,146,245}{\scriptsize 37.5} & \cellcolor[RGB]{208,222,252}{\scriptsize 3.5} & \cellcolor[RGB]{135,170,247}{\scriptsize 23.0} & \cellcolor[RGB]{52,112,243}{\scriptsize 65.2} & \cellcolor[RGB]{38,102,242}{\scriptsize 93.7} & \cellcolor[RGB]{51,111,243}{\scriptsize 65.8} & \cellcolor[RGB]{38,102,242}{\scriptsize 96.1} & \cellcolor[RGB]{114,156,246}{\scriptsize 31.4\scalebox{0.55}{$\pm$31}} \\
\cmidrule[0.5pt](l){2-19}
& {\scriptsize Gen.\ quality} & \cellcolor[RGB]{255,255,255}{\scriptsize 0.0\scalebox{0.55}{$\pm$0}} & \cellcolor[RGB]{217,228,252}{\scriptsize 2.3\scalebox{0.55}{$\pm$0}} & \cellcolor[RGB]{157,186,249}{\scriptsize 15.2\scalebox{0.55}{$\pm$4}} & \cellcolor[RGB]{132,168,247}{\scriptsize 24.0\scalebox{0.55}{$\pm$7}} & \cellcolor[RGB]{134,169,247}{\scriptsize 23.4\scalebox{0.55}{$\pm$8}} & \cellcolor[RGB]{142,175,248}{\scriptsize 20.1\scalebox{0.55}{$\pm$7}} & \cellcolor[RGB]{236,241,253}{\scriptsize 0.6\scalebox{0.55}{$\pm$0}} & \cellcolor[RGB]{208,221,252}{\scriptsize 3.5\scalebox{0.55}{$\pm$1}} & \cellcolor[RGB]{210,223,252}{\scriptsize 3.1\scalebox{0.55}{$\pm$1}} & \cellcolor[RGB]{120,160,247}{\scriptsize 28.7\scalebox{0.55}{$\pm$9}} & \cellcolor[RGB]{213,226,252}{\scriptsize 2.7\scalebox{0.55}{$\pm$1}} & \cellcolor[RGB]{147,179,248}{\scriptsize 18.3\scalebox{0.55}{$\pm$5}} & \cellcolor[RGB]{77,129,244}{\scriptsize 50.5\scalebox{0.55}{$\pm$17}} & \cellcolor[RGB]{70,124,244}{\scriptsize 54.4\scalebox{0.55}{$\pm$28}} & \cellcolor[RGB]{90,138,245}{\scriptsize 43.2\scalebox{0.55}{$\pm$19}} & \cellcolor[RGB]{56,114,243}{\scriptsize 62.8\scalebox{0.55}{$\pm$29}} &  \\
\bottomrule
\end{tabular}
}%  end resizebox
\end{table}

\begin{table}[htb]
\centering
\caption{Communication matrix at depth $d = 5$ (830 expressions). Each cell shows round-trip accuracy (\%). Rows are extractors, columns are generators. Dotted lines separate open-weight AR, diffusion, and frontier models.}
\vspace{2mm}
\label{tab:matrix_d5}
\small
\setlength{\tabcolsep}{2pt}
\setlength{\dashlinedash}{1pt}
\setlength{\dashlinegap}{1pt}
\resizebox{\textwidth}{!}{%
\begin{tabular}{c l c c c c c c c c c c !{\;\vrule width 0.3pt\;} c c !{\;\vrule width 0.3pt\;} c c c c !{\;\vrule width 0.75pt\;} c}
\toprule
\multicolumn{19}{c}{\textsc{Generators}} \\
 \multirow{20}{*}{\rotatebox{90}{\textsc{Extractors}}} & & \rotatebox{90}{\scriptsize Qwen3-0.6B} & \rotatebox{90}{\scriptsize Qwen3-1.7B} & \rotatebox{90}{\scriptsize Qwen3-4B} & \rotatebox{90}{\scriptsize Qwen3-8B} & \rotatebox{90}{\scriptsize Qwen3-14B} & \rotatebox{90}{\scriptsize Qwen3-32B} & \rotatebox{90}{\scriptsize Gemma-3-4B} & \rotatebox{90}{\scriptsize Gemma-3-12B} & \rotatebox{90}{\scriptsize Gemma-3-27B} & \rotatebox{90}{\scriptsize Phi-4} & \rotatebox{90}{\scriptsize Dream-7B} & \rotatebox{90}{\scriptsize LLaDA-8B} & \rotatebox{90}{\scriptsize C-Haiku-4.5} & \rotatebox{90}{\scriptsize GPT-5} & \rotatebox{90}{\scriptsize G-3-Flash} & \rotatebox{90}{\scriptsize G-3.1-Pro} & \rotatebox{90}{\scriptsize Extr.\ quality} \\
\cmidrule[0.5pt](l){2-19}
& {\scriptsize $\Guard$ pass \%} & \cellcolor[RGB]{228,247,228}{\scriptsize 21} & \cellcolor[RGB]{179,232,179}{\scriptsize 59} & \cellcolor[RGB]{136,219,136}{\scriptsize 93} & \cellcolor[RGB]{131,217,131}{\scriptsize 97} & \cellcolor[RGB]{128,217,128}{\scriptsize 99} & \cellcolor[RGB]{128,217,128}{\scriptsize 99} & \cellcolor[RGB]{130,217,130}{\scriptsize 98} & \cellcolor[RGB]{132,218,132}{\scriptsize 96} & \cellcolor[RGB]{130,217,130}{\scriptsize 97} & \cellcolor[RGB]{130,217,130}{\scriptsize 97} & \cellcolor[RGB]{173,230,173}{\scriptsize 64} & \cellcolor[RGB]{134,218,134}{\scriptsize 95} & \cellcolor[RGB]{128,216,128}{\scriptsize 100} & \cellcolor[RGB]{127,216,127}{\scriptsize 100} & \cellcolor[RGB]{127,216,127}{\scriptsize 100} & \cellcolor[RGB]{127,216,127}{\scriptsize 100} &  \\
\cmidrule[0.5pt](l){2-19}
& {\scriptsize Qwen3-0.6B} & \cellcolor[RGB]{255,255,255}{\scriptsize 0.0} & \cellcolor[RGB]{237,242,253}{\scriptsize 0.5} & \cellcolor[RGB]{222,232,253}{\scriptsize 1.7} & \cellcolor[RGB]{212,224,252}{\scriptsize 2.9} & \cellcolor[RGB]{217,228,252}{\scriptsize 2.3} & \cellcolor[RGB]{221,231,253}{\scriptsize 1.8} & \cellcolor[RGB]{246,248,254}{\scriptsize 0.1} & \cellcolor[RGB]{246,248,254}{\scriptsize 0.1} & \cellcolor[RGB]{242,246,254}{\scriptsize 0.2} & \cellcolor[RGB]{202,218,251}{\scriptsize 4.3} & \cellcolor[RGB]{255,255,255}{\scriptsize 0.0} & \cellcolor[RGB]{210,223,252}{\scriptsize 3.1} & \cellcolor[RGB]{190,209,251}{\scriptsize 6.6} & \cellcolor[RGB]{230,237,253}{\scriptsize 1.0} & \cellcolor[RGB]{235,241,253}{\scriptsize 0.6} & \cellcolor[RGB]{233,239,253}{\scriptsize 0.7} & \cellcolor[RGB]{223,232,253}{\scriptsize 1.6\scalebox{0.55}{$\pm$2}} \\
& {\scriptsize Qwen3-1.7B} & \cellcolor[RGB]{255,255,255}{\scriptsize 0.0} & \cellcolor[RGB]{226,234,253}{\scriptsize 1.3} & \cellcolor[RGB]{195,212,251}{\scriptsize 5.7} & \cellcolor[RGB]{175,199,250}{\scriptsize 10.0} & \cellcolor[RGB]{185,205,250}{\scriptsize 7.7} & \cellcolor[RGB]{186,206,250}{\scriptsize 7.6} & \cellcolor[RGB]{242,246,254}{\scriptsize 0.2} & \cellcolor[RGB]{223,232,253}{\scriptsize 1.6} & \cellcolor[RGB]{228,236,253}{\scriptsize 1.1} & \cellcolor[RGB]{157,186,249}{\scriptsize 15.1} & \cellcolor[RGB]{230,237,253}{\scriptsize 1.0} & \cellcolor[RGB]{183,204,250}{\scriptsize 8.1} & \cellcolor[RGB]{129,166,247}{\scriptsize 25.2} & \cellcolor[RGB]{180,202,250}{\scriptsize 8.9} & \cellcolor[RGB]{179,201,250}{\scriptsize 9.2} & \cellcolor[RGB]{166,192,249}{\scriptsize 12.4} & \cellcolor[RGB]{187,207,251}{\scriptsize 7.2\scalebox{0.55}{$\pm$6}} \\
& {\scriptsize Qwen3-4B} & \cellcolor[RGB]{255,255,255}{\scriptsize 0.0} & \cellcolor[RGB]{222,232,253}{\scriptsize 1.7} & \cellcolor[RGB]{168,193,249}{\scriptsize 12.0} & \cellcolor[RGB]{148,179,248}{\scriptsize 18.2} & \cellcolor[RGB]{139,173,248}{\scriptsize 21.3} & \cellcolor[RGB]{148,179,248}{\scriptsize 18.1} & \cellcolor[RGB]{237,242,253}{\scriptsize 0.5} & \cellcolor[RGB]{208,221,252}{\scriptsize 3.5} & \cellcolor[RGB]{209,223,252}{\scriptsize 3.3} & \cellcolor[RGB]{124,162,247}{\scriptsize 27.3} & \cellcolor[RGB]{222,232,253}{\scriptsize 1.7} & \cellcolor[RGB]{157,186,249}{\scriptsize 15.1} & \cellcolor[RGB]{82,132,244}{\scriptsize 47.7} & \cellcolor[RGB]{100,145,245}{\scriptsize 38.2} & \cellcolor[RGB]{109,152,246}{\scriptsize 33.7} & \cellcolor[RGB]{85,135,245}{\scriptsize 45.8} & \cellcolor[RGB]{148,180,248}{\scriptsize 18.0\scalebox{0.55}{$\pm$16}} \\
& {\scriptsize Qwen3-8B} & \cellcolor[RGB]{255,255,255}{\scriptsize 0.0} & \cellcolor[RGB]{223,232,253}{\scriptsize 1.6} & \cellcolor[RGB]{162,189,249}{\scriptsize 13.7} & \cellcolor[RGB]{149,180,248}{\scriptsize 17.8} & \cellcolor[RGB]{137,172,248}{\scriptsize 21.9} & \cellcolor[RGB]{142,175,248}{\scriptsize 20.4} & \cellcolor[RGB]{237,242,253}{\scriptsize 0.5} & \cellcolor[RGB]{212,224,252}{\scriptsize 2.9} & \cellcolor[RGB]{205,219,252}{\scriptsize 4.0} & \cellcolor[RGB]{121,160,247}{\scriptsize 28.4} & \cellcolor[RGB]{228,236,253}{\scriptsize 1.1} & \cellcolor[RGB]{154,183,249}{\scriptsize 16.3} & \cellcolor[RGB]{78,130,244}{\scriptsize 50.0} & \cellcolor[RGB]{90,138,245}{\scriptsize 43.1} & \cellcolor[RGB]{98,144,245}{\scriptsize 39.3} & \cellcolor[RGB]{75,128,244}{\scriptsize 51.4} & \cellcolor[RGB]{144,176,248}{\scriptsize 19.5\scalebox{0.55}{$\pm$18}} \\
& {\scriptsize Qwen3-14B} & \cellcolor[RGB]{255,255,255}{\scriptsize 0.0} & \cellcolor[RGB]{222,232,253}{\scriptsize 1.7} & \cellcolor[RGB]{158,186,249}{\scriptsize 14.8} & \cellcolor[RGB]{138,172,248}{\scriptsize 21.6} & \cellcolor[RGB]{131,168,247}{\scriptsize 24.2} & \cellcolor[RGB]{134,169,247}{\scriptsize 23.3} & \cellcolor[RGB]{235,241,253}{\scriptsize 0.6} & \cellcolor[RGB]{200,216,251}{\scriptsize 4.7} & \cellcolor[RGB]{202,218,251}{\scriptsize 4.3} & \cellcolor[RGB]{113,154,246}{\scriptsize 32.2} & \cellcolor[RGB]{221,231,253}{\scriptsize 1.8} & \cellcolor[RGB]{148,179,248}{\scriptsize 18.2} & \cellcolor[RGB]{63,119,243}{\scriptsize 58.8} & \cellcolor[RGB]{68,123,244}{\scriptsize 55.3} & \cellcolor[RGB]{74,127,244}{\scriptsize 51.8} & \cellcolor[RGB]{55,113,243}{\scriptsize 63.9} & \cellcolor[RGB]{133,169,247}{\scriptsize 23.6\scalebox{0.55}{$\pm$22}} \\
& {\scriptsize Qwen3-32B} & \cellcolor[RGB]{255,255,255}{\scriptsize 0.0} & \cellcolor[RGB]{222,232,253}{\scriptsize 1.7} & \cellcolor[RGB]{158,186,249}{\scriptsize 14.9} & \cellcolor[RGB]{138,172,248}{\scriptsize 21.7} & \cellcolor[RGB]{129,166,247}{\scriptsize 25.1} & \cellcolor[RGB]{130,167,247}{\scriptsize 24.6} & \cellcolor[RGB]{237,242,253}{\scriptsize 0.5} & \cellcolor[RGB]{208,221,252}{\scriptsize 3.5} & \cellcolor[RGB]{200,216,251}{\scriptsize 4.8} & \cellcolor[RGB]{114,155,246}{\scriptsize 31.4} & \cellcolor[RGB]{218,228,252}{\scriptsize 2.2} & \cellcolor[RGB]{146,178,248}{\scriptsize 18.8} & \cellcolor[RGB]{63,119,243}{\scriptsize 58.7} & \cellcolor[RGB]{57,115,243}{\scriptsize 62.2} & \cellcolor[RGB]{75,128,244}{\scriptsize 51.2} & \cellcolor[RGB]{44,106,242}{\scriptsize 70.8} & \cellcolor[RGB]{131,167,247}{\scriptsize 24.5\scalebox{0.55}{$\pm$23}} \\
& {\scriptsize Gemma-3-4B} & \cellcolor[RGB]{255,255,255}{\scriptsize 0.0} & \cellcolor[RGB]{228,236,253}{\scriptsize 1.1} & \cellcolor[RGB]{182,203,250}{\scriptsize 8.4} & \cellcolor[RGB]{166,192,249}{\scriptsize 12.5} & \cellcolor[RGB]{168,193,249}{\scriptsize 12.0} & \cellcolor[RGB]{174,198,250}{\scriptsize 10.4} & \cellcolor[RGB]{242,246,254}{\scriptsize 0.2} & \cellcolor[RGB]{214,226,252}{\scriptsize 2.7} & \cellcolor[RGB]{222,232,253}{\scriptsize 1.7} & \cellcolor[RGB]{152,182,248}{\scriptsize 16.7} & \cellcolor[RGB]{230,237,253}{\scriptsize 1.0} & \cellcolor[RGB]{172,196,250}{\scriptsize 10.8} & \cellcolor[RGB]{121,160,247}{\scriptsize 28.4} & \cellcolor[RGB]{151,181,248}{\scriptsize 17.1} & \cellcolor[RGB]{152,182,248}{\scriptsize 16.9} & \cellcolor[RGB]{151,181,248}{\scriptsize 17.2} & \cellcolor[RGB]{176,199,250}{\scriptsize 9.8\scalebox{0.55}{$\pm$8}} \\
& {\scriptsize Gemma-3-12B} & \cellcolor[RGB]{255,255,255}{\scriptsize 0.0} & \cellcolor[RGB]{226,234,253}{\scriptsize 1.3} & \cellcolor[RGB]{169,194,249}{\scriptsize 11.7} & \cellcolor[RGB]{156,185,249}{\scriptsize 15.4} & \cellcolor[RGB]{150,180,248}{\scriptsize 17.6} & \cellcolor[RGB]{151,181,248}{\scriptsize 17.2} & \cellcolor[RGB]{239,244,254}{\scriptsize 0.4} & \cellcolor[RGB]{215,226,252}{\scriptsize 2.5} & \cellcolor[RGB]{209,222,252}{\scriptsize 3.4} & \cellcolor[RGB]{128,165,247}{\scriptsize 25.4} & \cellcolor[RGB]{222,232,253}{\scriptsize 1.7} & \cellcolor[RGB]{159,187,249}{\scriptsize 14.6} & \cellcolor[RGB]{89,138,245}{\scriptsize 43.7} & \cellcolor[RGB]{103,147,246}{\scriptsize 36.9} & \cellcolor[RGB]{112,154,246}{\scriptsize 32.4} & \cellcolor[RGB]{93,140,245}{\scriptsize 41.8} & \cellcolor[RGB]{152,182,248}{\scriptsize 16.6\scalebox{0.55}{$\pm$15}} \\
& {\scriptsize Gemma-3-27B} & \cellcolor[RGB]{255,255,255}{\scriptsize 0.0} & \cellcolor[RGB]{221,231,253}{\scriptsize 1.8} & \cellcolor[RGB]{159,187,249}{\scriptsize 14.6} & \cellcolor[RGB]{144,176,248}{\scriptsize 19.5} & \cellcolor[RGB]{133,169,247}{\scriptsize 23.4} & \cellcolor[RGB]{136,171,248}{\scriptsize 22.4} & \cellcolor[RGB]{233,239,253}{\scriptsize 0.7} & \cellcolor[RGB]{204,219,252}{\scriptsize 4.1} & \cellcolor[RGB]{200,216,251}{\scriptsize 4.8} & \cellcolor[RGB]{117,157,246}{\scriptsize 30.4} & \cellcolor[RGB]{221,231,253}{\scriptsize 1.8} & \cellcolor[RGB]{152,182,248}{\scriptsize 16.7} & \cellcolor[RGB]{70,124,244}{\scriptsize 54.2} & \cellcolor[RGB]{67,122,243}{\scriptsize 56.1} & \cellcolor[RGB]{79,131,244}{\scriptsize 48.9} & \cellcolor[RGB]{55,113,243}{\scriptsize 63.7} & \cellcolor[RGB]{135,170,247}{\scriptsize 22.7\scalebox{0.55}{$\pm$21}} \\
& {\scriptsize Phi-4} & \cellcolor[RGB]{255,255,255}{\scriptsize 0.0} & \cellcolor[RGB]{222,232,253}{\scriptsize 1.7} & \cellcolor[RGB]{154,184,249}{\scriptsize 16.0} & \cellcolor[RGB]{135,170,247}{\scriptsize 22.7} & \cellcolor[RGB]{126,164,247}{\scriptsize 26.5} & \cellcolor[RGB]{132,168,247}{\scriptsize 24.1} & \cellcolor[RGB]{233,239,253}{\scriptsize 0.7} & \cellcolor[RGB]{200,216,251}{\scriptsize 4.7} & \cellcolor[RGB]{204,219,252}{\scriptsize 4.1} & \cellcolor[RGB]{108,151,246}{\scriptsize 34.1} & \cellcolor[RGB]{218,228,252}{\scriptsize 2.2} & \cellcolor[RGB]{144,177,248}{\scriptsize 19.4} & \cellcolor[RGB]{61,118,243}{\scriptsize 59.5} & \cellcolor[RGB]{56,115,243}{\scriptsize 62.8} & \cellcolor[RGB]{69,123,244}{\scriptsize 55.2} & \cellcolor[RGB]{41,104,242}{\scriptsize 72.5} & \cellcolor[RGB]{128,165,247}{\scriptsize 25.4\scalebox{0.55}{$\pm$24}} \\
\cmidrule[0.3pt](l){2-19}
& {\scriptsize Dream-7B} & \cellcolor[RGB]{255,255,255}{\scriptsize 0.0} & \cellcolor[RGB]{227,235,253}{\scriptsize 1.2} & \cellcolor[RGB]{193,211,251}{\scriptsize 6.0} & \cellcolor[RGB]{168,194,249}{\scriptsize 11.8} & \cellcolor[RGB]{181,202,250}{\scriptsize 8.7} & \cellcolor[RGB]{173,197,250}{\scriptsize 10.6} & \cellcolor[RGB]{239,244,254}{\scriptsize 0.4} & \cellcolor[RGB]{218,228,252}{\scriptsize 2.2} & \cellcolor[RGB]{221,231,253}{\scriptsize 1.8} & \cellcolor[RGB]{165,192,249}{\scriptsize 12.7} & \cellcolor[RGB]{227,235,253}{\scriptsize 1.2} & \cellcolor[RGB]{181,203,250}{\scriptsize 8.6} & \cellcolor[RGB]{131,168,247}{\scriptsize 24.2} & \cellcolor[RGB]{151,181,248}{\scriptsize 17.1} & \cellcolor[RGB]{148,179,248}{\scriptsize 18.1} & \cellcolor[RGB]{113,155,246}{\scriptsize 31.8} & \cellcolor[RGB]{176,199,250}{\scriptsize 9.8\scalebox{0.55}{$\pm$9}} \\
& {\scriptsize LLaDA-8B} & \cellcolor[RGB]{255,255,255}{\scriptsize 0.0} & \cellcolor[RGB]{224,233,253}{\scriptsize 1.4} & \cellcolor[RGB]{168,193,249}{\scriptsize 11.9} & \cellcolor[RGB]{154,183,249}{\scriptsize 16.3} & \cellcolor[RGB]{148,179,248}{\scriptsize 18.2} & \cellcolor[RGB]{150,181,248}{\scriptsize 17.5} & \cellcolor[RGB]{237,242,253}{\scriptsize 0.5} & \cellcolor[RGB]{213,225,252}{\scriptsize 2.8} & \cellcolor[RGB]{213,225,252}{\scriptsize 2.8} & \cellcolor[RGB]{135,170,247}{\scriptsize 22.8} & \cellcolor[RGB]{223,232,253}{\scriptsize 1.6} & \cellcolor[RGB]{158,186,249}{\scriptsize 14.9} & \cellcolor[RGB]{95,142,245}{\scriptsize 40.6} & \cellcolor[RGB]{105,149,246}{\scriptsize 35.9} & \cellcolor[RGB]{111,153,246}{\scriptsize 32.9} & \cellcolor[RGB]{102,147,246}{\scriptsize 37.3} & \cellcolor[RGB]{154,184,249}{\scriptsize 16.1\scalebox{0.55}{$\pm$14}} \\
\cmidrule[0.3pt](l){2-19}
& {\scriptsize C-Haiku-4.5} & \cellcolor[RGB]{255,255,255}{\scriptsize 0.0} & \cellcolor[RGB]{219,229,252}{\scriptsize 2.0} & \cellcolor[RGB]{155,184,249}{\scriptsize 15.8} & \cellcolor[RGB]{137,172,248}{\scriptsize 21.9} & \cellcolor[RGB]{124,163,247}{\scriptsize 27.0} & \cellcolor[RGB]{129,166,247}{\scriptsize 25.3} & \cellcolor[RGB]{237,242,253}{\scriptsize 0.5} & \cellcolor[RGB]{207,221,252}{\scriptsize 3.6} & \cellcolor[RGB]{200,216,251}{\scriptsize 4.8} & \cellcolor[RGB]{109,152,246}{\scriptsize 34.0} & \cellcolor[RGB]{215,226,252}{\scriptsize 2.5} & \cellcolor[RGB]{145,177,248}{\scriptsize 19.2} & \cellcolor[RGB]{58,115,243}{\scriptsize 61.9} & \cellcolor[RGB]{38,102,242}{\scriptsize 78.7} & \cellcolor[RGB]{65,121,243}{\scriptsize 57.1} & \cellcolor[RGB]{38,102,242}{\scriptsize 86.7} & \cellcolor[RGB]{123,162,247}{\scriptsize 27.6\scalebox{0.55}{$\pm$28}} \\
& {\scriptsize GPT-5} & \cellcolor[RGB]{255,255,255}{\scriptsize 0.0} & \cellcolor[RGB]{220,230,252}{\scriptsize 1.9} & \cellcolor[RGB]{149,180,248}{\scriptsize 17.8} & \cellcolor[RGB]{133,169,247}{\scriptsize 23.4} & \cellcolor[RGB]{120,160,247}{\scriptsize 28.7} & \cellcolor[RGB]{144,176,248}{\scriptsize 19.5} & \cellcolor[RGB]{233,239,253}{\scriptsize 0.7} & \cellcolor[RGB]{200,216,251}{\scriptsize 4.8} & \cellcolor[RGB]{202,217,251}{\scriptsize 4.5} & \cellcolor[RGB]{104,148,246}{\scriptsize 36.1} & \cellcolor[RGB]{214,226,252}{\scriptsize 2.7} & \cellcolor[RGB]{137,171,248}{\scriptsize 22.2} & \cellcolor[RGB]{51,111,243}{\scriptsize 66.4} & \cellcolor[RGB]{38,102,242}{\scriptsize 91.2} & \cellcolor[RGB]{49,109,242}{\scriptsize 67.5} & \cellcolor[RGB]{38,102,242}{\scriptsize 94.9} & \cellcolor[RGB]{117,158,246}{\scriptsize 30.1\scalebox{0.55}{$\pm$31}} \\
& {\scriptsize G-3-Flash} & \cellcolor[RGB]{255,255,255}{\scriptsize 0.0} & \cellcolor[RGB]{217,228,252}{\scriptsize 2.3} & \cellcolor[RGB]{154,184,249}{\scriptsize 16.1} & \cellcolor[RGB]{134,170,247}{\scriptsize 23.0} & \cellcolor[RGB]{125,163,247}{\scriptsize 26.9} & \cellcolor[RGB]{128,165,247}{\scriptsize 25.7} & \cellcolor[RGB]{233,239,253}{\scriptsize 0.7} & \cellcolor[RGB]{206,220,252}{\scriptsize 3.7} & \cellcolor[RGB]{200,216,251}{\scriptsize 4.8} & \cellcolor[RGB]{105,149,246}{\scriptsize 35.5} & \cellcolor[RGB]{216,227,252}{\scriptsize 2.4} & \cellcolor[RGB]{140,174,248}{\scriptsize 20.8} & \cellcolor[RGB]{53,112,243}{\scriptsize 64.7} & \cellcolor[RGB]{38,102,242}{\scriptsize 89.2} & \cellcolor[RGB]{57,115,243}{\scriptsize 62.3} & \cellcolor[RGB]{38,102,242}{\scriptsize 95.3} & \cellcolor[RGB]{118,158,246}{\scriptsize 29.6\scalebox{0.55}{$\pm$31}} \\
& {\scriptsize G-3.1-Pro} & \cellcolor[RGB]{255,255,255}{\scriptsize 0.0} & \cellcolor[RGB]{221,231,253}{\scriptsize 1.8} & \cellcolor[RGB]{155,184,249}{\scriptsize 15.8} & \cellcolor[RGB]{137,172,248}{\scriptsize 22.0} & \cellcolor[RGB]{123,162,247}{\scriptsize 27.6} & \cellcolor[RGB]{126,164,247}{\scriptsize 26.3} & \cellcolor[RGB]{239,244,254}{\scriptsize 0.4} & \cellcolor[RGB]{205,220,252}{\scriptsize 3.9} & \cellcolor[RGB]{202,217,251}{\scriptsize 4.5} & \cellcolor[RGB]{103,148,246}{\scriptsize 36.6} & \cellcolor[RGB]{216,227,252}{\scriptsize 2.4} & \cellcolor[RGB]{141,174,248}{\scriptsize 20.7} & \cellcolor[RGB]{55,114,243}{\scriptsize 63.5} & \cellcolor[RGB]{38,102,242}{\scriptsize 93.3} & \cellcolor[RGB]{53,113,243}{\scriptsize 64.6} & \cellcolor[RGB]{38,102,242}{\scriptsize 96.3} & \cellcolor[RGB]{117,158,246}{\scriptsize 30.0\scalebox{0.55}{$\pm$31}} \\
\cmidrule[0.5pt](l){2-19}
& {\scriptsize Gen.\ quality} & \cellcolor[RGB]{255,255,255}{\scriptsize 0.0\scalebox{0.55}{$\pm$0}} & \cellcolor[RGB]{223,232,253}{\scriptsize 1.6\scalebox{0.55}{$\pm$0}} & \cellcolor[RGB]{167,192,249}{\scriptsize 12.3\scalebox{0.55}{$\pm$4}} & \cellcolor[RGB]{150,180,248}{\scriptsize 17.5\scalebox{0.55}{$\pm$6}} & \cellcolor[RGB]{143,176,248}{\scriptsize 19.9\scalebox{0.55}{$\pm$8}} & \cellcolor[RGB]{147,179,248}{\scriptsize 18.4\scalebox{0.55}{$\pm$7}} & \cellcolor[RGB]{237,242,253}{\scriptsize 0.5\scalebox{0.55}{$\pm$0}} & \cellcolor[RGB]{210,223,252}{\scriptsize 3.2\scalebox{0.55}{$\pm$1}} & \cellcolor[RGB]{208,222,252}{\scriptsize 3.4\scalebox{0.55}{$\pm$1}} & \cellcolor[RGB]{126,164,247}{\scriptsize 26.4\scalebox{0.55}{$\pm$9}} & \cellcolor[RGB]{222,232,253}{\scriptsize 1.7\scalebox{0.55}{$\pm$1}} & \cellcolor[RGB]{156,185,249}{\scriptsize 15.5\scalebox{0.55}{$\pm$5}} & \cellcolor[RGB]{83,133,244}{\scriptsize 47.1\scalebox{0.55}{$\pm$17}} & \cellcolor[RGB]{79,131,244}{\scriptsize 49.2\scalebox{0.55}{$\pm$29}} & \cellcolor[RGB]{96,143,245}{\scriptsize 40.1\scalebox{0.55}{$\pm$20}} & \cellcolor[RGB]{69,123,244}{\scriptsize 55.2\scalebox{0.55}{$\pm$30}} &  \\
\bottomrule
\end{tabular}
}%  end resizebox
\end{table}

\begin{table}[htb]
\centering
\caption{Communication matrix at depth $d = 6$ (660 expressions). Each cell shows round-trip accuracy (\%). Rows are extractors, columns are generators. Dotted lines separate open-weight AR, diffusion, and frontier models.}
\vspace{2mm}
\label{tab:matrix_d6}
\small
\setlength{\tabcolsep}{2pt}
\setlength{\dashlinedash}{1pt}
\setlength{\dashlinegap}{1pt}
\resizebox{\textwidth}{!}{%
\begin{tabular}{c l c c c c c c c c c c !{\;\vrule width 0.3pt\;} c c !{\;\vrule width 0.3pt\;} c c c c !{\;\vrule width 0.75pt\;} c}
\toprule
\multicolumn{19}{c}{\textsc{Generators}} \\
 \multirow{20}{*}{\rotatebox{90}{\textsc{Extractors}}} & & \rotatebox{90}{\scriptsize Qwen3-0.6B} & \rotatebox{90}{\scriptsize Qwen3-1.7B} & \rotatebox{90}{\scriptsize Qwen3-4B} & \rotatebox{90}{\scriptsize Qwen3-8B} & \rotatebox{90}{\scriptsize Qwen3-14B} & \rotatebox{90}{\scriptsize Qwen3-32B} & \rotatebox{90}{\scriptsize Gemma-3-4B} & \rotatebox{90}{\scriptsize Gemma-3-12B} & \rotatebox{90}{\scriptsize Gemma-3-27B} & \rotatebox{90}{\scriptsize Phi-4} & \rotatebox{90}{\scriptsize Dream-7B} & \rotatebox{90}{\scriptsize LLaDA-8B} & \rotatebox{90}{\scriptsize C-Haiku-4.5} & \rotatebox{90}{\scriptsize GPT-5} & \rotatebox{90}{\scriptsize G-3-Flash} & \rotatebox{90}{\scriptsize G-3.1-Pro} & \rotatebox{90}{\scriptsize Extr.\ quality} \\
\cmidrule[0.5pt](l){2-19}
& {\scriptsize $\Guard$ pass \%} & \cellcolor[RGB]{227,246,227}{\scriptsize 21} & \cellcolor[RGB]{176,231,176}{\scriptsize 62} & \cellcolor[RGB]{136,219,136}{\scriptsize 93} & \cellcolor[RGB]{130,217,130}{\scriptsize 97} & \cellcolor[RGB]{128,216,128}{\scriptsize 100} & \cellcolor[RGB]{128,217,128}{\scriptsize 99} & \cellcolor[RGB]{129,217,129}{\scriptsize 99} & \cellcolor[RGB]{133,218,133}{\scriptsize 95} & \cellcolor[RGB]{134,218,134}{\scriptsize 95} & \cellcolor[RGB]{131,218,131}{\scriptsize 97} & \cellcolor[RGB]{174,230,174}{\scriptsize 63} & \cellcolor[RGB]{133,218,133}{\scriptsize 95} & \cellcolor[RGB]{128,217,128}{\scriptsize 99} & \cellcolor[RGB]{127,216,127}{\scriptsize 100} & \cellcolor[RGB]{127,216,127}{\scriptsize 100} & \cellcolor[RGB]{127,216,127}{\scriptsize 100} &  \\
\cmidrule[0.5pt](l){2-19}
& {\scriptsize Qwen3-0.6B} & \cellcolor[RGB]{255,255,255}{\scriptsize 0.0} & \cellcolor[RGB]{235,241,253}{\scriptsize 0.6} & \cellcolor[RGB]{229,236,253}{\scriptsize 1.1} & \cellcolor[RGB]{219,230,252}{\scriptsize 2.0} & \cellcolor[RGB]{229,236,253}{\scriptsize 1.1} & \cellcolor[RGB]{241,245,254}{\scriptsize 0.3} & \cellcolor[RGB]{245,248,254}{\scriptsize 0.2} & \cellcolor[RGB]{255,255,255}{\scriptsize 0.0} & \cellcolor[RGB]{255,255,255}{\scriptsize 0.0} & \cellcolor[RGB]{222,232,253}{\scriptsize 1.7} & \cellcolor[RGB]{255,255,255}{\scriptsize 0.0} & \cellcolor[RGB]{233,239,253}{\scriptsize 0.8} & \cellcolor[RGB]{208,222,252}{\scriptsize 3.5} & \cellcolor[RGB]{235,241,253}{\scriptsize 0.6} & \cellcolor[RGB]{235,241,253}{\scriptsize 0.6} & \cellcolor[RGB]{235,241,253}{\scriptsize 0.6} & \cellcolor[RGB]{232,239,253}{\scriptsize 0.8\scalebox{0.55}{$\pm$1}} \\
& {\scriptsize Qwen3-1.7B} & \cellcolor[RGB]{255,255,255}{\scriptsize 0.0} & \cellcolor[RGB]{227,235,253}{\scriptsize 1.2} & \cellcolor[RGB]{198,214,251}{\scriptsize 5.2} & \cellcolor[RGB]{193,211,251}{\scriptsize 6.1} & \cellcolor[RGB]{197,214,251}{\scriptsize 5.3} & \cellcolor[RGB]{199,216,251}{\scriptsize 4.8} & \cellcolor[RGB]{241,245,254}{\scriptsize 0.3} & \cellcolor[RGB]{238,243,254}{\scriptsize 0.5} & \cellcolor[RGB]{235,241,253}{\scriptsize 0.6} & \cellcolor[RGB]{186,206,250}{\scriptsize 7.4} & \cellcolor[RGB]{241,245,254}{\scriptsize 0.3} & \cellcolor[RGB]{202,217,251}{\scriptsize 4.4} & \cellcolor[RGB]{151,181,248}{\scriptsize 17.1} & \cellcolor[RGB]{189,208,251}{\scriptsize 6.8} & \cellcolor[RGB]{186,206,250}{\scriptsize 7.4} & \cellcolor[RGB]{186,206,250}{\scriptsize 7.4} & \cellcolor[RGB]{200,216,251}{\scriptsize 4.7\scalebox{0.55}{$\pm$4}} \\
& {\scriptsize Qwen3-4B} & \cellcolor[RGB]{255,255,255}{\scriptsize 0.0} & \cellcolor[RGB]{214,226,252}{\scriptsize 2.6} & \cellcolor[RGB]{173,197,250}{\scriptsize 10.6} & \cellcolor[RGB]{164,191,249}{\scriptsize 13.0} & \cellcolor[RGB]{161,188,249}{\scriptsize 14.1} & \cellcolor[RGB]{160,188,249}{\scriptsize 14.2} & \cellcolor[RGB]{241,245,254}{\scriptsize 0.3} & \cellcolor[RGB]{229,236,253}{\scriptsize 1.1} & \cellcolor[RGB]{227,235,253}{\scriptsize 1.2} & \cellcolor[RGB]{146,178,248}{\scriptsize 18.6} & \cellcolor[RGB]{233,239,253}{\scriptsize 0.8} & \cellcolor[RGB]{178,200,250}{\scriptsize 9.4} & \cellcolor[RGB]{100,145,245}{\scriptsize 38.3} & \cellcolor[RGB]{120,159,247}{\scriptsize 28.9} & \cellcolor[RGB]{128,165,247}{\scriptsize 25.6} & \cellcolor[RGB]{120,159,247}{\scriptsize 28.9} & \cellcolor[RGB]{164,191,249}{\scriptsize 13.0\scalebox{0.55}{$\pm$12}} \\
& {\scriptsize Qwen3-8B} & \cellcolor[RGB]{255,255,255}{\scriptsize 0.0} & \cellcolor[RGB]{216,227,252}{\scriptsize 2.4} & \cellcolor[RGB]{175,199,250}{\scriptsize 10.0} & \cellcolor[RGB]{170,195,250}{\scriptsize 11.5} & \cellcolor[RGB]{161,188,249}{\scriptsize 14.1} & \cellcolor[RGB]{159,187,249}{\scriptsize 14.7} & \cellcolor[RGB]{241,245,254}{\scriptsize 0.3} & \cellcolor[RGB]{231,238,253}{\scriptsize 0.9} & \cellcolor[RGB]{221,231,253}{\scriptsize 1.8} & \cellcolor[RGB]{149,180,248}{\scriptsize 17.9} & \cellcolor[RGB]{235,241,253}{\scriptsize 0.6} & \cellcolor[RGB]{178,200,250}{\scriptsize 9.4} & \cellcolor[RGB]{107,150,246}{\scriptsize 34.8} & \cellcolor[RGB]{110,153,246}{\scriptsize 33.2} & \cellcolor[RGB]{126,164,247}{\scriptsize 26.2} & \cellcolor[RGB]{112,154,246}{\scriptsize 32.3} & \cellcolor[RGB]{164,190,249}{\scriptsize 13.1\scalebox{0.55}{$\pm$12}} \\
& {\scriptsize Qwen3-14B} & \cellcolor[RGB]{255,255,255}{\scriptsize 0.0} & \cellcolor[RGB]{212,225,252}{\scriptsize 2.9} & \cellcolor[RGB]{168,193,249}{\scriptsize 12.0} & \cellcolor[RGB]{158,186,249}{\scriptsize 14.8} & \cellcolor[RGB]{149,180,248}{\scriptsize 17.9} & \cellcolor[RGB]{149,180,248}{\scriptsize 17.7} & \cellcolor[RGB]{235,241,253}{\scriptsize 0.6} & \cellcolor[RGB]{222,232,253}{\scriptsize 1.7} & \cellcolor[RGB]{219,230,252}{\scriptsize 2.0} & \cellcolor[RGB]{131,167,247}{\scriptsize 24.5} & \cellcolor[RGB]{233,239,253}{\scriptsize 0.8} & \cellcolor[RGB]{171,195,250}{\scriptsize 11.2} & \cellcolor[RGB]{91,139,245}{\scriptsize 42.9} & \cellcolor[RGB]{89,138,245}{\scriptsize 43.8} & \cellcolor[RGB]{98,144,245}{\scriptsize 39.2} & \cellcolor[RGB]{85,135,245}{\scriptsize 45.9} & \cellcolor[RGB]{150,181,248}{\scriptsize 17.4\scalebox{0.55}{$\pm$16}} \\
& {\scriptsize Qwen3-32B} & \cellcolor[RGB]{255,255,255}{\scriptsize 0.0} & \cellcolor[RGB]{211,224,252}{\scriptsize 3.0} & \cellcolor[RGB]{166,192,249}{\scriptsize 12.4} & \cellcolor[RGB]{158,186,249}{\scriptsize 14.8} & \cellcolor[RGB]{149,180,248}{\scriptsize 17.9} & \cellcolor[RGB]{145,177,248}{\scriptsize 19.1} & \cellcolor[RGB]{238,243,254}{\scriptsize 0.5} & \cellcolor[RGB]{222,232,253}{\scriptsize 1.7} & \cellcolor[RGB]{216,227,252}{\scriptsize 2.4} & \cellcolor[RGB]{132,168,247}{\scriptsize 24.1} & \cellcolor[RGB]{227,235,253}{\scriptsize 1.2} & \cellcolor[RGB]{172,196,250}{\scriptsize 10.9} & \cellcolor[RGB]{84,134,244}{\scriptsize 46.4} & \cellcolor[RGB]{74,127,244}{\scriptsize 52.3} & \cellcolor[RGB]{91,139,245}{\scriptsize 42.6} & \cellcolor[RGB]{75,128,244}{\scriptsize 51.7} & \cellcolor[RGB]{146,178,248}{\scriptsize 18.8\scalebox{0.55}{$\pm$19}} \\
& {\scriptsize Gemma-3-4B} & \cellcolor[RGB]{255,255,255}{\scriptsize 0.0} & \cellcolor[RGB]{221,231,253}{\scriptsize 1.8} & \cellcolor[RGB]{192,210,251}{\scriptsize 6.2} & \cellcolor[RGB]{190,209,251}{\scriptsize 6.7} & \cellcolor[RGB]{186,206,250}{\scriptsize 7.6} & \cellcolor[RGB]{194,212,251}{\scriptsize 5.8} & \cellcolor[RGB]{245,248,254}{\scriptsize 0.2} & \cellcolor[RGB]{235,241,253}{\scriptsize 0.6} & \cellcolor[RGB]{241,245,254}{\scriptsize 0.3} & \cellcolor[RGB]{174,198,250}{\scriptsize 10.3} & \cellcolor[RGB]{245,248,254}{\scriptsize 0.2} & \cellcolor[RGB]{196,213,251}{\scriptsize 5.5} & \cellcolor[RGB]{144,176,248}{\scriptsize 19.5} & \cellcolor[RGB]{177,200,250}{\scriptsize 9.5} & \cellcolor[RGB]{178,201,250}{\scriptsize 9.2} & \cellcolor[RGB]{167,193,249}{\scriptsize 12.1} & \cellcolor[RGB]{193,211,251}{\scriptsize 6.0\scalebox{0.55}{$\pm$5}} \\
& {\scriptsize Gemma-3-12B} & \cellcolor[RGB]{255,255,255}{\scriptsize 0.0} & \cellcolor[RGB]{224,233,253}{\scriptsize 1.5} & \cellcolor[RGB]{176,199,250}{\scriptsize 9.8} & \cellcolor[RGB]{172,197,250}{\scriptsize 10.8} & \cellcolor[RGB]{166,192,249}{\scriptsize 12.6} & \cellcolor[RGB]{167,193,249}{\scriptsize 12.1} & \cellcolor[RGB]{241,245,254}{\scriptsize 0.3} & \cellcolor[RGB]{233,239,253}{\scriptsize 0.8} & \cellcolor[RGB]{221,231,253}{\scriptsize 1.8} & \cellcolor[RGB]{156,185,249}{\scriptsize 15.5} & \cellcolor[RGB]{231,238,253}{\scriptsize 0.9} & \cellcolor[RGB]{185,205,250}{\scriptsize 7.7} & \cellcolor[RGB]{107,150,246}{\scriptsize 34.7} & \cellcolor[RGB]{132,168,247}{\scriptsize 24.1} & \cellcolor[RGB]{140,174,248}{\scriptsize 20.8} & \cellcolor[RGB]{121,160,247}{\scriptsize 28.6} & \cellcolor[RGB]{170,195,250}{\scriptsize 11.4\scalebox{0.55}{$\pm$11}} \\
& {\scriptsize Gemma-3-27B} & \cellcolor[RGB]{255,255,255}{\scriptsize 0.0} & \cellcolor[RGB]{214,226,252}{\scriptsize 2.6} & \cellcolor[RGB]{169,194,249}{\scriptsize 11.7} & \cellcolor[RGB]{164,190,249}{\scriptsize 13.2} & \cellcolor[RGB]{153,183,249}{\scriptsize 16.5} & \cellcolor[RGB]{150,181,248}{\scriptsize 17.4} & \cellcolor[RGB]{235,241,253}{\scriptsize 0.6} & \cellcolor[RGB]{227,235,253}{\scriptsize 1.2} & \cellcolor[RGB]{221,231,253}{\scriptsize 1.8} & \cellcolor[RGB]{140,174,248}{\scriptsize 20.9} & \cellcolor[RGB]{233,239,253}{\scriptsize 0.8} & \cellcolor[RGB]{172,197,250}{\scriptsize 10.8} & \cellcolor[RGB]{88,137,245}{\scriptsize 44.1} & \cellcolor[RGB]{92,140,245}{\scriptsize 42.1} & \cellcolor[RGB]{104,148,246}{\scriptsize 36.4} & \cellcolor[RGB]{85,135,245}{\scriptsize 46.1} & \cellcolor[RGB]{152,182,248}{\scriptsize 16.6\scalebox{0.55}{$\pm$16}} \\
& {\scriptsize Phi-4} & \cellcolor[RGB]{255,255,255}{\scriptsize 0.0} & \cellcolor[RGB]{211,224,252}{\scriptsize 3.0} & \cellcolor[RGB]{163,190,249}{\scriptsize 13.3} & \cellcolor[RGB]{158,186,249}{\scriptsize 15.0} & \cellcolor[RGB]{142,175,248}{\scriptsize 20.2} & \cellcolor[RGB]{146,178,248}{\scriptsize 18.9} & \cellcolor[RGB]{238,243,254}{\scriptsize 0.5} & \cellcolor[RGB]{224,233,253}{\scriptsize 1.5} & \cellcolor[RGB]{216,227,252}{\scriptsize 2.4} & \cellcolor[RGB]{128,165,247}{\scriptsize 25.6} & \cellcolor[RGB]{229,236,253}{\scriptsize 1.1} & \cellcolor[RGB]{166,192,249}{\scriptsize 12.6} & \cellcolor[RGB]{81,132,244}{\scriptsize 48.2} & \cellcolor[RGB]{67,122,243}{\scriptsize 55.9} & \cellcolor[RGB]{92,140,245}{\scriptsize 42.3} & \cellcolor[RGB]{69,123,244}{\scriptsize 55.2} & \cellcolor[RGB]{143,176,248}{\scriptsize 19.7\scalebox{0.55}{$\pm$19}} \\
\cmidrule[0.3pt](l){2-19}
& {\scriptsize Dream-7B} & \cellcolor[RGB]{255,255,255}{\scriptsize 0.0} & \cellcolor[RGB]{225,234,253}{\scriptsize 1.4} & \cellcolor[RGB]{197,214,251}{\scriptsize 5.3} & \cellcolor[RGB]{194,212,251}{\scriptsize 5.8} & \cellcolor[RGB]{191,209,251}{\scriptsize 6.5} & \cellcolor[RGB]{191,210,251}{\scriptsize 6.4} & \cellcolor[RGB]{245,248,254}{\scriptsize 0.2} & \cellcolor[RGB]{233,239,253}{\scriptsize 0.8} & \cellcolor[RGB]{238,243,254}{\scriptsize 0.5} & \cellcolor[RGB]{186,206,250}{\scriptsize 7.6} & \cellcolor[RGB]{241,245,254}{\scriptsize 0.3} & \cellcolor[RGB]{200,216,251}{\scriptsize 4.7} & \cellcolor[RGB]{161,188,249}{\scriptsize 14.1} & \cellcolor[RGB]{174,197,250}{\scriptsize 10.5} & \cellcolor[RGB]{168,194,249}{\scriptsize 11.8} & \cellcolor[RGB]{159,187,249}{\scriptsize 14.5} & \cellcolor[RGB]{195,213,251}{\scriptsize 5.6\scalebox{0.55}{$\pm$5}} \\
& {\scriptsize LLaDA-8B} & \cellcolor[RGB]{255,255,255}{\scriptsize 0.0} & \cellcolor[RGB]{219,230,252}{\scriptsize 2.0} & \cellcolor[RGB]{178,201,250}{\scriptsize 9.2} & \cellcolor[RGB]{177,200,250}{\scriptsize 9.5} & \cellcolor[RGB]{162,189,249}{\scriptsize 13.6} & \cellcolor[RGB]{173,197,250}{\scriptsize 10.6} & \cellcolor[RGB]{241,245,254}{\scriptsize 0.3} & \cellcolor[RGB]{229,236,253}{\scriptsize 1.1} & \cellcolor[RGB]{222,232,253}{\scriptsize 1.7} & \cellcolor[RGB]{162,189,249}{\scriptsize 13.6} & \cellcolor[RGB]{231,238,253}{\scriptsize 0.9} & \cellcolor[RGB]{182,204,250}{\scriptsize 8.3} & \cellcolor[RGB]{119,159,247}{\scriptsize 29.2} & \cellcolor[RGB]{129,166,247}{\scriptsize 25.3} & \cellcolor[RGB]{142,175,248}{\scriptsize 20.3} & \cellcolor[RGB]{119,159,247}{\scriptsize 29.4} & \cellcolor[RGB]{172,196,250}{\scriptsize 10.9\scalebox{0.55}{$\pm$10}} \\
\cmidrule[0.3pt](l){2-19}
& {\scriptsize C-Haiku-4.5} & \cellcolor[RGB]{255,255,255}{\scriptsize 0.0} & \cellcolor[RGB]{213,225,252}{\scriptsize 2.7} & \cellcolor[RGB]{167,193,249}{\scriptsize 12.1} & \cellcolor[RGB]{157,185,249}{\scriptsize 15.3} & \cellcolor[RGB]{144,176,248}{\scriptsize 19.5} & \cellcolor[RGB]{143,175,248}{\scriptsize 20.0} & \cellcolor[RGB]{241,245,254}{\scriptsize 0.3} & \cellcolor[RGB]{222,232,253}{\scriptsize 1.7} & \cellcolor[RGB]{217,228,252}{\scriptsize 2.3} & \cellcolor[RGB]{128,165,247}{\scriptsize 25.6} & \cellcolor[RGB]{229,236,253}{\scriptsize 1.1} & \cellcolor[RGB]{175,199,250}{\scriptsize 10.0} & \cellcolor[RGB]{79,131,244}{\scriptsize 48.9} & \cellcolor[RGB]{41,104,242}{\scriptsize 72.9} & \cellcolor[RGB]{85,135,245}{\scriptsize 46.1} & \cellcolor[RGB]{55,114,243}{\scriptsize 63.5} & \cellcolor[RGB]{139,173,248}{\scriptsize 21.4\scalebox{0.55}{$\pm$23}} \\
& {\scriptsize GPT-5} & \cellcolor[RGB]{255,255,255}{\scriptsize 0.0} & \cellcolor[RGB]{206,220,252}{\scriptsize 3.8} & \cellcolor[RGB]{161,188,249}{\scriptsize 14.1} & \cellcolor[RGB]{151,181,248}{\scriptsize 17.1} & \cellcolor[RGB]{140,173,248}{\scriptsize 21.1} & \cellcolor[RGB]{128,165,247}{\scriptsize 25.5} & \cellcolor[RGB]{233,239,253}{\scriptsize 0.8} & \cellcolor[RGB]{218,229,252}{\scriptsize 2.1} & \cellcolor[RGB]{213,225,252}{\scriptsize 2.7} & \cellcolor[RGB]{121,160,247}{\scriptsize 28.6} & \cellcolor[RGB]{227,235,253}{\scriptsize 1.2} & \cellcolor[RGB]{162,189,249}{\scriptsize 13.6} & \cellcolor[RGB]{61,118,243}{\scriptsize 60.0} & \cellcolor[RGB]{38,102,242}{\scriptsize 90.8} & \cellcolor[RGB]{68,123,244}{\scriptsize 55.8} & \cellcolor[RGB]{45,107,242}{\scriptsize 70.0} & \cellcolor[RGB]{128,165,247}{\scriptsize 25.4\scalebox{0.55}{$\pm$28}} \\
& {\scriptsize G-3-Flash} & \cellcolor[RGB]{255,255,255}{\scriptsize 0.0} & \cellcolor[RGB]{210,223,252}{\scriptsize 3.2} & \cellcolor[RGB]{165,191,249}{\scriptsize 12.7} & \cellcolor[RGB]{155,184,249}{\scriptsize 15.8} & \cellcolor[RGB]{144,177,248}{\scriptsize 19.4} & \cellcolor[RGB]{141,174,248}{\scriptsize 20.6} & \cellcolor[RGB]{238,243,254}{\scriptsize 0.5} & \cellcolor[RGB]{224,233,253}{\scriptsize 1.5} & \cellcolor[RGB]{214,226,252}{\scriptsize 2.6} & \cellcolor[RGB]{124,162,247}{\scriptsize 27.1} & \cellcolor[RGB]{227,235,253}{\scriptsize 1.2} & \cellcolor[RGB]{167,193,249}{\scriptsize 12.1} & \cellcolor[RGB]{70,125,244}{\scriptsize 54.1} & \cellcolor[RGB]{38,102,242}{\scriptsize 86.5} & \cellcolor[RGB]{72,126,244}{\scriptsize 53.2} & \cellcolor[RGB]{46,107,242}{\scriptsize 69.7} & \cellcolor[RGB]{133,168,247}{\scriptsize 23.8\scalebox{0.55}{$\pm$26}} \\
& {\scriptsize G-3.1-Pro} & \cellcolor[RGB]{255,255,255}{\scriptsize 0.0} & \cellcolor[RGB]{217,228,252}{\scriptsize 2.3} & \cellcolor[RGB]{162,189,249}{\scriptsize 13.8} & \cellcolor[RGB]{154,183,249}{\scriptsize 16.2} & \cellcolor[RGB]{145,177,248}{\scriptsize 19.2} & \cellcolor[RGB]{139,173,248}{\scriptsize 21.4} & \cellcolor[RGB]{235,241,253}{\scriptsize 0.6} & \cellcolor[RGB]{222,232,253}{\scriptsize 1.7} & \cellcolor[RGB]{213,225,252}{\scriptsize 2.7} & \cellcolor[RGB]{120,159,247}{\scriptsize 28.9} & \cellcolor[RGB]{227,235,253}{\scriptsize 1.2} & \cellcolor[RGB]{165,191,249}{\scriptsize 12.9} & \cellcolor[RGB]{66,121,243}{\scriptsize 57.0} & \cellcolor[RGB]{38,102,242}{\scriptsize 90.3} & \cellcolor[RGB]{74,127,244}{\scriptsize 52.0} & \cellcolor[RGB]{48,109,242}{\scriptsize 68.0} & \cellcolor[RGB]{131,167,247}{\scriptsize 24.3\scalebox{0.55}{$\pm$27}} \\
\cmidrule[0.5pt](l){2-19}
& {\scriptsize Gen.\ quality} & \cellcolor[RGB]{255,255,255}{\scriptsize 0.0\scalebox{0.55}{$\pm$0}} & \cellcolor[RGB]{216,228,252}{\scriptsize 2.3\scalebox{0.55}{$\pm$1}} & \cellcolor[RGB]{175,199,250}{\scriptsize 10.0\scalebox{0.55}{$\pm$4}} & \cellcolor[RGB]{169,194,249}{\scriptsize 11.7\scalebox{0.55}{$\pm$4}} & \cellcolor[RGB]{160,188,249}{\scriptsize 14.2\scalebox{0.55}{$\pm$6}} & \cellcolor[RGB]{160,188,249}{\scriptsize 14.3\scalebox{0.55}{$\pm$7}} & \cellcolor[RGB]{239,243,254}{\scriptsize 0.4\scalebox{0.55}{$\pm$0}} & \cellcolor[RGB]{227,235,253}{\scriptsize 1.2\scalebox{0.55}{$\pm$1}} & \cellcolor[RGB]{222,232,253}{\scriptsize 1.7\scalebox{0.55}{$\pm$1}} & \cellcolor[RGB]{146,178,248}{\scriptsize 18.6\scalebox{0.55}{$\pm$8}} & \cellcolor[RGB]{232,239,253}{\scriptsize 0.8\scalebox{0.55}{$\pm$0}} & \cellcolor[RGB]{179,201,250}{\scriptsize 9.0\scalebox{0.55}{$\pm$3}} & \cellcolor[RGB]{102,147,246}{\scriptsize 37.1\scalebox{0.55}{$\pm$16}} & \cellcolor[RGB]{92,140,245}{\scriptsize 42.1\scalebox{0.55}{$\pm$29}} & \cellcolor[RGB]{116,157,246}{\scriptsize 30.6\scalebox{0.55}{$\pm$17}} & \cellcolor[RGB]{98,144,245}{\scriptsize 39.0\scalebox{0.55}{$\pm$22}} &  \\
\bottomrule
\end{tabular}
}%  end resizebox
\end{table}

% ─────────────────────────────────────────────────────────────────────────────
\section{SHAP Attribution Methodology and Additional Experiments}
\label{app:shap}
% ─────────────────────────────────────────────────────────────────────────────

To quantify which features drive round-trip success, we fit a
gradient-boosted decision tree (GBT) classifier to the binary outcome
(correct/incorrect) of each \emph{triple} (expression, generator,
extractor).

\textbf{Model.}
We use scikit-learn's~\citep{scikit-learn} \texttt{GradientBoostingClassifier} with 200
trees, maximum depth~4, learning rate~0.1, and row subsampling rate~0.8.
The classifier is trained on the full set of triples. A held-out 80/20
split yields a sanity-check accuracy of 88.1\%, confirming the features
carry substantial predictive signal.  For context, the majority-class
baseline (always predicting failure) achieves 87.8\%, so the GBT's
improvement is modest in absolute terms, and the value of the classifier
lies in the SHAP decomposition rather than raw accuracy.  The
corresponding AUC-ROC scores are $0.812$ for the high-level
classifier and $0.822$ for the AST classifier, consistent with a
model that captures substantial but not saturating signal on the
round-trip outcome.

\textbf{Attribution.}
We compute exact SHAP values via \texttt{TreeExplainer}~\citep{lundberg2019explainable},
which exploits the tree structure to compute Shapley values in
polynomial time without sampling.  For each feature~$f$, we report
\emph{mean absolute SHAP} normalised to a percentage of the total
attribution:
\begin{equation}
  \mathrm{Attribution}(f)
  \;=\;
  \frac{
    \frac{1}{n}\sum_{t=1}^{n} |\phi_f^{(t)}|
  }{
    \sum_{f'} \frac{1}{n}\sum_{t=1}^{n} |\phi_{f'}^{(t)}|
  }
  \;\times\; 100\%,
\end{equation}
where $\phi_f^{(t)}$ is the SHAP value of feature~$f$ for triple~$t$
and the sum in the denominator runs over all features.

\textbf{Direction.}
We additionally report the Pearson correlation between each feature's
raw values and its SHAP values across triples.  A positive correlation
means higher feature values push the prediction toward success; a
negative correlation means higher values push toward failure.

% \textbf{Two feature sets.}
% We run the analysis twice: once with six \emph{global} features
% (generator size, extractor size, family match, depth, operator count,
% $R_{\text{math}}$) and once with the fine-grained \emph{AST} features
% defined in \Cref{sec:features} (per-operator counts, skeleton identity,
% center of mass, variable reuse, etc.).  The global analysis identifies
% broad drivers; the AST analysis pinpoints which structural properties
% matter most.

% ─────────────────────────────────────────────────────────────────────────────
\subsection{Complete Structural SHAP Attribution}
\label{app:shap_full}
% ─────────────────────────────────────────────────────────────────────────────

The main text (\Cref{sec:why_fails}) reports the high-level SHAP
analysis with six summary features and summarizes the key findings of
the structural analysis. \Cref{tab:shap_full} lists all 16 features
from the AST-level gradient-boosted classifier, sorted by mean
absolute SHAP value.

After controlling for model size ($50.8\%$ combined), four structural
findings stand out:
\begin{itemize}
  \item \textbf{Division is the hardest operator.}
    Division ($n_{\div}$: 6.4\%) is the single largest structural
    predictor, followed by subtraction ($n_{-}$: 4.1\%) and
    multiplication ($n_{\times}$: 3.7\%); addition ($n_{+}$: 1.4\%)
    matters least. The two non-commutative operators thus rank
    first and second among the four. Non-commutativity, not
    multiplicativity, drives operator difficulty: argument order
    ($a \div b \neq b \div a$, $a - b \neq b - a$) must be conveyed
    unambiguously in prose, whereas the commutative operators tolerate
    reordering. Consistent with this, the deepest tree level
    containing a non-commutative operator (5.5\%, $\downarrow$) is the
    largest structural predictor after depth itself, confirming that
    ordering ambiguity compounds with nesting.
  \item \textbf{Tree shape matters beyond size and depth.} Skeleton
    identity (5.5\%) is a strong predictor alongside depth (7.4\%),
    indicating that expressions with different branching topologies
    but identical $(k,d)$ can differ substantially in round-trip
    accuracy. Center of mass (3.0\%, $\downarrow$) reveals the
    direction of this effect: right-branching trees, which require
    the word problem to hold an incomplete computation open while
    describing a nested subexpression, tend to be harder
    than left-branching ones.
  \item \textbf{Variable reuse introduces ambiguity.} Expressions
    where a variable appears in multiple leaves (3.5\%,
    $\downarrow$) require coreference in the word problem, forcing
    the generator to signal that two quantities refer to the same
    variable. More unique variables (3.0\%, $\uparrow$) have the
    opposite effect, reducing coreference demands and improving
    round-trip fidelity.
  \item \textbf{Root operator sets the frame.} Addition as root
    (2.2\%, $\downarrow$) and subtraction as root (2.0\%,
    $\downarrow$) are the most penalising root choices, whereas
    multiplicative roots ($\div$: 0.9\%, $\times$: 0.7\%, both
    $\uparrow$) provide more flexibility at the top of the tree.
\end{itemize}

\begin{table}[htb]
\centering
\caption{Complete SHAP feature attribution for the structural analysis.
All features from the AST-level gradient-boosted classifier are shown,
sorted by mean $|\text{SHAP}|$ as a percentage of total attribution.
Green ($\uparrow$) indicates higher feature values increase
round-trip success; red ($\downarrow$) indicates higher values
decrease success.}
\label{tab:shap_full}
\small
\setlength{\tabcolsep}{4pt}
\begin{tabular}{l r}
\toprule
Feature & Attribution \\
\midrule
{\scriptsize Generator size} & \cellcolor[RGB]{46,125,34}{\scriptsize 34.3\% $\uparrow$} \\
{\scriptsize Extractor size} & \cellcolor[RGB]{154,192,148}{\scriptsize 16.5\% $\uparrow$} \\
{\scriptsize Depth} & \cellcolor[RGB]{242,208,208}{\scriptsize 7.4\% $\downarrow$} \\
{\scriptsize $n_{\div}$} & \cellcolor[RGB]{244,215,215}{\scriptsize 6.4\% $\downarrow$} \\
{\scriptsize Non-comm.\ depth} & \cellcolor[RGB]{245,220,220}{\scriptsize 5.5\% $\downarrow$} \\
{\scriptsize Skeleton ID} & \cellcolor[RGB]{245,220,220}{\scriptsize 5.5\% $\downarrow$} \\
{\scriptsize $n_{-}$} & \cellcolor[RGB]{248,229,229}{\scriptsize 4.1\% $\downarrow$} \\
{\scriptsize $n_{\times}$} & \cellcolor[RGB]{248,231,231}{\scriptsize 3.7\% $\downarrow$} \\
{\scriptsize Variable reuse} & \cellcolor[RGB]{249,232,232}{\scriptsize 3.5\% $\downarrow$} \\
{\scriptsize Center of mass} & \cellcolor[RGB]{249,236,236}{\scriptsize 3.0\% $\downarrow$} \\
{\scriptsize Unique variables} & \cellcolor[RGB]{236,243,235}{\scriptsize 3.0\% $\uparrow$} \\
{\scriptsize Root $=$\,+} & \cellcolor[RGB]{251,241,241}{\scriptsize 2.2\% $\downarrow$} \\
{\scriptsize Root $=$\,$-$} & \cellcolor[RGB]{251,242,242}{\scriptsize 2.0\% $\downarrow$} \\
{\scriptsize $n_{+}$} & \cellcolor[RGB]{246,249,245}{\scriptsize 1.4\% $\uparrow$} \\
{\scriptsize Root $=$\,$\div$} & \cellcolor[RGB]{249,251,249}{\scriptsize 0.9\% $\uparrow$} \\
{\scriptsize Root $=$\,$\times$} & \cellcolor[RGB]{250,252,250}{\scriptsize 0.7\% $\uparrow$} \\
\bottomrule
\end{tabular}
\end{table}

% ─────────────────────────────────────────────────────────────────────────────
\section{Skeleton Difficulty and Center of Mass}
\label{app:com}
% ─────────────────────────────────────────────────────────────────────────────

The SHAP analysis identifies skeleton identity (5.5\%) and center of
mass (3.0\%) as significant predictors of difficulty.  We now examine
these effects directly by comparing skeletons that share the same
operator count and depth.

\begin{figure}[htb]
  \centering
  \includegraphics[width=\linewidth]{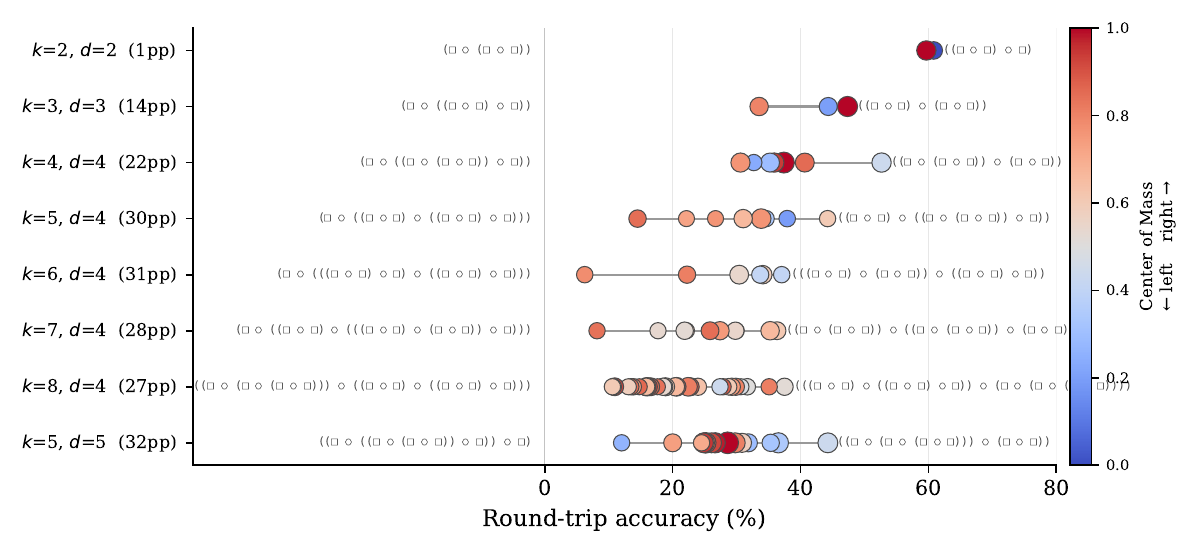}
  \caption{Skeleton difficulty within $(k, d)$ groups.  Each point is
  one skeleton; the $x$-axis shows its round-trip accuracy averaged
  across all models and expressions sharing that skeleton.  Colour
  encodes center of mass (blue = left-branching, red =
  right-branching).  The spread in parentheses on each row label shows
  the gap between the easiest and hardest skeleton in that group.
  Within-group weighted $r = -0.148$ (CoM vs.\ accuracy, $p = 0.007$, permutation test).}
  \label{fig:skeleton}
\end{figure}

\Cref{fig:skeleton} plots the round-trip accuracy of each skeleton,
grouped by $(k, d)$.  The spread within groups is substantial: at
$(k{=}4, d{=}4)$, the seven observed skeletons span 22.1 percentage
points, from 52.7\% for the easiest to 30.6\% for the hardest.  At
$(k{=}2, d{=}2)$, the only two skeletons differ by 1.1 percentage
points (left-branching: 60.8\% vs.\ right-branching: 59.7\%).  These
differences are controlled for operator count and depth by
construction; they reflect the effect of tree shape alone.

\textbf{Center of mass as predictor.}
Points in \Cref{fig:skeleton} are coloured by their normalised center
of mass (CoM), ranging from blue (left-branching, CoM~$= 0$) to red
(right-branching, CoM~$= 1$).  A visual pattern emerges: within each
group, bluer points tend to appear to the right (higher accuracy) and
redder points to the left (lower accuracy).  The within-group weighted
correlation between CoM and accuracy is $r = -0.148$ ($p = 0.007$,
one-tailed permutation test with 10000 shuffles, $N = 455$ skeletons),
indicating that right-branching trees tend to be harder after
controlling for $(k, d)$.

\textbf{Why right-branching is harder.}
Left-branching trees (e.g.\ $((A + B) \times C) - D$) correspond to
a sequence of operations applied to a running result, which maps
naturally onto left-to-right prose: ``start with $A$ plus $B$, multiply
by $C$, subtract $D$.''  Right-branching trees (e.g.\ $A - (B \times
(C + D))$) require the generator to introduce a nested subcomputation
before the main operation, demanding forward references or
subordinate clauses that are harder to produce and parse unambiguously.
An analogous right-branching deficit has been reported for LSTM/GRU
learners on synthetic interpreted languages~\citep{paperno2022interpreted},
suggesting the bias is not specific to transformer LLMs.

% ─────────────────────────────────────────────────────────────────────────────
\section{Error Taxonomy Details}
\label{app:error_taxonomy}
% ─────────────────────────────────────────────────────────────────────────────

\Cref{tab:error_taxonomy} shows the per-extractor error profile across
all generators and depths.  The classification algorithm processes each
incorrect extraction in the following order: (1)~attempt to parse
$\exprest$ into a symbolic expression tree; if parsing fails, label
\errparse{}; (2)~count the operators in both $\expr$ and
$\exprest$; if they differ, label \errarity{};
(3)~compare the unlabelled tree skeletons; if they differ, label
\errstruct{}; (4)~otherwise label \errcontent{}
(the skeleton matches but leaf values or operator labels differ).

\begin{table}[htb]
\centering
\caption{Per-extractor error profile.  Each row shows the
percentage of that extractor's errors falling into each
category, aggregated across all generators and depths.
Models are grouped by family and sorted by size.}
\label{tab:error_taxonomy}
\small
\setlength{\tabcolsep}{5pt}
\begin{tabular}{l rrrr}
\toprule
Extractor & $\mathrm{Err}_{\mathrm{parse}}$ & $\mathrm{Err}_{\mathrm{arity}}$ & $\mathrm{Err}_{\mathrm{struct}}$ & $\mathrm{Err}_{\mathrm{content}}$ \\
\midrule
Qwen3-0.6B & 0.1\% & 89.1\% & 10.3\% & 0.5\% \\
Qwen3-1.7B & 0.4\% & 74.4\% & 23.4\% & 1.8\% \\
Qwen3-4B & 1.2\% & 67.8\% & 29.3\% & 1.7\% \\
Qwen3-8B & 1.6\% & 66.2\% & 30.7\% & 1.6\% \\
Qwen3-14B & 1.8\% & 65.7\% & 31.3\% & 1.2\% \\
Qwen3-32B & 1.6\% & 64.7\% & 32.4\% & 1.2\% \\
Gemma-3-4B & 5.4\% & 67.9\% & 25.5\% & 1.1\% \\
Gemma-3-12B & 0.3\% & 70.6\% & 27.9\% & 1.3\% \\
Gemma-3-27B & 0.6\% & 68.0\% & 29.8\% & 1.6\% \\
Phi-4 & 2.3\% & 67.0\% & 28.8\% & 1.9\% \\
Dream-7B & 10.5\% & 69.6\% & 18.6\% & 1.3\% \\
LLaDA-8B & 2.6\% & 77.2\% & 17.6\% & 2.5\% \\
C-Haiku-4.5 & 1.9\% & 67.0\% & 29.7\% & 1.5\% \\
GPT-5 & 2.1\% & 70.3\% & 26.2\% & 1.4\% \\
Gemini-3-Flash & 0.6\% & 73.3\% & 24.9\% & 1.2\% \\
Gemini-3.1-Pro & 1.4\% & 71.0\% & 26.3\% & 1.2\% \\
\bottomrule
\end{tabular}
\end{table}

\begin{figure}[htb]
  \centering
  \includegraphics[width=0.55\linewidth]{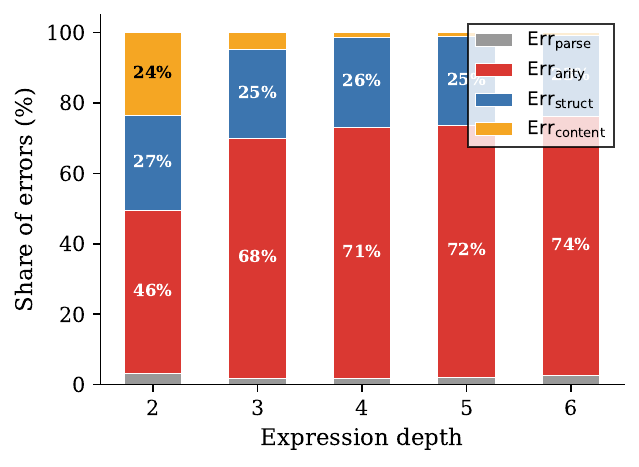}
  \caption{Distribution of error categories by expression depth.
  \errarity{} (lost or extra operators) dominates at every
  depth and grows from 46\% at depth~2 to 74\% at depth~6.
  \errcontent{} shrinks correspondingly, indicating that
  models that preserve the correct operator count and tree shape
  rarely misidentify leaves.  \errparse{} errors are
  small (2.0\% overall).}
  \label{fig:error_taxonomy}
\end{figure}

\Cref{fig:error_taxonomy} breaks the same categories down by expression depth, showing that arity errors grow with depth while content errors shrink.

% ─────────────────────────────────────────────────────────────────────────────
\section{Correlation with External Benchmarks}
\label{app:correlation}
% ─────────────────────────────────────────────────────────────────────────────

A compact summary appears in \Cref{sec:scaling_ext}; this appendix
provides the full per-benchmark breakdown.

We compute Spearman rank correlations between per-model round-trip
scores and published results on 9 external benchmarks spanning four
categories:
math (GSM8K~\citep{cobbe2021gsm8k}, MATH~\citep{hendrycks2021math},
MGSM~\citep{shi2023mgsm}),
code (MBPP~\citep{austin2021mbpp}, LiveCodeBench~\citep{jain2024livecodebench}),
general knowledge (MMLU~\citep{hendrycks2021mmlu},
MMLU-Pro~\citep{wang2024mmlupro}, BBH~\citep{suzgun2023bbh}),
and reasoning (GPQA~\citep{rein2023gpqagraduatelevelgoogleproofqa}). Benchmark scores are drawn exclusively from official
technical reports and vendor announcements
\citep{qwen3,gemma3,phi4,anthropic2025claude37,claude_haiku_4_5,singh2025openai};
benchmarks with verified scores for fewer than seven models are excluded.
For each benchmark we correlate its model rankings with three
round-trip scores (\Cref{sec:matrix}):
\emph{generation quality} (column-average of the communication matrix),
\emph{extraction quality} (row-average), and
\emph{self-communication} (diagonal cell).
Because frontier vendors did not publish standard-mode scores on most
benchmarks, the analysis is dominated by the 10 open-weight models;
each benchmark uses pairwise deletion, so the effective $n$ varies
(7--10; see \Cref{tab:correlation_benchmarks}).
\Cref{tab:benchmark_provenance} lists the source document, evaluation
protocol, and coverage for each model family.
Because evaluation protocols differ across families (e.g.\ base-model
few-shot for Qwen3 vs.\ 0-shot instruct for Gemma~3), the
correlations should be read as rough indicators rather than precise
estimates.

\begin{table}[htb]
\centering
\caption{Provenance of external benchmark scores. Each score traces to
the listed source; \textbf{null} entries are omitted from the
correlation analysis via pairwise deletion.}
\label{tab:benchmark_provenance}
\small
\setlength{\tabcolsep}{3pt}
\begin{tabular}{@{}l l l l@{}}
\toprule
Model family & Source & Eval mode & Benchmarks (of 9) \\
\midrule
Qwen3 (0.6--32B) & \citet{qwen3} & base few-shot\textsuperscript{\dag} & 9/9 \\
Gemma~3 IT (4--27B) & \citet{gemma3} & 0-shot instruct & 7/9 \\
Phi-4 (14B) & \citet{phi4} & simple-evals & 5/9 \\
Claude Haiku~4.5 & \citet{claude_haiku_4_5} & \multicolumn{2}{l}{no standard-mode scores published} \\
GPT-5 & \citet{singh2025openai} & non-reasoning & 1/9 \\
\bottomrule
\end{tabular}
\par\smallskip
{\scriptsize \textsuperscript{\dag}MATH-500, LiveCodeBench, and GPQA-Diamond use instruct non-thinking scores
where base-model scores are unavailable.}
\end{table}

\Cref{tab:correlation_benchmarks} reports the per-benchmark correlations
sorted by average $\rho$ across the three round-trip phases.
Several patterns emerge:

\textbf{Extraction quality is consistently well-predicted.}
For 7 of 9 benchmarks, extraction $\rho$ reaches $0.83$ or higher,
peaking at
MMLU ($\rho = 0.98$) and MGSM ($\rho = 0.96$). The models that
extract expressions most accurately are, broadly, those that score
highest on standard evaluations.

\textbf{Generation quality is harder to predict.}
Generation $\rho$ ranges from $0.47$ (MMLU) to $0.89$ (MGSM) and
fails to reach significance for MMLU, MMLU-Pro, MATH, and
GPQA-Diamond individually. The gap between extraction and generation
$\rho$ is largest for MMLU ($0.98$ vs.\ $0.47$) and GPQA-Diamond
($0.95$ vs.\ $0.50$), confirming that faithfully \emph{generating}
a word problem from an expression is a skill not fully predicted by
standard evaluations.

\textbf{Code benchmarks correlate strongly.}
MBPP and LiveCodeBench both reach $\rho = 0.93$ for extraction and
$0.77$ for generation, placing code among the best-predicted
categories rather than the weakest.

\textbf{Math benchmarks are the weakest category.}
MATH shows uniformly moderate correlations
($\rho = 0.52$/$0.50$/$0.30$ for generation/extraction/self-communication),
and GSM8K follows a similar pattern ($0.68$/$0.72$/$0.50$).
MGSM is an outlier ($0.89$/$0.96$/$0.89$, $n{=}7$).
Standard math-solving ability overlaps only partially with the
structural linearization our protocol measures.

\Cref{tab:correlation_categories} aggregates these findings at the
category level via Fisher-$z$ averaging of per-benchmark $\rho$
values. Extraction correlates strongly across all four categories
($\rho = 0.82$--$0.95$). Generation is highest for code
($\rho = 0.77$) and math ($\rho = 0.74$) and lowest for reasoning
and general knowledge ($\rho = 0.50$--$0.55$).

\begin{table}[htb]
\centering
\caption{Fisher-$z$ averaged Spearman $\rho$ between round-trip scores and external benchmarks, grouped by category: {\color[HTML]{E53935}\textbf{Math}}, {\color[HTML]{1E88E5}\textbf{Code}}, {\color[HTML]{43A047}\textbf{General}}, {\color[HTML]{8E24AA}\textbf{Reasoning}}.}
\vspace{2mm}
\label{tab:correlation_categories}
\small
\setlength{\tabcolsep}{4pt}
\begin{tabular}{l c c c c}
\toprule
 & {\color[HTML]{1E88E5}\textbf{Code}} & {\color[HTML]{8E24AA}\textbf{Reasoning}} & {\color[HTML]{43A047}\textbf{General}} & {\color[HTML]{E53935}\textbf{Math}} \\
{\scriptsize benchmarks} & {\scriptsize 2} & {\scriptsize 1} & {\scriptsize 3} & {\scriptsize 3} \\
\midrule
{\scriptsize Generation Acc.} & \cellcolor[RGB]{102,137,235}{\color{black}\scriptsize 0.77} & \cellcolor[RGB]{154,178,242}{\color{black}\scriptsize 0.50} & \cellcolor[RGB]{144,170,240}{\color{black}\scriptsize 0.55} & \cellcolor[RGB]{108,142,236}{\color{black}\scriptsize 0.74} \\
{\scriptsize Extraction Acc.} & \cellcolor[RGB]{69,112,231}{\color{black}\scriptsize 0.93} & \cellcolor[RGB]{65,109,230}{\color{black}\scriptsize 0.95} & \cellcolor[RGB]{67,110,230}{\color{black}\scriptsize 0.94} & \cellcolor[RGB]{92,129,234}{\color{black}\scriptsize 0.82} \\
{\scriptsize Self-Consistency} & \cellcolor[RGB]{102,137,235}{\color{black}\scriptsize 0.77} & \cellcolor[RGB]{75,116,231}{\color{black}\scriptsize 0.90} & \cellcolor[RGB]{84,123,233}{\color{black}\scriptsize 0.86} & \cellcolor[RGB]{127,156,238}{\color{black}\scriptsize 0.64} \\
\bottomrule
\end{tabular}
\end{table}

\begin{table}[htb]
\centering
\caption{Spearman rank correlation ($\rho$) between round-trip scores and external benchmarks, sorted by average $\rho$. Columns are colored by category: {\color[HTML]{E53935}\textbf{Math}}, {\color[HTML]{1E88E5}\textbf{Code}}, {\color[HTML]{43A047}\textbf{General}}, {\color[HTML]{8E24AA}\textbf{Reasoning}}.}
\vspace{2mm}
\label{tab:correlation_benchmarks}
\small
\setlength{\tabcolsep}{2pt}
\begin{tabular}{l c c c c c c c c c}
\toprule
 & \rotatebox{90}{\scriptsize \color[HTML]{E53935}\textbf{MGSM}} & \rotatebox{90}{\scriptsize \color[HTML]{1E88E5}\textbf{LiveCodeBench}} & \rotatebox{90}{\scriptsize \color[HTML]{1E88E5}\textbf{MBPP}} & \rotatebox{90}{\scriptsize \color[HTML]{43A047}\textbf{MMLU}} & \rotatebox{90}{\scriptsize \color[HTML]{8E24AA}\textbf{GPQA-Diamond}} & \rotatebox{90}{\scriptsize \color[HTML]{43A047}\textbf{MMLU-Pro}} & \rotatebox{90}{\scriptsize \color[HTML]{43A047}\textbf{BBH}} & \rotatebox{90}{\scriptsize \color[HTML]{E53935}\textbf{GSM8K}} & \rotatebox{90}{\scriptsize \color[HTML]{E53935}\textbf{MATH}} \\
\midrule
{\scriptsize $n$} & {\scriptsize 7} & {\scriptsize 9} & {\scriptsize 9} & {\scriptsize 10} & {\scriptsize 10} & {\scriptsize 10} & {\scriptsize 9} & {\scriptsize 9} & {\scriptsize 10} \\
\midrule[0.25pt]
{\scriptsize Generation Acc.} & \cellcolor[RGB]{77,118,232}{\color{black}\scriptsize 0.89} & \cellcolor[RGB]{102,137,235}{\color{black}\scriptsize 0.77} & \cellcolor[RGB]{102,137,235}{\color{black}\scriptsize 0.77} & \cellcolor[RGB]{160,182,242}{\color{black}\scriptsize 0.47} & \cellcolor[RGB]{154,178,242}{\color{black}\scriptsize 0.50} & \cellcolor[RGB]{154,178,242}{\color{black}\scriptsize 0.50} & \cellcolor[RGB]{122,153,238}{\color{black}\scriptsize 0.67} & \cellcolor[RGB]{119,150,237}{\color{black}\scriptsize 0.68} & \cellcolor[RGB]{152,176,241}{\color{black}\scriptsize 0.52} \\
{\scriptsize Extraction Acc.} & \cellcolor[RGB]{63,107,230}{\color{black}\scriptsize 0.96} & \cellcolor[RGB]{69,112,231}{\color{black}\scriptsize 0.93} & \cellcolor[RGB]{69,112,231}{\color{black}\scriptsize 0.93} & \cellcolor[RGB]{59,104,229}{\color{black}\scriptsize 0.98} & \cellcolor[RGB]{65,109,230}{\color{black}\scriptsize 0.95} & \cellcolor[RGB]{70,113,231}{\color{black}\scriptsize 0.93} & \cellcolor[RGB]{89,127,233}{\color{black}\scriptsize 0.83} & \cellcolor[RGB]{112,145,236}{\color{black}\scriptsize 0.72} & \cellcolor[RGB]{154,178,242}{\color{black}\scriptsize 0.50} \\
{\scriptsize Self-Consistency} & \cellcolor[RGB]{77,118,232}{\color{black}\scriptsize 0.89} & \cellcolor[RGB]{102,137,235}{\color{black}\scriptsize 0.77} & \cellcolor[RGB]{102,137,235}{\color{black}\scriptsize 0.77} & \cellcolor[RGB]{62,107,230}{\color{black}\scriptsize 0.97} & \cellcolor[RGB]{75,116,231}{\color{black}\scriptsize 0.90} & \cellcolor[RGB]{97,133,234}{\color{black}\scriptsize 0.79} & \cellcolor[RGB]{129,158,238}{\color{black}\scriptsize 0.63} & \cellcolor[RGB]{155,178,242}{\color{black}\scriptsize 0.50} & \cellcolor[RGB]{195,209,247}{\color{black}\scriptsize 0.30} \\
\bottomrule
\end{tabular}
\end{table}

% ─────────────────────────────────────────────────────────────────────────────
\section{Round-Trip Examples}
\label{app:examples}
% ─────────────────────────────────────────────────────────────────────────────

\Cref{tab:examples} presents example round-trips generated by
Qwen3-32B and frontier models (as both $\Gen$ and $\Ext$) at increasing complexity.  Each
row shows the skeleton (tree shape), the source expression, the
generated word problem, and the expression recovered by the extractor.
All five examples were successfully verified by SymPy, despite the
recovered expressions differing syntactically from the originals
(e.g.\ reordering commutative operands or distributing subtraction).

\begin{table}[htb]
\centering
\caption{Round-trip examples at increasing operator
count~$k$ and depth~$d$.  Skeletons use the notation from
\Cref{fig:skeletons}: $\square$ = leaf, $\circ$ = operator.
The ``Recovered'' column shows the extractor's output, which SymPy
confirms as symbolically equivalent to the original expression.
The first four rows use Qwen3-32B as both $\Gen$ and $\Ext$; the
last two use frontier models.}
\label{tab:examples}
\small
\begin{tabular}{@{}l l p{5.6cm} l@{}}
\toprule
$(k,d)$ & \makecell[l]{Skeleton \\ Expression} & Word Problem & Recovered \\
\midrule

$(2,2)$ &
\makecell[l]{$(\square\circ\square)\circ\square$\\[2pt]
$(E - H) - H$} &
A farmer had \{E\} acres of land.\ He used \{H\} acres for planting
corn and another \{H\} acres for growing soybeans.\ How many acres
were left unplanted? &
$E-(H+H)$ \\[8pt]

$(3,3)$ &
\makecell[l]{$\square\circ(\square\circ(\square\circ\square))$\\[2pt]
$D-(1-(D+C))$} &
A construction site has \{D\} tons of steel.\ Workers use 1 ton for a
beam, then recover \{D\} tons from an old structure and add \{C\} tons
from storage.\ How much steel is available now? &
$D-1+D+C$ \\[8pt]

$(5,4)$ &
\makecell[l]{$(((\square\circ\square)\circ\square)\circ\square)\circ(\square\circ\square)$\\[2pt]
$(((F/3)+1)-C)-(A+9)$} &
A traveler drove \{F\} miles and then took a break after every 3
miles.\ They added 1 rest stop but skipped \{C\} due to road
closures.\ How many rest stops were left after subtracting the \{A\}
planned stops plus 9 extra ones? &
$(F/3{+}1{-}C){-}(A{+}9)$ \\[8pt]

$(6,5)$ &
\makecell[l]{$((\square\circ((\square\circ\square)\circ\square))\circ\square)\circ(\square\circ\square)$\\[2pt]
$((E-((C{\times}6){\times}B))-9)+(G/8)$} &
A research team counted \{E\} fish in a coral reef.\ They observed
that \{C\} divers each removed 6 fish per hour for \{B\} hours, and 9
fish swam away.\ Later, \{G\} baby fish were evenly distributed among
8 areas.\ How many fish remained in the reef? &
$E{-}(C{\times}6{\times}B{+}9){+}G/8$ \\
\midrule

\multicolumn{4}{@{}l}{\itshape Frontier models} \\[2pt]

$(5,5)$ &
\makecell[l]{$G{\times}(H{+}G/(C{+}H{+}C))$\\[1pt]
{\scriptsize $\Gen$/$\Ext$: Gemini-3.1-Pro}} &
A dispatcher calculates the total fleet size by combining \{C\} cargo
jets, \{H\} transport helicopters, and another \{C\} light aircraft.\
A stockpile of \{G\} crates of aviation fuel is divided evenly among
this entire fleet.\ Each aircraft takes its share of crates, adds
\{H\} emergency supply boxes to its payload, and multiplies this new
load by \{G\} scheduled flights to find the total cargo transported. &
$(H{+}(G/(C{+}H{+}C))){\times}G$ \\[8pt]

$(5,5)$ &
\makecell[l]{$C{-}F{\times}(F{-}(7{+}F{+}D))$\\[1pt]
{\scriptsize $\Gen$/$\Ext$: GPT-5}} &
In a basketball game, start with 7 bonus points, then add \{F\} for
free throws and \{D\} for assists, and take this total away from
\{F\}.\ Multiply that result by \{F\}, then take that amount away from
\{C\} to find the team's final tally. &
$C{-}((F{-}(7{+}F{+}D)){\times}F)$ \\

\bottomrule
\end{tabular}
\end{table}

% ─────────────────────────────────────────────────────────────────────────────
\section{Round-Trip Failure Examples}
\label{app:failure_examples}
% ─────────────────────────────────────────────────────────────────────────────

\Cref{tab:examples} shows successful round-trips.
\Cref{tab:failure_examples} complements it with representative failures
covering all four error categories defined in \Cref{app:error_taxonomy},
including both extractor and generator faults.
All examples are drawn from strong model pairs ($\geq$14B) and
frontier models to illustrate that these failures are not trivial.

\begin{table}[htb]
\centering
\caption{Representative round-trip failures covering all four error
categories, with a mix of extractor and generator faults.  For each
row: the source expression, the word problem written by $\Gen$, the
expression recovered by $\Ext$, and a diagnosis identifying whether
the fault lies in generation or extraction, based on the consensus
vote across the 16 extractors.  Underlining highlights the locus of
the error in the recovered expression.}
\label{tab:failure_examples}
\small
\begin{tabular}{@{}p{1.4cm} p{4.8cm} p{2.8cm} p{4.6cm}@{}}
\toprule
Category & Expression \& Word Problem & Recovered & Diagnosis \\
\midrule

\errparse
\newline
{\scriptsize $\Gen$: GPT-5}
\newline
{\scriptsize $\Ext$: GPT-5}
&
$((F+H) \times (10 + F/E))$
\newline\smallskip
{\scriptsize ``You bake \{F\} trays of cookies and \{H\} trays of
brownies, then count all the trays together.\ For each tray, you set
the glaze amount by splitting \{F\} cups evenly among \{E\} bowls and
then adding 10 more cups, and you make that amount for every tray.''}
&
\texttt{(F + H)\,\underline{(F/E + 10)}}
\newline\smallskip
{\scriptsize Juxtaposition without an explicit operator, so the
expression does not parse.}
&
{\scriptsize \textbf{Extractor fault.}
The expected expression is the product of ``trays'' and ``glaze per
tray,'' but the extractor drops the explicit $\times$ and emits two
parenthesised groups side by side, which the parser rejects.}
\\
\midrule[0.25pt]

\errarity
\newline
{\scriptsize $\Gen$: Qwen3-14B}
\newline
{\scriptsize $\Ext$: GPT-5}
&
$G + 3 + 7 - 7$
\newline\smallskip
{\scriptsize ``A soccer team scored \{G\} goals in the first half, then
scored 3 more goals, and later scored 7 additional goals before giving
up 7 in the second half.\ How many goals did they have at the end of
the game?''}
&
$(G + 3) + 7$\ \underline{\emph{[drops $-7$]}}
\newline\smallskip
{\scriptsize 2 operators instead of 3.}
&
{\scriptsize \textbf{Extractor fault.}
The word problem clearly narrates four events, including ``giving up
7,'' but the extractor stops after the third and omits the final
subtraction, effectively simplifying $+7-7$ away.}
\\
\midrule[0.25pt]

\errarity
\newline
{\scriptsize $\Gen$: Qwen3-14B}
\newline
{\scriptsize $\Ext$: Phi-4}
&
$D + F \times 9 + 8$
\newline\smallskip
{\scriptsize ``A designer creates \{D\} dresses and \{F\} fashion sets,
each set requiring 9 dresses.\ After making 8 additional dresses, how
many total dresses has the designer created?''}
&
$\underline{(C / F)}$
\newline\smallskip
{\scriptsize 1 operator instead of 3, and with wrong variable ($C$
does not appear in the problem).}
&
{\scriptsize \textbf{Generator fault.}
The narration flattens the two additions and the multiplication into a
single sentence and loses the link between ``sets,'' ``each requiring 9
dresses,'' and the final ``$+8$'' term.
The extractors, including the frontier models, fail to recover
the three-operation expression, and several emit a single division
using a variable ($C$) that never appears in the problem.}
\\
\midrule[0.25pt]

\errstruct
\newline
{\scriptsize $\Gen$: GPT-5}
\newline
{\scriptsize $\Ext$: Qwen3-14B}
&
$E + B - (1 - 9)$
\newline\smallskip
{\scriptsize ``During a forest survey, a ranger counts \{E\} birch
saplings and then adds \{B\} pine saplings to the tally.\ From this
total, they subtract the result of taking 9 away from 1 to correct an
earlier note.\ How many saplings are recorded now?''}
&
$(E + B) - \underline{(9 - 1)}$
\newline\smallskip
{\scriptsize Same three operators ($+$, $-$, $-$), with operands of the
inner subtraction swapped.}
&
{\scriptsize \textbf{Extractor fault.}
The word problem states ``taking 9 away from 1,'' i.e.\ $1 - 9$, but
the extractor flips the operands to the more natural $9 - 1$,
preserving the outer structure while inverting the inner subtraction.}
\\
\midrule[0.25pt]

\errcontent
\newline
{\scriptsize $\Gen$: Qwen3-14B}
\newline
{\scriptsize $\Ext$: GPT-5}
&
$B \times F + E - D$
\newline\smallskip
{\scriptsize ``During a cycling race, \{B\} cyclists each rode \{F\}
laps around the track.\ They then received an extra \{E\} bonus points
for completing the race, but \{D\} points were deducted for penalties.
What is the total points earned by all cyclists?''}
&
$B \times \underline{(F + E - D)}$
\newline\smallskip
{\scriptsize Same skeleton after canonicalisation, with $B$ distributed
over a parenthesised sum rather than multiplied only with $F$.}
&
{\scriptsize \textbf{Extractor fault.}
The word problem distinguishes ``laps per cyclist'' ($B \times F$)
from the additive bonus $+E$ and penalty $-D$, but the extractor binds
all three addends under the multiplication, changing the value at
every operand assignment.
Most extractors recover the intended grouping despite the mild
unit slip (laps vs.\ points), indicating the English is clear enough
for most readers.}
\\
\midrule[0.25pt]

\errcontent
\newline
{\scriptsize $\Gen$: Qwen3-32B}
\newline
{\scriptsize $\Ext$: Claude Haiku 4.5}
&
$F + 8 + E$
\newline\smallskip
{\scriptsize ``\{F\} flower bulbs were planted in the spring, and by
summer there were 8 more blooming flowers.\ Then, \{E\} extra flowers
were added to the garden bed.''}
&
$\underline{D} + 8 + E$
\newline\smallskip
{\scriptsize Same skeleton, with leaf $F$ replaced by $D$.}
&
{\scriptsize \textbf{Generator fault.}
The word problem starts with ``\{F\} flower bulbs,'' but the surface
form places the variable in sentence-initial position with no
anchoring cue, and extractors consistently read the leading token as
$D$.
The extractors exhibit the same $F\!\to\!D$ confusion: they
recover the tree skeleton correctly but mis-bind the leading
variable, a universal failure that points to the surface form rather
than any individual extractor.}
\\

\bottomrule
\end{tabular}
\end{table}

% ─────────────────────────────────────────────────────────────────────────────
\section{Fault Attribution Details}
\label{app:fault_attribution}
% ─────────────────────────────────────────────────────────────────────────────

This appendix provides implementation details and per-model breakdowns
for the fault attribution analysis of \Cref{sec:fault_results}, based
on the extractor vote $P(e' \mid \wpr)$ (\Cref{eq:consensus}) and
extraction agreement $p^{*}$ (\Cref{sec:fault_method}).

\subsection*{Computing extractor votes}

The symbolic equivalence $\equiv$ in \Cref{eq:consensus} is evaluated
via SymPy: each predicted string is parsed into an expression and
simplified; two predictions are equivalent if their simplified forms
match.  Predictions that fail to parse are treated as distinct.
This is applied to all $N$ extractor outputs per word problem to
compute $P(e' \mid \wpr)$ and, for non-decodable word problems,
the extraction agreement $p^{*}$ (\Cref{sec:fault_method}).  Simplification is
parallelized across expressions using \texttt{ProcessPoolExecutor}.

\subsection*{Aggregate results}

\Cref{tab:fault_attr} reports the aggregate breakdown.  Of 392276
errors, 73.6\% arise from non-decodable word problems
($P(\expr \mid \wpr) = 0$), matching the headline rate
reported in \Cref{sec:fault_results}.

\begin{table}[htb]
\centering
\caption{Fault attribution summary for all incorrect round-trips.}
\label{tab:fault_attr}
\small
\begin{tabular}{@{}lrr@{}}
\toprule
Category & Count & \% of errors \\
\midrule
Extractor fault ($P(\expr \mid \wpr) > 0$) & 103668 & 26.4 \\
Non-decodable ($P(\expr \mid \wpr) = 0$)   & 288608 & 73.6 \\
\bottomrule
\end{tabular}
\end{table}

\subsection*{Per-depth breakdown}

\Cref{tab:fault_depth} shows the per-depth breakdown.
At depth~2, 37.9\% of errors are extractor faults (decodable word
problems); as depth increases, the extractor-fault rate falls to
24.7\% at depth~6. Deeper word problems are longer and more
ambiguous, causing extractor outputs to diverge.

\begin{table}[htb]
\centering
\caption{Fault attribution by expression depth.  Ext-fault \%
is the share of errors from decodable word problems.}
\label{tab:fault_depth}
\small
\begin{tabular}{@{}crr@{}}
\toprule
Depth & Errors & Ext-fault \% \\
\midrule
2 &    1622 & 37.9 \\
3 &   13926 & 29.2 \\
4 &  128749 & 26.8 \\
5 &  133516 & 27.1 \\
6 &  114463 & 24.7 \\
\bottomrule
\end{tabular}
\end{table}

\subsection*{Per-generator breakdown}

\Cref{tab:fault_gen} breaks down the two fault-attribution metrics by
generator model, ordered by parameter count.

\begin{table}[htb]
\centering
\caption{Fault attribution by generator model.  Larger generators
produce more decodable word problems (higher ext-fault \%).}
\label{tab:fault_gen}
\small
\begin{tabular}{@{}lrr@{}}
\toprule
Generator & Errors & Ext-fault \% \\
\midrule
Qwen3-0.6B  &  6889 &  0.1 \\
Qwen3-1.7B  & 19630 &  3.8 \\
Qwen3-4B    & 28240 & 14.3 \\
Gemma-3-4B  & 35235 &  1.3 \\
Dream-7B    & 18753 &  5.5 \\
LLaDA-8B    & 23910 & 19.6 \\
Qwen3-8B    & 27813 & 17.4 \\
Gemma-3-12B & 34419 &  5.2 \\
Qwen3-14B   & 28676 & 21.0 \\
Phi-4       & 24817 & 29.5 \\
Gemma-3-27B & 33939 &  6.0 \\
Qwen3-32B   & 29524 & 26.6 \\
Haiku-4.5      & 19624 & 59.6 \\
Gemini-3-Flash & 23456 & 68.6 \\
Gemini-3.1-Pro & 17838 & 90.2 \\
GPT-5          & 19513 & 96.9 \\
\bottomrule
\end{tabular}
\end{table}

\textbf{Small models produce systematic errors.}
Qwen3-0.6B and Gemma-3-4B have the lowest extractor-fault rates
($<$2\%).  Their word problems are wrong in a consistent,
predictable way: nearly all extractors decode the same incorrect
expression, and the word problems are almost never decodable.
In contrast, GPT-5 shows the opposite pattern: 97\% of
its errors are extractor faults, meaning the word problem is
faithful but extractors fail to recover the target expression.

\textbf{Large models shift errors to the extractor.}
The frontier generators (Haiku-4.5, Gemini-3-Flash, Gemini-3.1-Pro,
GPT-5) have extractor-fault rates of 60--97\%: their word problems
are far more often decodable, so when failures occur they are
predominantly attributable to extractor limitations.

\subsection*{Takeaway}

Fault attribution reinforces a central finding of this paper:
generation is harder than extraction.  Over 73\% of errors arise from
non-decodable word problems, with the rate reaching essentially 100\% for the
smallest generator and dropping to 3\% for GPT-5.
This asymmetry is consistent with the
generation-extraction gap observed in \Cref{sec:comm_matrix_results} and the
IRT generation-clarity coefficients in \Cref{app:irt}.  Improving
round-trip fidelity will require advances primarily on the encoding
side, particularly for deep expressions where generators most
frequently distort the intended arithmetic structure.

\subsection{Misattribution Sensitivity}
\label{app:fault_sensitivity}

The consensus-based attribution (\Cref{sec:fault_method}) classifies a
word problem as a generator fault when no extractor in the panel
recovers the target expression. A faithful word problem will be
misclassified whenever every extractor independently fails on it. We
bound this probability as follows.

Let $f_i$ denote the success rate of extractor~$i$ on confirmed-faithful
word problems (those where at least one extractor succeeds). Assuming
conditional independence across extractors, the probability that a
faithful word problem is misclassified as a generator fault is
\begin{equation}
  P(\text{misattribution})
  \;=\;
  \prod_{i=1}^{N}(1 - f_i).
  \label{eq:misattr}
\end{equation}

\Cref{tab:per_ext_f} reports $f_i$ for all 16 extractors, estimated from
14232 confirmed-faithful word problems. Values range from 0.064 (Qwen3-0.6B)
to 0.820 (GPT-5), with a mean of $\bar{f} = 0.55$. Using the
per-extractor values in \Cref{eq:misattr} gives
$P(\text{misattribution}) = 4.5 \times 10^{-7}$, yielding fewer than
1 expected false attribution among the 18038 word problems classified
as generator faults by the consensus rule.

The independence assumption is optimistic: extractors may share failure
modes on the same hard word problems. However, the panel spans six
architectural families and four frontier APIs, limiting correlated
failures. Even under moderate positive correlation, the false-attribution
rate remains negligible given the diversity of the panel.

\begin{table}[htb]
\centering
\caption{Per-extractor success rate $f_i$ on 14232 confirmed-faithful
  word problems (those where at least one extractor recovers the
  target). Models sorted by $f_i$.}
\label{tab:per_ext_f}
\small
\begin{tabular}{@{} l r r @{}}
\toprule
Extractor & Correct / 14232 & $f_i$ \\
\midrule
Qwen3-0.6B        &   913 & 0.064 \\
Qwen3-1.7B        &  3154 & 0.222 \\
Dream-7B          &  3959 & 0.278 \\
Gemma-3-4B        &  4179 & 0.294 \\
LLaDA-8B          &  6349 & 0.446 \\
Gemma-3-12B       &  6890 & 0.484 \\
Qwen3-4B          &  7281 & 0.512 \\
Qwen3-8B          &  7782 & 0.547 \\
Gemma-3-27B       &  9028 & 0.634 \\
Qwen3-14B         &  9393 & 0.660 \\
Qwen3-32B         &  9783 & 0.687 \\
Phi-4             &  9992 & 0.702 \\
Haiku-4.5         & 10708 & 0.752 \\
Gemini-3-Flash    & 11404 & 0.801 \\
Gemini-3.1-Pro    & 11557 & 0.812 \\
GPT-5             & 11672 & 0.820 \\
\midrule
Mean               &       & 0.545 \\
\bottomrule
\end{tabular}
\end{table}

\FloatBarrier
\subsection*{Panel-composition sensitivity}
\label{app:panel_sensitivity}
The consensus-based attribution depends on the extractor panel: a
generation is classified as faithful whenever any single extractor
recovers the target, so the generator-fault rate is a function of panel
composition. \Cref{tab:fault_panel} reports the rate under nine panel
subsets, all reusing the same extraction data. Across these subsets the
rate ranges 73.6--91.1\%, with the full 16-model panel yielding the
lowest value: every restricted panel classifies more failures as
generator faults, because smaller panels afford fewer chances for
``at least one'' extractor to succeed. Within same-size panels
($|P|{=}3$), the Gemma-3-only panel gives 85.8\% and the
other-open-weight panel gives 82.6\%, indicating that stronger extractors are
if anything more likely to collectively fail on genuinely bad word
problems. The diversity-controlled ``one per family'' panel (77.3\%)
is close to Qwen3-only (79.3\%) and above the full open-weight panel
(74.4\%), offering no evidence that correlated failures within a single
family inflate the headline figure.

\begin{table}[htb]
\centering
\caption{Generator-fault rate under different extractor panel subsets
  (incl-guard convention). The full-panel rate of 73.6\% is the
  lowest observation; every restricted panel yields a higher rate.}
\label{tab:fault_panel}
\small
\begin{tabular}{@{} l r r r @{}}
\toprule
Panel                          & $|P|$ & Errors  & GenF \% \\
\midrule
Full (all models)              & 16 & 392276 & 73.6 \\
Open-weight only               & 12 & 308537 & 74.4 \\
Qwen3 only                     &  6 & 155314 & 79.3 \\
Gemma-3 only                   &  3 &  76713 & 85.8 \\
Other (Phi/Dream/LLaDA)        &  3 &  76510 & 82.6 \\
Frontier only                  &  4 &  83739 & 91.1 \\
Strong ($\geq$14B + frontier)  &  8 & 174623 & 84.8 \\
Weak ($\leq$4B)                &  4 & 113553 & 83.7 \\
One per family (median)        &  6 & 146976 & 77.3 \\
\bottomrule
\end{tabular}
\end{table}

% ─────────────────────────────────────────────────────────────────────────────
\section{Latent-Ability Model (IRT)}
\label{app:irt}
% ─────────────────────────────────────────────────────────────────────────────

As a validation of the additive decomposition in \Cref{sec:matrix}, we
fit a Rasch model~\citep{Rasch1980ProbabilisticMF} to the binary
per-example outcomes in the communication matrix.  The fit uses all 16
models (12 open-weight and 4 frontier).  For each triple (extractor~$i$,
generator~$j$, expression~$k$), we model the probability of a correct
round-trip as
\begin{equation}
  \log \frac{P(\text{correct})}{1 - P(\text{correct})}
  \;=\;
  \mu + \theta_i + \beta_j - \delta_k,
  \label{eq:rasch}
\end{equation}
where $\theta_i$ is the extraction skill of model~$i$, $\beta_j$ is
the generation clarity of model~$j$, and $\delta_k$ is the difficulty
of expression~$k$.  Equivalently, the predicted success probability is
\begin{equation}
  P(\text{correct})
  \;=\;
  \frac{1}{1 + \exp\!\bigl[-(\mu + \theta_i + \beta_j - \delta_k)\bigr]}.
  \label{eq:rasch_prob}
\end{equation}
For example, given the estimated global baseline $\mu = -1.90$, pairing Phi-4 ($\theta = +0.53$, $\beta = +1.07$) on a
median-difficulty expression ($\delta = 0.1$) gives
$P = \sigma(-1.90 + 0.53 + 1.07 - 0.1) \approx 40.1\%$, while
a hard expression ($\delta = 1.2$, $\sim$90th percentile) yields
$P = \sigma(-1.90 + 0.53 + 1.07 - 1.2) \approx 18.2\%$.
Parameters are estimated by $L_2$-regularised logistic regression
(regularisation strength $C = 10$).

\Cref{tab:irt_coefficients} reports the estimated coefficients for each
model alongside its raw marginal accuracy from the communication
matrix.  Spearman rank correlations between the IRT coefficients and the
matrix marginals in \Cref{tab:matrix} are $\rho = 1.00$ for extraction
and $\rho = 0.96$ for generation, confirming that the simple averages
faithfully summarise latent ability.  Among the three parameter groups,
generation clarity has the largest variance
($\sigma_\beta = 1.93$, compared with
$\sigma_\delta = 1.00$ for expression difficulty and
$\sigma_\theta = 0.98$ for extraction), indicating that variation in
how well models encode structure exceeds variation in expression
difficulty or extraction skill.  Including all three parameter
groups reduces the log-loss from 0.44 (model-only, $\theta + \beta$) to
0.38 (full model with $\delta$), confirming that expression
difficulty adds predictive signal beyond the model coefficients.

\begin{table}[htb]
\centering
\caption{Rasch model coefficients and matrix marginal accuracies.
  $\theta$: extraction skill; $\beta$: generation clarity;
  Raw Ext.\ / Gen.: row / column averages of the $16 \times 16$ communication
  matrix in \Cref{tab:matrix} (guard failures counted as errors).
  Models are sorted by family and size.}
\label{tab:irt_coefficients}
\small
\begin{tabular}{l rr rr}
\toprule
Model & $\theta$ & Raw Ext.\ (\%) & $\beta$ & Raw Gen.\ (\%) \\
\midrule
Qwen3-0.6B & -2.89 & 2.3 & -4.23 & 0.1 \\
Qwen3-1.7B & -1.43 & 8.0 & -1.61 & 2.6 \\
Qwen3-4B & -0.08 & 18.6 & +0.02 & 13.4 \\
Qwen3-8B & +0.11 & 19.9 & +0.53 & 19.0 \\
Qwen3-14B & +0.42 & 24.0 & +0.55 & 20.5 \\
Qwen3-32B & +0.49 & 25.0 & +0.44 & 18.8 \\
Gemma-3-4B & -0.92 & 10.7 & -3.20 & 0.7 \\
Gemma-3-12B & -0.14 & 17.6 & -1.42 & 3.3 \\
Gemma-3-27B & +0.36 & 23.0 & -1.74 & 3.3 \\
Phi-4 & +0.53 & 25.5 & +1.07 & 25.8 \\
Dream-7B & -1.07 & 10.1 & -1.60 & 2.2 \\
LLaDA-8B & -0.34 & 16.2 & +0.66 & 15.9 \\
Haiku-4.5 & +0.67 & 27.3 & +2.10 & 46.5 \\
GPT-5 & +0.81 & 29.8 & +2.22 & 50.2 \\
Gemini-3-Flash & +0.77 & 29.1 & +1.74 & 39.6 \\
Gemini-3.1-Pro & +0.82 & 29.5 & +2.59 & 54.4 \\
\midrule
Spearman $\rho$ & \multicolumn{2}{c}{1.00} & \multicolumn{2}{c}{0.96} \\
\bottomrule
\end{tabular}
\end{table}

% \begin{figure}[htb]
%   \centering
%   \includegraphics[width=0.75\linewidth]{figures/irt_broadcaster_listener.pdf}
%   \caption{IRT-coefficient analogue of \Cref{fig:broadcaster}a.  Each
%   model's generation clarity~$\beta$ (horizontal) is plotted against its
%   extraction skill~$\theta$ (vertical), both estimated by the Rasch model
%   in \Cref{eq:rasch}.  The generation-extraction asymmetry observed in the
%   raw marginals is preserved almost exactly (Spearman $\rho = 1.00$ for
%   extraction, $\rho = 0.99$ for generation), confirming that the simple
%   row and column averages faithfully summarise latent ability.}
%   \label{fig:irt_scatter}
% \end{figure}

\section{Fine-Tuning: Assembly Domain Details}
\label{app:finetuning_assembly}

This appendix provides the full domain specification for both the
nested assembly fine-tuning data, along with
training examples and qualitative analysis supporting the fine-tuning
experiment in \Cref{sec:finetuning}.

\subsection{Assembly Domain Specification}
\label{app:assembly_generation}

Expressions are random trees over eleven assembly operators of mixed
arity: three unary (\texttt{label}, \texttt{inspect},
\texttt{polish}), six binary (three commutative: \texttt{weld},
\texttt{glue}, \texttt{rivet}; three non-commutative: \texttt{mount},
\texttt{pour}, \texttt{load}), and four ternary (two commutative:
\texttt{mix}, \texttt{assort}; two non-commutative: \texttt{layer},
\texttt{thread}). Operators are applied to coloured-object leaves
drawn from 13 colours $\times$ 13 objects. Tree depth is sampled
uniformly from 2 to 5, matching the benchmark range. Each operator
maps to multiple paraphrase templates describing physical assembly
steps in a \emph{procedural} style (ordinal step markers). Every
(expression, NL) pair
yields two training examples ($\expr \!\to\! \wpr$ and
$\wpr \!\to\! \expr$). We generate 500 pairs per depth with a 90/10
train/test split, producing approximately 3600 training messages.

% \subsection{Flat Control Specification}
% \label{app:flat_control}

% The flat control dataset uses nine of the eleven operators (excluding
% the ternary non-commutative \texttt{layer} and \texttt{thread}, whose
% NL templates inherently describe sequential ordering that would leak
% compositional structure), the same coloured-object
% vocabulary, NL styles, and bidirectional task format as the nested
% assembly data. The key difference is that no expression contains
% recursive composition: at each depth level $k$, each training example
% contains $k$ independent single-operator expressions rather than a
% depth-$k$ tree. The $k$ descriptions are joined by discourse
% separators (``Separately,'' ``Also,'' ``Meanwhile,'' etc.)\ to form a
% multi-sentence passage. This matches the nested data in total operator
% count, entity count, and token budget per depth level, isolating
% recursive nesting as the only variable.

\subsection{Training Examples}
\label{app:assembly_examples}

\Cref{tab:assembly_examples} shows three nested assembly expressions
rendered in procedural style, illustrating the
mixed arity (unary \texttt{polish}, binary \texttt{mount}, ternary
\texttt{assort}/\texttt{load}/\texttt{mix}) and non-commutative
argument ordering. 
% \Cref{tab:flat_examples} shows the corresponding
% flat control examples at $k = 2, 3, 5$ independent operations.

\begin{table}[htb]
\centering
\caption{\textbf{Sample assembly expressions} (depth 3) with
procedural NL renderings.}
\vspace{2mm}
\label{tab:assembly_examples}
\small
\setlength{\tabcolsep}{5pt}
\begin{tabular}{@{} p{3.8cm} p{10.2cm} @{}}
\toprule
\textbf{Expression} & \textbf{Natural-language rendering} \\
\midrule
\texttt{assort(label(copper gem), gold ring, yellow card)}
  & First, mark the copper gem with a tag to get the first component.\ Second, assort the first component, the gold ring, and the yellow card into one set as the final product. \\
\midrule
\texttt{mount(rivet(bronze ball, copper coin), black cube)}
  & First, clinch the bronze ball to the copper coin with a pop rivet to get the first component.\ Second, position the first component on the face of the black cube as the final product. \\
\midrule
\texttt{polish(layer(silver ring, gold rod, blue cube))}
  & First, deposit the silver ring at the bottom, the gold rod in the centre, and the blue cube at the top to get the first component.\ Second, buff the first component to a shine as the final product. \\
\bottomrule
\end{tabular}
\end{table}

% \begin{table}[htb]
% \centering
% \caption{\textbf{Sample flat control examples} at $k = 2, 3, 5$ independent
% operations. Same vocabulary and operators as \Cref{tab:assembly_examples}
% but no recursive nesting; operations are joined by discourse separators.}
% \vspace{2mm}
% \label{tab:flat_examples}
% \small
% \setlength{\tabcolsep}{5pt}
% \begin{tabular}{@{} c p{5cm} p{8cm} @{}}
% \toprule
% $k$ & \textbf{Expressions} & \textbf{Natural-language rendering} \\
% \midrule
% 2 &
% \texttt{pour(green rod, gold star); rivet(blue cube, purple rod)}
% &
% They decanted the green rod into the gold star.\ Also, the blue cube
% and the purple rod were bolted with a rivet gun. \\[6pt]
% \midrule
% 3 &
% \texttt{assort(orange card, yellow cube, green gem); label(orange stone); polish(bronze rod)}
% &
% The orange card, the yellow cube, and the green gem assorted together.\
% Meanwhile, a worker labelled the orange stone.\ Also, polish the bronze
% rod. \\[6pt]
% \midrule
% 5 &
% \texttt{mix(purple bead, yellow bead, red star); assort(copper ring, silver rod, green knot); mount(pink bead, copper gem); assort(pink card, red ball, orange card); glue(purple cube, blue ball)}
% &
% Blend the purple bead, the yellow bead, and the red star in a bowl.\
% Independently, the copper ring, the silver rod, and the green knot were
% assorted into one set.\ Meanwhile, mount the pink bead onto the copper
% gem.\ Meanwhile, arrange the pink card, the red ball, and the orange
% card into a set.\ In addition, they glued the purple cube and the blue
% ball together. \\
% \bottomrule
% \end{tabular}
% \end{table}

\subsection{Procedural assembly FT: failure taxonomy}
The $2.1$~pp guard-pass drop reported for assembly FT
(\Cref{tab:ft_assembly_summary}) is concentrated in two models, not
shared across the panel. \Cref{tab:guard_taxonomy_ft_proced} gives
the per-model $\Guard$ pass rate (operator adjacency, the only
category counted in the guard column) for the base and assembly
FT conditions. Gemma-3-4B and Gemma-3-12B account for nearly all
of the aggregate drop ($-22.9$ and $-17.5$~pp). Phi-4 and
Qwen3-14B/32B stay within $\pm 0.7$~pp of their base, and
Qwen3-0.6B/1.7B improve ($+32.6$ and $+7.0$~pp) because their
base outputs leak heavily and FT reduces that tendency. We do not
have a mechanistic account for the Gemma-specific drop. The \emph{missing-variable}
fault (a generator error under our reporting rule, not a guard
failure) rises by $+7.0$~pp on average under FT, dominated again
by Qwen3-0.6B ($+42.2$) and Qwen3-1.7B ($+17.8$), where the
smallest generators drop placeholders more often after training.

\begin{table}[htb]
\centering
\small
\setlength{\tabcolsep}{5pt}
\caption{\textbf{$\Guard$ pass rate under base and procedural
assembly FT} (operator adjacency, the only failure category
counted in the $\Guard$ column of \Cref{tab:ft_assembly_summary}).
Values are the share of 1790 problems that clear the operator-leak
check. Two Gemma models concentrate the aggregate drop.}
\label{tab:guard_taxonomy_ft_proced}
\vspace{1mm}
\begin{tabular}{l rrr}
\toprule
Model & Base \% & FT$_{\text{asm}}$ \% & $\Delta$ pp \\
\midrule
Qwen3-0.6B  & 19.6 & 52.2 & $+32.6$ \\
Qwen3-1.7B  & 58.9 & 65.9 & $+7.0$  \\
Qwen3-4B    & 93.9 & 82.2 & $-11.7$ \\
Qwen3-8B    & 97.5 & 94.0 & $-3.5$  \\
Qwen3-14B   & 99.2 & 98.5 & $-0.7$  \\
Qwen3-32B   & 98.4 & 97.7 & $-0.7$  \\
Gemma-3-4B  & 98.0 & 75.1 & $\mathbf{-22.9}$ \\
Gemma-3-12B & 96.2 & 78.7 & $\mathbf{-17.5}$ \\
Gemma-3-27B & 96.1 & 92.7 & $-3.4$  \\
Phi-4       & 96.6 & 96.5 & $-0.1$  \\
\midrule
Panel mean  & 85.4 & 83.3 & $-2.1$  \\
\bottomrule
\end{tabular}
\end{table}

\subsection{Qualitative Analysis}
\label{app:ft_qualitative}

\Cref{tab:ft_assembly_qualitative} illustrates how assembly 
% and flat-control
fine-tuning change the style of generated word problems for Phi-4.

\begin{table}[htb]
\centering
\caption{Word problems generated by Phi-4 (base
% , flat-control
  % $\text{FT}_{\text{flat}}$, 
  and assembly fine-tuned FT) for three
  expressions. Assembly fine-tuning shifts output toward explicit
  arithmetic descriptions.
  % ; 
  % flat-control fine-tuning produces word
  % problems closer to the base model, sometimes leaking mathematical
  % notation.
  }
\vspace{2mm}
\label{tab:ft_assembly_qualitative}
\small
\setlength{\tabcolsep}{5pt}
\begin{tabular}{@{} p{2.0cm} p{12cm} @{}}
\toprule
\multicolumn{2}{@{}l}{\textbf{Expression:} \texttt{( C - ( A + B + 2 ) )} \quad [depth 3]} \\
\midrule
Base & A real estate agent is calculating the remaining budget for a property renovation.\ The total budget is \{C\}, and the costs include \{A\} for materials, \{B\} for labor, and an additional 2 for miscellaneous expenses.\ The agent needs to determine how much budget is left after these expenses. \\[4pt]
% $\text{FT}_{\text{flat}}$ & A real estate agent had \{C\} properties listed.\ She sold \{A\} properties last month, \{B\} properties this month, and 2 properties next month.\ How many properties does she still have listed? \\[4pt]
FT   & A developer had \{C\} plots of land.\ They sold \{A\} plots in the spring, \{B\} plots in the summer, and set aside 2 for a park.\ How many plots remained? \\
\midrule
\multicolumn{2}{@{}l}{\textbf{Expression:} \texttt{( B - C * ( A + G ) )} \quad [depth 3]} \\
\midrule
Base & A delivery truck starts with \{B\} gallons of fuel.\ It uses \{C\} gallons for each mile driven, and it travels a total of \{A\} miles plus an additional \{G\} miles for detours.\ How many gallons of fuel does the truck have left after completing the trip? \\[4pt]
% $\text{FT}_{\text{flat}}$ & A team has \{C\} players available.\ They need to form a sub-team by excluding \{B\} times \{A\} divided by \{G\} players.\ How many players are on the sub-team? \\[4pt]
FT   & A bus company had \{B\} buses in service.\ They assigned \{A\} buses to the north route and \{G\} buses to the south route, then multiplied the total by \{C\} to get the number of buses used for express service.\ How many buses remained in service? \\
\midrule
\multicolumn{2}{@{}l}{\textbf{Expression:} \texttt{( G * ( D + C ) / H )} \quad [depth 3]} \\
\midrule
Base & A musician is composing a piece and wants to determine the total number of beats in a section.\ They have \{G\} measures, each containing \{D\} beats plus an additional \{C\} beats for a special rhythm.\ To find the average number of beats per measure, they divide the total beats by \{H\}. \\[4pt]
% $\text{FT}_{\text{flat}}$ & A music producer distributed \{G\} new songs across \{D\} standard and \{C\} premium streaming platforms.\ This gave an average of \{H\} songs per platform.\ How many songs were on each platform? \\[4pt]
FT   & First, add \{D\} and \{C\} together to get the first component.\ Second, multiply \{G\} by the first component to get the second component.\ Third, divide the second component by \{H\} to get the third component.\ Fourth, the third component is the final product. \\
\bottomrule
\end{tabular}
\end{table}

The base model produces fluent narratives whose structure often diverges
from the target expression: in the first example, \texttt{C - (A + B + 2)}
becomes ``remaining budget'' after listed expenses, flattening the
subtraction-of-a-sum into a single subtractive relation.
The assembly-fine-tuned model makes operations explicit and sequential:
narrative outputs preserve operator order (``sold \ldots\ in the spring,
\{B\} plots in the summer, and set aside 2 for a park''), and the
third example collapses to a pure step-marker form (``First, add \ldots\
Second, multiply \ldots\ Third, divide \ldots''). This structural
fidelity comes at a stylistic cost: two Gemma models contribute most of
the guard-rate drop reported in \Cref{tab:guard_taxonomy_ft_proced}.
% The flat-control model produces word problems stylistically similar to
% the base model; in the second example it resorts to ``\{B\} times
% \{A\} divided by \{G\},'' leaking mathematical phrasing that
% approaches the guard threshold.

% ─────────────────────────────────────────────────────────────────────────────
\section{Shot-Configuration Grid}
\label{app:shot_grid}

\Cref{tab:shot_grid_base,tab:shot_grid_asm,tab:shot_grid_arith} report
generation quality ($\genqual{}$) and extraction quality ($\extqual{}$)
across all evaluated $(\gamma, \varepsilon)$ shot configurations for
each training condition. The base figure justifies the
$(\gamma{=}0, \varepsilon{=}3)$ configuration used in
\Cref{tab:matrix}; the FT figures justify the per-condition selections
used in \Cref{tab:ft_assembly_summary}. Bold marks the per-row maximum
in each panel. All values (\%) are averaged over the 10-model
open-weight extractor panel with guard failures counted as errors.

\begin{figure}[htb]
\centering
\begin{minipage}[t]{0.48\linewidth}
\centering
{\small\textbf{(a) Generation quality ($\genqual{}$)}}
\vspace{1mm}

\small
\setlength{\tabcolsep}{4pt}
\begin{tabular}{l cc cc}
\toprule
 & \multicolumn{2}{c}{$\gamma=0$} & \multicolumn{2}{c}{$\gamma=3$} \\
\cmidrule(lr){2-3}\cmidrule(lr){4-5}
Model & $\varepsilon{=}0$ & $\varepsilon{=}3$ & $\varepsilon{=}0$ & $\varepsilon{=}3$ \\
\midrule
Qwen3-0.6B  &  0.1 & 0.1 & \textbf{0.3} & \textbf{0.3} \\
Qwen3-1.7B  &  \textbf{2.6} & \textbf{2.6} & 0.3 & 0.3 \\
Qwen3-4B    & 13.7 & \textbf{14.0} & 2.9 & 3.1 \\
Qwen3-8B    & 19.7 & \textbf{20.5} & 5.3 & 5.6 \\
Qwen3-14B   & 20.7 & \textbf{21.4} & 10.6 & 11.3 \\
Qwen3-32B   & 18.3 & \textbf{19.4} & 10.0 & 10.7 \\
Gemma-3-4B  &  0.8 & 0.8 & 1.0 & \textbf{1.1} \\
Gemma-3-12B &  3.8 & \textbf{3.9} & 2.6 & 2.5 \\
Gemma-3-27B &  3.6 & 3.8 & \textbf{4.1} & \textbf{4.1} \\
Phi-4       & 25.9 & \textbf{26.9} & 10.4 & 10.9 \\
\bottomrule
\end{tabular}
\end{minipage}%
\hfill
\begin{minipage}[t]{0.48\linewidth}
\centering
{\small\textbf{(b) Extraction quality ($\extqual{}$)}}
\vspace{1mm}

\small
\setlength{\tabcolsep}{4pt}
\begin{tabular}{l cc cc}
\toprule
 & \multicolumn{2}{c}{$\gamma=0$} & \multicolumn{2}{c}{$\gamma=3$} \\
\cmidrule(lr){2-3}\cmidrule(lr){4-5}
Model & $\varepsilon{=}0$ & $\varepsilon{=}3$ & $\varepsilon{=}0$ & $\varepsilon{=}3$ \\
\midrule
Qwen3-0.6B  &  2.6 & \textbf{3.0} & 0.7 & 0.7 \\
Qwen3-1.7B  &  3.1 & \textbf{6.6} & 1.2 & 2.1 \\
Qwen3-4B    & \textbf{12.5} & \textbf{12.4} & 5.4 & 5.4 \\
Qwen3-8B    & 12.8 & \textbf{12.9} & 5.5 & 6.0 \\
Qwen3-14B   & 14.3 & \textbf{14.5} & 6.5 & 7.1 \\
Qwen3-32B   & 14.0 & \textbf{14.7} & 6.7 & 7.0 \\
Gemma-3-4B  & \textbf{8.6} & 8.4 & 3.1 & 3.3 \\
Gemma-3-12B & \textbf{11.9} & 11.7 & 5.2 & 5.3 \\
Gemma-3-27B & 14.0 & \textbf{14.2} & 6.4 & 6.5 \\
Phi-4       & \textbf{15.3} & 15.1 & 6.7 & 6.7 \\
\bottomrule
\end{tabular}
\end{minipage}
\caption{\textbf{Base condition: shot-configuration grid.}}
\label{tab:shot_grid_base}
\end{figure}

\begin{figure}[htb]
\centering
\begin{minipage}[t]{0.48\linewidth}
\centering
{\small\textbf{(a) Generation quality ($\genqual{}$)}}
\vspace{1mm}

\small
\setlength{\tabcolsep}{4pt}
\begin{tabular}{l cc cc}
\toprule
 & \multicolumn{2}{c}{$\gamma=0$} & \multicolumn{2}{c}{$\gamma=3$} \\
\cmidrule(lr){2-3}\cmidrule(lr){4-5}
Model & $\varepsilon{=}0$ & $\varepsilon{=}3$ & $\varepsilon{=}0$ & $\varepsilon{=}3$ \\
\midrule
Qwen3-0.6B  &  0.0 &  0.0 & \textbf{0.3} &  0.1 \\
Qwen3-1.7B  &  0.2 &  0.1 &  2.6 & \textbf{6.5} \\
Qwen3-4B    & \textbf{20.8} & 11.2 & 12.1 & 17.3 \\
Qwen3-8B    & 22.3 & 22.8 & 23.3 & \textbf{32.6} \\
Qwen3-14B   & 20.3 & 24.8 & \textbf{27.6} & 26.2 \\
Qwen3-32B   & 23.7 & 21.3 & 20.5 & \textbf{27.6} \\
Gemma-3-4B  &  6.0 & \textbf{7.4} &  3.4 &  5.5 \\
Gemma-3-12B &  7.9 & 11.5 & 12.0 & \textbf{16.7} \\
Gemma-3-27B & 23.2 & 23.1 & \textbf{25.8} & 20.4 \\
Phi-4       & \textbf{36.9} & 35.2 & 34.1 & 32.7 \\
\bottomrule
\end{tabular}
\end{minipage}%
\hfill
\begin{minipage}[t]{0.48\linewidth}
\centering
{\small\textbf{(b) Extraction quality ($\extqual{}$)}}
\vspace{1mm}

\small
\setlength{\tabcolsep}{4pt}
\begin{tabular}{l cc cc}
\toprule
 & \multicolumn{2}{c}{$\gamma=0$} & \multicolumn{2}{c}{$\gamma=3$} \\
\cmidrule(lr){2-3}\cmidrule(lr){4-5}
Model & $\varepsilon{=}0$ & $\varepsilon{=}3$ & $\varepsilon{=}0$ & $\varepsilon{=}3$ \\
\midrule
Qwen3-0.6B  &  3.3 & \textbf{3.9} &  2.5 &  3.7 \\
Qwen3-1.7B  &  4.1 &  7.3 &  3.4 & \textbf{8.4} \\
Qwen3-4B    & 18.2 & 17.7 & 18.3 & \textbf{20.9} \\
Qwen3-8B    & 19.6 & 18.3 & 17.6 & \textbf{21.1} \\
Qwen3-14B   & 21.9 & 20.7 & 21.8 & \textbf{24.2} \\
Qwen3-32B   & 22.8 & 22.6 & 22.5 & \textbf{25.3} \\
Gemma-3-4B  &  9.1 &  9.5 & 10.0 & \textbf{11.9} \\
Gemma-3-12B & 17.3 & 14.5 & 17.3 & \textbf{19.2} \\
Gemma-3-27B & 20.0 & 20.5 & 22.8 & \textbf{24.2} \\
Phi-4       & 24.9 & 22.6 & 25.6 & \textbf{26.6} \\
\bottomrule
\end{tabular}
\end{minipage}
\caption{\textbf{Assembly FT (procedural) condition: shot-configuration grid.} Panel means are $16.13 / 15.75 / 16.17 / 18.54$ for $(\gamma,\varepsilon) \in \{(0,0),(0,3),(3,0),(3,3)\}$, so the panel optimum is $(\gamma{=}3,\varepsilon{=}3)$. Per-row optima vary: Phi-4 and Qwen3-4B peak at $(0,0)$, Qwen3-14B and Gemma-3-27B at $(3,0)$.}
\label{tab:shot_grid_asm}
\end{figure}

\begin{figure}[htb]
\centering
\begin{minipage}[t]{0.48\linewidth}
\centering
{\small\textbf{(a) Generation quality ($\genqual{}$)}}
\vspace{1mm}

\small
\setlength{\tabcolsep}{6pt}
\begin{tabular}{l cc}
\toprule
 & \multicolumn{2}{c}{$\gamma=0$} \\
\cmidrule(lr){2-3}
Model & $\varepsilon{=}0$ & $\varepsilon{=}3$ \\
\midrule
Qwen3-0.6B  & 47.7 & \textbf{51.9} \\
Qwen3-1.7B  & 57.0 & \textbf{57.2} \\
Qwen3-4B    & 53.2 & \textbf{55.3} \\
Qwen3-8B    & 56.4 & \textbf{59.5} \\
Qwen3-14B   & 54.8 & \textbf{55.9} \\
Qwen3-32B   & 56.0 & \textbf{58.0} \\
Gemma-3-4B  & 53.9 & \textbf{55.4} \\
Gemma-3-12B & 56.5 & \textbf{57.8} \\
Gemma-3-27B & 55.0 & \textbf{58.4} \\
Phi-4       & 56.6 & \textbf{58.2} \\
\bottomrule
\end{tabular}
\end{minipage}%
\hfill
\begin{minipage}[t]{0.48\linewidth}
\centering
{\small\textbf{(b) Extraction quality ($\extqual{}$)}}
\vspace{1mm}

\small
\setlength{\tabcolsep}{6pt}
\begin{tabular}{l cc}
\toprule
 & \multicolumn{2}{c}{$\gamma=0$} \\
\cmidrule(lr){2-3}
Model & $\varepsilon{=}0$ & $\varepsilon{=}3$ \\
\midrule
Qwen3-0.6B  & 13.7 & \textbf{14.1} \\
Qwen3-1.7B  & 17.2 & \textbf{34.8} \\
Qwen3-4B    & 62.8 & \textbf{64.4} \\
Qwen3-8B    & 64.7 & \textbf{66.2} \\
Qwen3-14B   & 71.0 & \textbf{72.7} \\
Qwen3-32B   & 70.0 & \textbf{73.6} \\
Gemma-3-4B  & \textbf{39.2} & 37.1 \\
Gemma-3-12B & \textbf{62.2} & 59.8 \\
Gemma-3-27B & \textbf{72.2} & 70.6 \\
Phi-4       & 74.0 & \textbf{74.4} \\
\bottomrule
\end{tabular}
\end{minipage}
\caption{\textbf{Arithmetic FT condition: shot-configuration grid.}
  Arithmetic FT was evaluated with $\gamma{=}0$ only (in-domain generators
  do not benefit from exemplars).}
\label{tab:shot_grid_arith}
\end{figure}

% ─────────────────────────────────────────────────────────────────────────────
\section{Lossless-Channel Extraction Baselines}
\label{app:sexpr}
% ─────────────────────────────────────────────────────────────────────────────

To quantify the cost of using natural language as the communication
channel, we replace the NL word problem with an unambiguous structured
representation and measure how accurately the extractor recovers the
original infix expression.  We evaluate two formats:

\textbf{S-expression (prefix notation).}
Each expression is deterministically converted to prefix notation,
\eg $(A + B) \times C$ becomes \texttt{(* (+ A B) C)}.

\textbf{JSON abstract syntax tree.}
Each expression is converted to a nested JSON dictionary,
\eg $(A + B) \times C$ becomes
\texttt{\{"op": "*", "left": \{"op": "+", "left": "A", "right": "B"\}, "right": "C"\}}.

Both formats encode tree structure without ambiguity, so extraction
accuracy on either task upper-bounds what any faithful NL encoding
could achieve for a given extractor.

The S-expression vs.\ NL comparison scales with operator count for the
Qwen3 family: for the strongest extractors (14B--32B), S-expression
accuracy stays high even at $k{=}8$ (78--82\%), whereas NL extraction
falls off sharply as expressions grow more complex. The NL tax is thus
negligible at low complexity and widens at high complexity, confirming
that the natural-language bottleneck specifically targets structurally
complex expressions; \Cref{fig:broadcaster}-right quantifies this gap
against the best available lossless format across extractor sizes.

\textbf{Format dependence at mid-scale.}
Comparing the two lossless channels across model sizes reveals that the
structured-format ceiling is not format-invariant. JSON extraction
substantially exceeds S-expression extraction for mid-range models
(Qwen3-4B: 63.2\% vs.\ 40.8\%; Qwen3-8B: 75.9\% vs.\ 37.5\%), while
both converge at 32B (${\approx}\,88\%$). This gap likely reflects
pretraining exposure: models encounter far more JSON than
S-expressions in their training corpora. The NL tax should therefore be
measured against the best available lossless format
($\max(\text{S-expr}, \text{JSON})$), as reported in
\Cref{fig:broadcaster}-right.

% ─────────────────────────────────────────────────────────────────────────────
\section{Propositional-Logic Replication}
\label{app:logic}
% ─────────────────────────────────────────────────────────────────────────────

To test whether the communication bottleneck is specific to arithmetic
or reflects a more general structural limitation, we replicate the
protocol on propositional logic on a reduced $14 \times 14$ roster
(the arithmetic frontier additions Gemini-3-Flash and Gemini-3.1-Pro
are omitted to limit compute). Formulas are built from
five connectives (AND, OR, NOT, IMPLIES, XOR) over propositional
variables $P$--$W$, at depths 2--5. Correctness is judged by
\emph{logical equivalence} via SymPy, so surface-form variation
(e.g.\ commutativity of AND) does not penalise the extractor.
A domain-adapted leakage guard forbids logical operator words and
symbols in the generated scenario, mirroring the arithmetic guard.

\textbf{Protocol.}
The generator receives a formula and a randomly sampled real-world
domain (access control, safety regulations, game rules, etc.) and must
produce a 2--3 sentence scenario using proposition placeholders
$\{P\}, \{Q\}, \ldots$ without any logical vocabulary.
The extractor receives the scenario and must recover the original
formula using only the connectives AND, OR, NOT, IMPLIES, XOR.
The same 14 models serve as both generators and extractors.

\textbf{Example (depth 3).}
\begin{quote}
\small
\begin{verbatim}
Formula: ( ( P AND Q ) IMPLIES R )
Domain: safety regulations

Scenario: When equipment {P} is running and the temperature
reading {Q} exceeds the threshold, the automatic shutdown
protocol {R} activates.
\end{verbatim}
\end{quote}

\textbf{Results.}
\Cref{tab:logic_matrix} shows the full logic communication matrix.
Overall accuracy is lower than in arithmetic (mean cell accuracy
$8.7\%$ vs.\ $14.9\%$ on the shared 14-model block), consistent with the additional ambiguity
introduced by logical connectives that lack the concrete semantics of
arithmetic operators. Despite this difficulty shift, the relative
ordering of models is largely preserved.

\begin{figure}[htb]
  \centering
  \includegraphics[width=\textwidth]{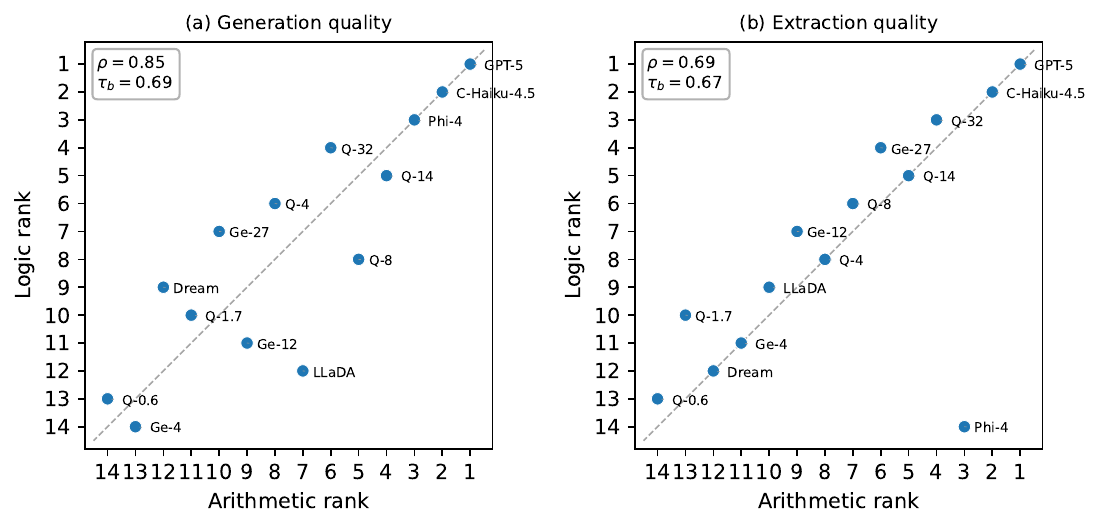}
  \caption{Arithmetic vs.\ logic rank correlation for generation quality
    (left) and extraction quality (right). Each point is one model;
    the dashed diagonal marks perfect agreement. Spearman $\rho$ and
    Kendall $\tau_b$ are shown in each panel. Phi-4 (bottom-right in
    panel b) is the sole large outlier, dropping from extraction rank 3
    to rank 14.}
  \label{fig:rank_correlation}
\end{figure}

\Cref{fig:rank_correlation} compares arithmetic and logic ranks for
both generation and extraction quality. Generation ranks are well
preserved (Spearman $\rho = 0.85$, Kendall $\tau_b = 0.69$): GPT-5
and Claude Haiku 4.5 lead in both domains. Extraction ranks show
comparable pairwise agreement ($\tau_b = 0.67$), though the Spearman
correlation is lower ($\rho = 0.69$) due to a single large
displacement: Phi-4 drops from extraction rank 3 in arithmetic to
rank 14 in logic, showing a strong imbalance in terms of extraction abilities (ranked 3rd on arithmetic). This suggests that its strong arithmetic extraction reflects
format-specific parsing fluency rather than general structural
competence.

Excluding Phi-4, extraction $\rho$ rises above generation $\rho$,
indicating that for all other models the extraction ranking transfers
at least as reliably as the generation ranking.

\begin{table}[htb]
\centering
\caption{\textbf{The $\boldsymbol{14 \times 14}$ communication matrix on all generated expressions (guard failures counted as errors).} Each cell shows round-trip accuracy (\%).  Rows are extractors, columns are generators.  The rightmost column and bottom row show extraction quality (row average) and generation quality (column average). Generator and extractor are run with prompts $\promptg{0}$ and $\prompte{\varepsilon}$ respectively (configuration $\gamma{=}0, \varepsilon{=}3)$; see \Cref{app:shot_grid} for the full ablation. The $\pm$ values denote standard deviation across the 14 pairings in each row or column. The $\Guard$~\% row reports each model's leakage-guard pass rate as a generator. Thin rules separate open-weight auto-regressive, diffusion, and frontier models.}
\vspace{2mm}
\label{tab:logic_matrix}
\small
\setlength{\tabcolsep}{2pt}
\setlength{\dashlinedash}{1pt}
\setlength{\dashlinegap}{1pt}
\resizebox{\textwidth}{!}{%
\begin{tabular}{c l c c c c c c c c c c !{\;\vrule width 0.3pt\;} c c !{\;\vrule width 0.3pt\;} c c !{\;\vrule width 0.75pt\;} c}
\toprule
\multicolumn{17}{c}{\textsc{Generators}} \\
 \multirow{18}{*}{\rotatebox{90}{\textsc{Extractors}}} & & \rotatebox{90}{\scriptsize Qwen3-0.6B} & \rotatebox{90}{\scriptsize Qwen3-1.7B} & \rotatebox{90}{\scriptsize Qwen3-4B} & \rotatebox{90}{\scriptsize Qwen3-8B} & \rotatebox{90}{\scriptsize Qwen3-14B} & \rotatebox{90}{\scriptsize Qwen3-32B} & \rotatebox{90}{\scriptsize Gemma-3-4B} & \rotatebox{90}{\scriptsize Gemma-3-12B} & \rotatebox{90}{\scriptsize Gemma-3-27B} & \rotatebox{90}{\scriptsize Phi-4} & \rotatebox{90}{\scriptsize Dream-7B} & \rotatebox{90}{\scriptsize LLaDA-8B} & \rotatebox{90}{\scriptsize C-Haiku-4.5} & \rotatebox{90}{\scriptsize GPT-5} & \rotatebox{90}{\scriptsize Extr.\ quality} \\
\cmidrule[0.5pt](l){2-17}
& {\scriptsize $\Guard$ pass \%} & \cellcolor[RGB]{138,219,138}{\scriptsize 92} & \cellcolor[RGB]{132,218,132}{\scriptsize 96} & \cellcolor[RGB]{134,218,134}{\scriptsize 95} & \cellcolor[RGB]{131,217,131}{\scriptsize 97} & \cellcolor[RGB]{129,217,129}{\scriptsize 98} & \cellcolor[RGB]{128,217,128}{\scriptsize 99} & \cellcolor[RGB]{138,220,138}{\scriptsize 91} & \cellcolor[RGB]{134,218,134}{\scriptsize 95} & \cellcolor[RGB]{129,217,129}{\scriptsize 98} & \cellcolor[RGB]{128,217,128}{\scriptsize 99} & \cellcolor[RGB]{143,221,143}{\scriptsize 87} & \cellcolor[RGB]{143,221,143}{\scriptsize 87} & \cellcolor[RGB]{129,217,129}{\scriptsize 98} & \cellcolor[RGB]{127,216,127}{\scriptsize 100} &  \\
\cmidrule[0.5pt](l){2-17}
& {\scriptsize Qwen3-0.6B} & \cellcolor[RGB]{225,234,253}{\scriptsize 1.4} & \cellcolor[RGB]{213,225,252}{\scriptsize 2.7} & \cellcolor[RGB]{208,221,252}{\scriptsize 3.5} & \cellcolor[RGB]{220,230,252}{\scriptsize 1.9} & \cellcolor[RGB]{206,220,252}{\scriptsize 3.8} & \cellcolor[RGB]{205,220,252}{\scriptsize 3.9} & \cellcolor[RGB]{224,233,253}{\scriptsize 1.5} & \cellcolor[RGB]{218,228,252}{\scriptsize 2.2} & \cellcolor[RGB]{210,223,252}{\scriptsize 3.2} & \cellcolor[RGB]{195,212,251}{\scriptsize 5.7} & \cellcolor[RGB]{213,225,252}{\scriptsize 2.7} & \cellcolor[RGB]{219,229,252}{\scriptsize 2.0} & \cellcolor[RGB]{195,213,251}{\scriptsize 5.6} & \cellcolor[RGB]{194,212,251}{\scriptsize 5.8} & \cellcolor[RGB]{209,222,252}{\scriptsize 3.3\scalebox{0.55}{$\pm$1}} \\
& {\scriptsize Qwen3-1.7B} & \cellcolor[RGB]{213,226,252}{\scriptsize 2.7} & \cellcolor[RGB]{203,218,251}{\scriptsize 4.3} & \cellcolor[RGB]{194,212,251}{\scriptsize 5.9} & \cellcolor[RGB]{196,213,251}{\scriptsize 5.5} & \cellcolor[RGB]{182,203,250}{\scriptsize 8.4} & \cellcolor[RGB]{187,207,251}{\scriptsize 7.4} & \cellcolor[RGB]{218,229,252}{\scriptsize 2.1} & \cellcolor[RGB]{209,222,252}{\scriptsize 3.4} & \cellcolor[RGB]{197,214,251}{\scriptsize 5.2} & \cellcolor[RGB]{168,193,249}{\scriptsize 12.0} & \cellcolor[RGB]{202,217,251}{\scriptsize 4.4} & \cellcolor[RGB]{204,219,252}{\scriptsize 4.0} & \cellcolor[RGB]{160,188,249}{\scriptsize 14.2} & \cellcolor[RGB]{144,176,248}{\scriptsize 19.6} & \cellcolor[RGB]{188,208,251}{\scriptsize 7.1\scalebox{0.55}{$\pm$5}} \\
& {\scriptsize Qwen3-4B} & \cellcolor[RGB]{211,224,252}{\scriptsize 3.0} & \cellcolor[RGB]{203,218,251}{\scriptsize 4.3} & \cellcolor[RGB]{184,205,250}{\scriptsize 7.9} & \cellcolor[RGB]{196,213,251}{\scriptsize 5.5} & \cellcolor[RGB]{177,200,250}{\scriptsize 9.5} & \cellcolor[RGB]{180,202,250}{\scriptsize 8.9} & \cellcolor[RGB]{218,228,252}{\scriptsize 2.2} & \cellcolor[RGB]{204,219,252}{\scriptsize 4.0} & \cellcolor[RGB]{192,210,251}{\scriptsize 6.2} & \cellcolor[RGB]{159,187,249}{\scriptsize 14.6} & \cellcolor[RGB]{195,213,251}{\scriptsize 5.6} & \cellcolor[RGB]{206,220,252}{\scriptsize 3.7} & \cellcolor[RGB]{146,178,248}{\scriptsize 18.8} & \cellcolor[RGB]{107,151,246}{\scriptsize 34.5} & \cellcolor[RGB]{179,201,250}{\scriptsize 9.2\scalebox{0.55}{$\pm$8}} \\
& {\scriptsize Qwen3-8B} & \cellcolor[RGB]{214,226,252}{\scriptsize 2.6} & \cellcolor[RGB]{200,216,251}{\scriptsize 4.8} & \cellcolor[RGB]{183,204,250}{\scriptsize 8.2} & \cellcolor[RGB]{192,210,251}{\scriptsize 6.2} & \cellcolor[RGB]{173,197,250}{\scriptsize 10.5} & \cellcolor[RGB]{172,196,250}{\scriptsize 10.9} & \cellcolor[RGB]{218,229,252}{\scriptsize 2.1} & \cellcolor[RGB]{203,218,251}{\scriptsize 4.2} & \cellcolor[RGB]{185,206,250}{\scriptsize 7.7} & \cellcolor[RGB]{155,185,249}{\scriptsize 15.7} & \cellcolor[RGB]{199,215,251}{\scriptsize 5.0} & \cellcolor[RGB]{203,218,251}{\scriptsize 4.2} & \cellcolor[RGB]{139,173,248}{\scriptsize 21.5} & \cellcolor[RGB]{95,142,245}{\scriptsize 40.8} & \cellcolor[RGB]{174,198,250}{\scriptsize 10.3\scalebox{0.55}{$\pm$10}} \\
& {\scriptsize Qwen3-14B} & \cellcolor[RGB]{214,226,252}{\scriptsize 2.6} & \cellcolor[RGB]{200,216,251}{\scriptsize 4.8} & \cellcolor[RGB]{180,202,250}{\scriptsize 8.8} & \cellcolor[RGB]{189,208,251}{\scriptsize 6.9} & \cellcolor[RGB]{170,195,250}{\scriptsize 11.3} & \cellcolor[RGB]{169,194,249}{\scriptsize 11.6} & \cellcolor[RGB]{218,228,252}{\scriptsize 2.2} & \cellcolor[RGB]{200,216,251}{\scriptsize 4.7} & \cellcolor[RGB]{185,205,250}{\scriptsize 7.8} & \cellcolor[RGB]{151,181,248}{\scriptsize 17.1} & \cellcolor[RGB]{201,217,251}{\scriptsize 4.6} & \cellcolor[RGB]{207,221,252}{\scriptsize 3.6} & \cellcolor[RGB]{135,170,247}{\scriptsize 22.8} & \cellcolor[RGB]{101,146,245}{\scriptsize 37.8} & \cellcolor[RGB]{174,197,250}{\scriptsize 10.5\scalebox{0.55}{$\pm$9}} \\
& {\scriptsize Qwen3-32B} & \cellcolor[RGB]{214,226,252}{\scriptsize 2.6} & \cellcolor[RGB]{200,216,251}{\scriptsize 4.8} & \cellcolor[RGB]{178,200,250}{\scriptsize 9.4} & \cellcolor[RGB]{184,204,250}{\scriptsize 8.0} & \cellcolor[RGB]{169,194,249}{\scriptsize 11.7} & \cellcolor[RGB]{163,190,249}{\scriptsize 13.4} & \cellcolor[RGB]{219,230,252}{\scriptsize 2.0} & \cellcolor[RGB]{195,213,251}{\scriptsize 5.6} & \cellcolor[RGB]{183,204,250}{\scriptsize 8.1} & \cellcolor[RGB]{145,178,248}{\scriptsize 19.0} & \cellcolor[RGB]{196,213,251}{\scriptsize 5.4} & \cellcolor[RGB]{202,217,251}{\scriptsize 4.5} & \cellcolor[RGB]{126,164,247}{\scriptsize 26.2} & \cellcolor[RGB]{69,124,244}{\scriptsize 54.8} & \cellcolor[RGB]{166,192,249}{\scriptsize 12.5\scalebox{0.55}{$\pm$13}} \\
& {\scriptsize Gemma-3-4B} & \cellcolor[RGB]{217,228,252}{\scriptsize 2.2} & \cellcolor[RGB]{206,220,252}{\scriptsize 3.8} & \cellcolor[RGB]{192,210,251}{\scriptsize 6.2} & \cellcolor[RGB]{202,218,251}{\scriptsize 4.4} & \cellcolor[RGB]{186,206,250}{\scriptsize 7.4} & \cellcolor[RGB]{189,209,251}{\scriptsize 6.8} & \cellcolor[RGB]{218,229,252}{\scriptsize 2.1} & \cellcolor[RGB]{211,224,252}{\scriptsize 3.0} & \cellcolor[RGB]{201,217,251}{\scriptsize 4.6} & \cellcolor[RGB]{169,194,249}{\scriptsize 11.6} & \cellcolor[RGB]{205,219,252}{\scriptsize 4.0} & \cellcolor[RGB]{210,223,252}{\scriptsize 3.2} & \cellcolor[RGB]{160,188,249}{\scriptsize 14.3} & \cellcolor[RGB]{155,184,249}{\scriptsize 15.9} & \cellcolor[RGB]{191,210,251}{\scriptsize 6.4\scalebox{0.55}{$\pm$4}} \\
& {\scriptsize Gemma-3-12B} & \cellcolor[RGB]{215,227,252}{\scriptsize 2.5} & \cellcolor[RGB]{200,216,251}{\scriptsize 4.7} & \cellcolor[RGB]{183,204,250}{\scriptsize 8.2} & \cellcolor[RGB]{192,210,251}{\scriptsize 6.3} & \cellcolor[RGB]{173,197,250}{\scriptsize 10.7} & \cellcolor[RGB]{172,196,250}{\scriptsize 10.9} & \cellcolor[RGB]{218,229,252}{\scriptsize 2.1} & \cellcolor[RGB]{200,216,251}{\scriptsize 4.7} & \cellcolor[RGB]{185,205,250}{\scriptsize 7.8} & \cellcolor[RGB]{156,185,249}{\scriptsize 15.6} & \cellcolor[RGB]{197,214,251}{\scriptsize 5.3} & \cellcolor[RGB]{202,218,251}{\scriptsize 4.4} & \cellcolor[RGB]{142,175,248}{\scriptsize 20.1} & \cellcolor[RGB]{106,150,246}{\scriptsize 35.2} & \cellcolor[RGB]{176,199,250}{\scriptsize 9.9\scalebox{0.55}{$\pm$9}} \\
& {\scriptsize Gemma-3-27B} & \cellcolor[RGB]{213,225,252}{\scriptsize 2.7} & \cellcolor[RGB]{202,217,251}{\scriptsize 4.5} & \cellcolor[RGB]{181,202,250}{\scriptsize 8.7} & \cellcolor[RGB]{186,206,250}{\scriptsize 7.4} & \cellcolor[RGB]{170,195,250}{\scriptsize 11.4} & \cellcolor[RGB]{171,195,250}{\scriptsize 11.2} & \cellcolor[RGB]{218,229,252}{\scriptsize 2.1} & \cellcolor[RGB]{201,217,251}{\scriptsize 4.5} & \cellcolor[RGB]{187,207,251}{\scriptsize 7.3} & \cellcolor[RGB]{150,181,248}{\scriptsize 17.4} & \cellcolor[RGB]{199,216,251}{\scriptsize 4.9} & \cellcolor[RGB]{202,217,251}{\scriptsize 4.5} & \cellcolor[RGB]{137,172,248}{\scriptsize 22.0} & \cellcolor[RGB]{98,144,245}{\scriptsize 38.9} & \cellcolor[RGB]{173,197,250}{\scriptsize 10.5\scalebox{0.55}{$\pm$10}} \\
& {\scriptsize Phi-4} & \cellcolor[RGB]{222,232,253}{\scriptsize 1.7} & \cellcolor[RGB]{216,227,252}{\scriptsize 2.4} & \cellcolor[RGB]{228,236,253}{\scriptsize 1.1} & \cellcolor[RGB]{232,239,253}{\scriptsize 0.8} & \cellcolor[RGB]{228,236,253}{\scriptsize 1.1} & \cellcolor[RGB]{218,229,252}{\scriptsize 2.1} & \cellcolor[RGB]{235,241,253}{\scriptsize 0.6} & \cellcolor[RGB]{239,243,254}{\scriptsize 0.4} & \cellcolor[RGB]{238,243,254}{\scriptsize 0.5} & \cellcolor[RGB]{218,229,252}{\scriptsize 2.1} & \cellcolor[RGB]{233,240,253}{\scriptsize 0.7} & \cellcolor[RGB]{238,243,254}{\scriptsize 0.5} & \cellcolor[RGB]{207,221,252}{\scriptsize 3.6} & \cellcolor[RGB]{204,219,252}{\scriptsize 4.1} & \cellcolor[RGB]{223,233,253}{\scriptsize 1.5\scalebox{0.55}{$\pm$1}} \\
\cmidrule[0.3pt](l){2-17}
& {\scriptsize Dream-7B} & \cellcolor[RGB]{221,231,253}{\scriptsize 1.8} & \cellcolor[RGB]{211,224,252}{\scriptsize 3.0} & \cellcolor[RGB]{212,224,252}{\scriptsize 2.9} & \cellcolor[RGB]{215,226,252}{\scriptsize 2.5} & \cellcolor[RGB]{203,218,251}{\scriptsize 4.2} & \cellcolor[RGB]{208,222,252}{\scriptsize 3.4} & \cellcolor[RGB]{224,233,253}{\scriptsize 1.5} & \cellcolor[RGB]{223,232,253}{\scriptsize 1.6} & \cellcolor[RGB]{217,228,252}{\scriptsize 2.3} & \cellcolor[RGB]{192,210,251}{\scriptsize 6.2} & \cellcolor[RGB]{218,228,252}{\scriptsize 2.2} & \cellcolor[RGB]{218,229,252}{\scriptsize 2.1} & \cellcolor[RGB]{186,206,250}{\scriptsize 7.4} & \cellcolor[RGB]{179,201,250}{\scriptsize 9.1} & \cellcolor[RGB]{207,221,252}{\scriptsize 3.6\scalebox{0.55}{$\pm$2}} \\
& {\scriptsize LLaDA-8B} & \cellcolor[RGB]{215,227,252}{\scriptsize 2.4} & \cellcolor[RGB]{206,220,252}{\scriptsize 3.7} & \cellcolor[RGB]{190,209,251}{\scriptsize 6.6} & \cellcolor[RGB]{191,209,251}{\scriptsize 6.5} & \cellcolor[RGB]{179,201,250}{\scriptsize 9.0} & \cellcolor[RGB]{179,201,250}{\scriptsize 9.1} & \cellcolor[RGB]{216,228,252}{\scriptsize 2.3} & \cellcolor[RGB]{204,219,252}{\scriptsize 4.1} & \cellcolor[RGB]{189,208,251}{\scriptsize 6.9} & \cellcolor[RGB]{159,187,249}{\scriptsize 14.4} & \cellcolor[RGB]{203,218,251}{\scriptsize 4.3} & \cellcolor[RGB]{203,218,251}{\scriptsize 4.2} & \cellcolor[RGB]{156,185,249}{\scriptsize 15.6} & \cellcolor[RGB]{123,162,247}{\scriptsize 27.6} & \cellcolor[RGB]{182,203,250}{\scriptsize 8.3\scalebox{0.55}{$\pm$7}} \\
\cmidrule[0.3pt](l){2-17}
& {\scriptsize C-Haiku-4.5} & \cellcolor[RGB]{214,226,252}{\scriptsize 2.6} & \cellcolor[RGB]{198,215,251}{\scriptsize 5.1} & \cellcolor[RGB]{177,199,250}{\scriptsize 9.7} & \cellcolor[RGB]{182,204,250}{\scriptsize 8.3} & \cellcolor[RGB]{165,191,249}{\scriptsize 12.8} & \cellcolor[RGB]{162,189,249}{\scriptsize 13.8} & \cellcolor[RGB]{220,230,252}{\scriptsize 1.9} & \cellcolor[RGB]{201,217,251}{\scriptsize 4.6} & \cellcolor[RGB]{176,199,250}{\scriptsize 9.8} & \cellcolor[RGB]{145,177,248}{\scriptsize 19.2} & \cellcolor[RGB]{194,212,251}{\scriptsize 5.8} & \cellcolor[RGB]{198,214,251}{\scriptsize 5.2} & \cellcolor[RGB]{122,161,247}{\scriptsize 28.0} & \cellcolor[RGB]{65,121,243}{\scriptsize 57.3} & \cellcolor[RGB]{164,190,249}{\scriptsize 13.1\scalebox{0.55}{$\pm$14}} \\
& {\scriptsize GPT-5} & \cellcolor[RGB]{212,225,252}{\scriptsize 2.8} & \cellcolor[RGB]{199,215,251}{\scriptsize 4.9} & \cellcolor[RGB]{176,199,250}{\scriptsize 9.9} & \cellcolor[RGB]{182,203,250}{\scriptsize 8.5} & \cellcolor[RGB]{161,188,249}{\scriptsize 14.0} & \cellcolor[RGB]{159,187,249}{\scriptsize 14.5} & \cellcolor[RGB]{217,228,252}{\scriptsize 2.3} & \cellcolor[RGB]{201,217,251}{\scriptsize 4.6} & \cellcolor[RGB]{183,204,250}{\scriptsize 8.2} & \cellcolor[RGB]{139,173,248}{\scriptsize 21.1} & \cellcolor[RGB]{196,213,251}{\scriptsize 5.5} & \cellcolor[RGB]{198,215,251}{\scriptsize 5.0} & \cellcolor[RGB]{121,160,247}{\scriptsize 28.3} & \cellcolor[RGB]{38,102,242}{\scriptsize 80.4} & \cellcolor[RGB]{158,186,249}{\scriptsize 15.0\scalebox{0.55}{$\pm$19}} \\
\cmidrule[0.5pt](l){2-17}
& {\scriptsize Gen.\ quality} & \cellcolor[RGB]{216,227,252}{\scriptsize 2.4\scalebox{0.55}{$\pm$0}} & \cellcolor[RGB]{204,219,252}{\scriptsize 4.1\scalebox{0.55}{$\pm$1}} & \cellcolor[RGB]{189,208,251}{\scriptsize 6.9\scalebox{0.55}{$\pm$3}} & \cellcolor[RGB]{195,213,251}{\scriptsize 5.6\scalebox{0.55}{$\pm$2}} & \cellcolor[RGB]{179,202,250}{\scriptsize 9.0\scalebox{0.55}{$\pm$4}} & \cellcolor[RGB]{179,201,250}{\scriptsize 9.1\scalebox{0.55}{$\pm$4}} & \cellcolor[RGB]{220,230,252}{\scriptsize 1.9\scalebox{0.55}{$\pm$0}} & \cellcolor[RGB]{207,221,252}{\scriptsize 3.7\scalebox{0.55}{$\pm$1}} & \cellcolor[RGB]{193,211,251}{\scriptsize 6.1\scalebox{0.55}{$\pm$3}} & \cellcolor[RGB]{162,189,249}{\scriptsize 13.7\scalebox{0.55}{$\pm$5}} & \cellcolor[RGB]{202,218,251}{\scriptsize 4.3\scalebox{0.55}{$\pm$1}} & \cellcolor[RGB]{207,221,252}{\scriptsize 3.6\scalebox{0.55}{$\pm$1}} & \cellcolor[RGB]{149,180,248}{\scriptsize 17.8\scalebox{0.55}{$\pm$8}} & \cellcolor[RGB]{111,153,246}{\scriptsize 33.0\scalebox{0.55}{$\pm$21}} &  \\
\bottomrule
\end{tabular}
}%  end resizebox
\end{table}

% ─────────────────────────────────────────────────────────────────────────────
\section{Compute Resources}
\label{app:compute}
% ─────────────────────────────────────────────────────────────────────────────

All GPU experiments were run on NVIDIA A100 80GB nodes.
Frontier models were accessed over API, we report call counts
(one request per generated or extracted example). 

\begin{table}[h]
  \centering
  \caption{Compute consumed by experiments reported in this paper.
    GPU hours are wall-clock on NVIDIA A100 80GB nodes, summed across
    generator and extractor roles. API calls count requests to frontier
    models (one request per generated or extracted example) and are
    listed separately because they do not consume our GPU budget.}
  \small
  \begin{tabular}{lrrl}
    \toprule
    Experiment & A100 80GB (h) & API calls & Section \\
    \midrule
    Communication matrix (arithmetic)     & 182.5 & 85{,}644 & \Cref{sec:comm_matrix_results} \\
    Communication matrix (logic)          & 119.2 & 72{,}714 & \Cref{sec:scaling_ext} \\
    Lossless baseline (S-expression)      &   0.6 &     --   & \Cref{sec:comm_matrix_results} \\
    Lossless baseline (JSON AST)          &   0.5 &     --   & \Cref{sec:comm_matrix_results} \\
    Fine-tuning (arithmetic)              &  68.5 &     --   & \Cref{sec:ft_results} \\
    Fine-tuning (assembly)                &  57.3 &     --   & \Cref{sec:ft_results} \\
    Temperature sweep (Phi-4)             &  13.3 &     --   & \Cref{app:temperature} \\
    \midrule
    \textbf{Total}                        & \textbf{441.9} & \textbf{158{,}358} & \\
    \bottomrule
  \end{tabular}
  
  \label{tab:compute}
\end{table}

\ifnotarxiv{
    \input{broader_impact}
    \newpage
    \input{checklist}
}{
    % Optional else branch (leave empty if not needed)
}

\end{document}